\documentclass{siamart251216}

\usepackage{xcolor}  
\usepackage{hyperref}      
\usepackage{xurl}
\usepackage{url}           
\usepackage{graphicx}
\usepackage[sort]{natbib}
\usepackage{amsmath,amssymb,bm}
\usepackage[shortlabels]{enumitem}
\usepackage{setspace}
\usepackage{multirow,tabularx}
\usepackage{adjustbox}
\usepackage{setspace}
\usepackage{booktabs}
\usepackage{array}
\usepackage{placeins}

\newcommand{\as}[1]{\vert#1\vert}
\newcommand{\magg}[1]{\vert#1\vert ^{2}}
\newcommand{\norm}[1]{\left\Vert#1\right\Vert}

\newcommand{\dx}{{\rm d}x}
\newcommand{\dy}{{\rm d}y}

\usepackage{bm}
\renewcommand{\vec}[1]{\bm{#1}}

\DeclareMathOperator*{\NN}{NN}

\title{Learning functional components of partial differential equation models from data using neural networks}
\date{}

\author{Torkel E. Loman\thanks{Mathematical Institute, University of Oxford, United Kingdom (\email{torkel.loman@maths.ox.ac.uk}).}
\and
Yurij Salmaniw\thanks{Department of Mathematical and Physical Sciences, University of Sheffield, United Kingdom (\email{y.salmaniw@sheffield.ac.uk}).} 
\and
Antonio Le{\'o}n Villares\thanks{Department of Engineering, University of Oxford, United Kingdom (\email{antonio.leonvillares@stx.ox.ac.uk}).}
\and 
Jos\'e A. Carrillo\thanks{Mathematical Institute, University of Oxford, United Kingdom (\email{jose.carrillo@maths.ox.ac.uk}).} 
\and
Ruth E. Baker\thanks{Mathematical Institute, University of Oxford, United Kingdom (\email{ruth.baker@maths.ox.ac.uk}).}} 

\begin{document}

\maketitle

% ------------------------------------------------------------------------------------------------

\begin{abstract}

Partial differential equation (PDE) models frequently contain unknown functional terms that cannot be measured directly, limiting their predictive utility. While data-driven methods for estimating scalar PDE parameters are well established, the recovery of unknown functions remains comparatively underexplored. Here, we show that standard parameter estimation workflows can be extended to infer functional components of PDEs directly from data. Our approach embeds neural networks within the PDE framework, allowing unknown functions to be learned during training with high accuracy. Using nonlocal aggregation--diffusion equations as a case study, we infer interaction kernels and external potentials from steady-state observations. We systematically examine how reconstruction accuracy depends on factors such as the number and diversity of available solutions, sampling density, and measurement noise. The resulting framework retains the advantages of conventional PDE calibration approaches while extending them to functional inference: once trained, the PDE model can be used in the standard way to analyse system behaviour and generate predictions.

\end{abstract}

% --------------------------------------------------------------------------

%\tableofcontents

\section{Introduction}

Partial differential equations (PDEs) are a common language for modelling spatio-temporal phenomena across disciplines. They underpin mechanistic and phenomenological descriptions in biology and ecology~\cite{CantrellCosner2003,EdelsteinKeshet2005,Murray2002}, govern fluid, thermal, and pattern-forming systems in physics and chemistry~\cite{Batchelor1967,CarslawJaeger1959,CrossHohenberg1993,Kuramoto1984,LandauLifshitz1987}, and are workhorses for transport phenomena in chemical engineering~\cite{bird2002transport}. Beyond the natural sciences, PDEs are used in traffic and crowd modelling~\cite{Hughes2002,BellomoDogbe2011}, in image processing and computer vision~\cite{PeronaMalik1990,ChanVese2001}, and in finance~\cite{BlackScholes1973,WilmottHowisonDewynne1995}. A key reason for this ubiquity is their flexibility: the quantities governing dynamics are frequently spatial (or spatio-temporal) fields rather than scalar parameters. Examples include spatially dependent growth or mortality rates, source--sink fields, advection velocities, heterogeneous diffusivities or mobility maps, boundary fluxes, and interaction kernels governing agent--agent interactions.

This flexibility, however, comes at a cost. Predictive use of PDE models requires explicit specification of the functional forms appearing in the equations, together with reliable parameter estimates. In practice, neither the governing functions nor their parameters are typically known in full. Instead, a common structure arises across applications: the system state can often be observed, whereas key spatial components of the governing equations are only partially characterised or entirely inaccessible experimentally. In ecology, for example, population dynamics are frequently modelled using reaction--diffusion equations~\cite{Murray2002,CantrellCosner2003,EdelsteinKeshet2005}. Although population densities can often be estimated from field observations, the underlying spatially varying growth rates or carrying-capacity landscapes---quantities central to ecological forecasting and management~\cite{BerestyckiHamelRoques2005JMB}---are far more difficult to determine directly. In physics and chemical engineering, thermal and mass-transport processes are similarly ubiquitous~\cite{CarslawJaeger1959,bird2002transport}. While temperatures or concentrations may be measurable, the associated source terms, conductivities, or diffusivities are often uncertain or inaccessible~\cite{Isakov2017}. Related challenges occur in phase-field and pattern-formation models, where heterogeneous tilts or mobility maps encode effective energetic landscapes that cannot be directly observed experimentally~\cite{CrossHohenberg1993,Kuramoto1984,AllenCahn1979,Fife1979}. In nonlocal aggregation--diffusion systems describing processes such as cell sorting, the unknown structure may instead take the form of an external stimulus or an interaction kernel~\cite{MR3948738}. 

These examples highlight a common inverse problem: given partial observations of a system state, infer the unobserved sources, kernels, or other spatially varying functions appearing in the governing PDEs. Classical source-identification problems for parabolic equations in heat and transport theory provide a well-established instance of this framework~\cite{Isakov2017,Yamamoto2009Carleman}. However, most traditional parameter estimation methods are formulated for finite-dimensional parameter spaces. Extending these approaches to recover entire functional forms---such as spatially varying coefficients or interaction kernels---requires representations that are both mathematically principled and computationally tractable. In this work, we show how standard parameter fitting methodologies can be extended naturally to infer functional parameters in PDE models. Using a prototypical aggregation--diffusion system as a case study, we examine the reconstruction problem in detail, starting from idealised noise-free data and progressing towards more realistic settings involving incomplete and noisy observations. In particular, we investigate how qualitative model behaviour and stability properties affect identifiability, and how reconstruction accuracy deteriorates under sparse sampling and measurement noise.

We adopt a \textit{Universal PDE} (UPDE) framework~\cite{Rackauckas2020}, which provides a flexible yet interpretable representation of unknown spatial components while remaining computationally tractable. The core idea originates from non-spatial systems of ordinary differential equations, where UDEs were introduced as a means of embedding neural networks within differential equation models~\cite{Rackauckas2020}. In that setting, neural network parameters are trained alongside conventional scalar parameters, enabling unresolved or unknown dynamics to be learned directly from data. This strategy exploits the universal approximation capability of neural networks: sufficiently expressive architectures can approximate mappings $f:\mathbb{R}^n\to\mathbb{R}^m$ to arbitrary accuracy~\cite{HORNIK1991251}. Although UDE-type constructions have since been extended to PDEs, they have primarily been used as flexible black-box representations of complex or poorly understood dynamics~\cite{bolibar_2023,Xiajun2024,Xiaofeng2025}. Our objective here is fundamentally different. Rather than replacing the governing physics, we use the UPDE framework to parameterise unknown functions within an otherwise mechanistic PDE model. Neural networks therefore act as structured function approximators for spatially varying coefficients, potentials, or interaction kernels. Once trained, the inferred functions can either be interpreted as quantitative reconstructions of underlying spatial structure or used directly within the validated mechanistic model to generate forward predictions.

We illustrate this workflow using a prototypical aggregation--diffusion equation as a case study. This model is particularly well suited to our aims: it arises in a wide range of applications, is supported by a substantial theoretical and numerical literature that aids interpretation of inferred structures, and admits a formulation that is naturally suited to functional inference. In particular, the model can be written in terms of a nonlinear operator $\mathcal{T}$ whose fixed points coincide with steady-state solutions of the original PDE. This fixed-point formulation provides an efficient numerical framework for computing equilibria directly, irrespective of their dynamical stability or other properties \cite{carrillosalmaniwvillares2025}. The fixed-point structure also yields a natural cost functional for inference. In conventional parameter estimation, one seeks parameter values that minimise a cost function measuring disagreement between model predictions and observations. Within the UPDE framework, the same principle extends from finite-dimensional parameter vectors to unknown functions. Given observations of steady states, we define a cost functional using the fixed-point residual $u \mapsto |\mathcal{T}(u)-u|$. Since this residual vanishes precisely at numerical equilibria of the forward solver, it provides a natural measure of the mismatch between an observed steady state and the fixed-point equation induced by a candidate function. Moreover, because $\mathcal{T}$ is inherited directly from the numerical scheme, the resulting cost functional remains fully consistent with the underlying PDE: it preserves the equation structure, incorporates boundary conditions intrinsically, and avoids the need to differentiate potentially noisy data.

In this work, we show that perfect (infinite and noise-free) data permit the recovery of both single and multiple functional model components simultaneously. To bridge theory and practice, we then apply our framework to synthetic measurement data designed to mimic experimental conditions: recovery remains possible under sparse spatial sampling and in the presence of measurement noise, although accuracy degrades as noise levels increase. More generally, the success of reconstruction is governed by the interplay of three factors: the structure of the PDE (which functional components are unknown and how they enter the model); the available data (their coverage, resolution, and noise level); and the chosen solutions or experimental configurations (their number, diversity, and informativeness). In some cases, failure can be explained analytically, for example through structural non-identifiability, where recovery is provably impossible regardless of data quality. In others, the inverse problem is theoretically well-posed but reconstruction is hindered by practical limitations such as insufficient or noisy observations. We characterise these distinct regimes and provide simple diagnostics for distinguishing between them. Taken together, these results demonstrate that functional parameter recovery in PDE models is not merely a theoretical possibility but a practical one. With exact data, recovery is guaranteed at the theoretical level; with partial or noisy observations, it remains feasible across a broad class of systems. More broadly, our findings support the deployment of functional inference in PDE models and highlight the importance of systematic experimental design in settings that might otherwise appear non-identifiable.

% ------------------------------------------------------------------------------------------------

\section{The aggregation--diffusion equation}\label{section:methods_nlPDE}

As a concrete case study, we consider a canonical, nondimensionalised aggregation--diffusion equation with linear diffusion on the one-dimensional torus:
\begin{equation}\label{eq:model}
  \partial_t u
  = \partial_x ^2 u
  + \kappa\partial_x ( u\partial_x [W*u] ) + \partial_x ( u \partial_x V),
  \qquad x\in\mathbb{T},\ t>0,
\end{equation}
with nonnegative initial datum $u_0(x)$ of unit mass. The density $u(x,t)$ evolves under the combined effects of constant linear diffusion, nonlocal interactions mediated by the periodic interaction kernel $W(x)$ through the spatial convolution $(W\!\ast u)(x)=\int_{\mathbb{T}}W(x-y)u(y,t)\mathrm{d}y$, and transport induced by an external periodic potential $V(x)$. Our aim is to determine whether it is possible to recover the functions $W(x)$ and $V(x)$ given measurements of $u(x,t)$. Throughout, we assume that $\kappa \geq 0$ and that $\int_\mathbb{T} \lvert W(x) \rvert \dx = 1$ since only the product $\kappa{W}$ can be recovered and normalisation of $W$ then uniquely determines $\kappa$. In Section~\ref{sec:identifiability}, we consider only the product $\mathcal{W}=\kappa{W}$ when performing structural identifiability analysis to determine whether $W(x)$ and $V(x)$ can be recovered using ideal, noise-free data.

We select Equation~\eqref{eq:model} because it combines broad modelling relevance with an analytical structure that is particularly well suited to a systematic study of functional inference. In particular, the model possesses a \emph{gradient-flow structure}: its dynamics can be formulated as the gradient flow of an energy functional, and under appropriate conditions solutions converge to steady states as $t\to+\infty$, typically without sustained oscillations. As a result, steady states are both dynamically meaningful and practically accessible. The model also exhibits pattern-forming instabilities whose onset and local branches can be characterised rigorously through linear stability and bifurcation analysis~\cite{Carrillo2020LongTime,carrillo2025longtimebehaviourbifurcationanalysis}, providing a well-understood ground-truth structure against which the performance and limitations of the inference framework can be assessed. A further advantage is that equilibria coincide with fixed points of a nonlinear operator $\mathcal{T}$ (Proposition~\ref{prop:fixedpoint_iff_SS}). Steady states can therefore be characterised and computed directly as solutions of $\mathcal{T}u=u$, without time integration. In practice, these solutions can be obtained efficiently and accurately using a Newton--Krylov method~\cite{carrillosalmaniwvillares2025}. We refer the reader to Supplementary Information~\ref{sec:appendix_analytical_details} for further analytical details and references for Equation~\eqref{eq:model}.

The fixed-point formulation for the steady-states yields a natural loss functional, namely the residual $|\mathcal{T}u-u|$, which forms the basis of our learning procedure for $W(x)$ and $V(x)$. Accordingly, rather than constructing a residual for the fully time-dependent problem (as in, e.g.,~\cite{MR4896516}), we formulate the inverse problem using observations of steady-state densities. From an inference perspective, this corresponds to recovering model functions from data observed at a single asymptotic time point. Although such data contain less information than complete dynamical trajectories---and cannot, for example, distinguish between behaviours arising from different initial conditions that converge to the same equilibrium---they nevertheless provide a stringent test of functional recoverability while retaining computational tractability of the inference task.

The stationary problem reads
\begin{equation}\label{eq:model_SS}
  0= \partial_x \left( \partial_x u
  + \kappa u\partial_x [ W*u ]  +  u \partial_x V \right),
  \qquad x\in\mathbb{T}.
\end{equation}
For simplicity, we assume\footnote{Note that the tophat kernel does not satisfy this property, though we consider it in several instances because it remains smooth enough to ensure the stationary problem remains well-posed.}
\begin{equation}\label{hyp:regularity_of_inputs}
 W,V \in H^2(\mathbb{T}) \text{ with $W$ even periodic and $V$ periodic}.
\end{equation}
Then, a function $u_\text{ss}(x)$ solves Equation~\eqref{eq:model_SS} if and only if $u_\text{ss}(x)$ is a fixed point of the nonlinear map $\mathcal{T} : L^2 (\mathbb{T}) \mapsto L^2(\mathbb{T})$ where
 \begin{align}\label{eq:fixed_point_map_methods}
    \mathcal{T} u :=  \frac{\exp (- [ \kappa  W*u +  V] )}{Z(u)} \quad \text{with} \quad Z (u) := \int_\mathbb{T} \exp (-[ \kappa W*u +  V] )\,\dx.
\end{align}

% ------------------------------------------------------------------------------------------------

\section{Structural identifiability}\label{sec:identifiability}

In inverse problems, it is not uncommon for multiple distinct parameterisations to reproduce the same observations equally well; when this occurs, the underlying components cannot be uniquely inferred, and the model is said to be non-identifiable~\cite{WIELAND202160,simpson_parameter_2025}. Our first objective is to determine when the components $V(x)$ and $\mathcal{W}(x)=\kappa{W}(x)$ can be uniquely recovered under idealised, noise-free observations of the steady-state profiles: this is known as \emph{structural identifiability}. In subsequent sections, we then explore the question of \emph{practical identifiability}---the extent to which recovery remains stable and reliable in realistic computational settings with sparse or noisy data. For Equation~\eqref{eq:model_SS}, we understand structural identifiability in the following sense:

% ------------------------------------------------------------------------------------------------

\begin{definition}[Structural identifiability]\label{def:structural_identifiability}
    Let $\mathcal{F}$ denote an admissible class of unknown model components appearing in Equation~\eqref{eq:model_SS}, and let $\mathcal{O}$ denote the corresponding observation map assigning to each $\vec{f} \in \mathcal{F}$ the observed steady-state profile data generated by Equation~\eqref{eq:model_SS}. We say that the unknown model components are \emph{structurally identifiable} from the steady-state profile if $\mathcal{O}$ is injective on $\mathcal{F}$, that is, if 
    \[
    \mathcal{O}(\vec{f}_1)=\mathcal{O}(\vec{f}_2)
    \quad \Longrightarrow \quad
    \vec{f}_1=\vec{f}_2
    \text{\, for all \,} \vec{f}_1, \vec{f}_2 \in \mathcal{F}.
    \]
    Equivalently, distinct admissible model components must generate distinct observed steady-state profile data.
\end{definition}

We remark that the injectivity condition appearing in Definition~\ref{def:structural_identifiability} is understood globally over the admissible class $\mathcal{F}$. That is, we require uniqueness among all admissible model components, rather than within a local neighbourhood of a given set of parameters. Moreover, while Equation~\eqref{eq:model_SS} is written such that the interaction strength $\kappa$ is explicit, the nonlocal term depends only on the product $\kappa W$. In our explorations we therefore work with the effective interaction kernel $\mathcal{W}=\kappa\, W$ and formulate the structural identifiability results below in terms of the pair $(\mathcal{W}, V)$. This produces no loss of information once a normalisation of the kernel $W$ has been fixed; indeed, under the convention $\int_\mathbb{T} \as{W(x)} \dx = 1$, identification of $\mathcal{W}$ automatically determines $\kappa$ via $\kappa = \int_\mathbb{T}\mathcal{W}(x)\dx$.

% ------------------------------------------------------------------------------------------------

The unknown component $\vec{f}$ will be either a single interaction kernel $\mathcal{W}$ or a pair $(\mathcal{W},V)$ consisting of an interaction kernel and an external potential. Observations consist of either a single non-trivial steady-state profile $u_\text{ss}(x)$ or a collection of distinct profiles $(u_{1,\text{ss}}(x),\ldots,u_{K,\mathrm{ss}}(x))$. Accordingly, we consider observation maps of the form
\begin{equation*}
\mathcal O_1: \vec{f} \mapsto u_\text{ss},
\qquad
\mathcal O_2: \vec{f} \mapsto (u_{1,\text{ss}},\ldots,u_{K,\mathrm{ss}}),
\end{equation*}
where $\vec{f}\in\{\mathcal{W},(\mathcal{W},V)\}$. 

For a periodic function $g$ on $\mathbb T$, let $\widetilde g(k)$ denote its $k^{\mathrm{th}}$ Fourier coefficient. Then we have:

% ------------------------------------------------------------------------------------------------

\begin{theorem}\label{thm:identifiability}
Consider the stationary problem given by Equation~\eqref{eq:model_SS} on $\mathbb{T}$, and suppose that the effective interaction kernel $\mathcal{W}=\kappa W$ and external potential $V$ satisfy the conditions given in Equation~\eqref{hyp:regularity_of_inputs}. The following hold:

\begin{enumerate}[(i)]
    \item \textbf{Identifiability of $\mathcal{W}$ from a single profile.} Suppose that $V$ is known, so that $\mathcal{W}$ is the only unknown component. Let $u_\text{ss}$ be a non-trivial steady-state profile of Equation~\eqref{eq:model_SS}. Then $\widetilde{\mathcal{W}} (k)$ is uniquely recoverable from $u_\text{ss}$ at every Fourier mode $k$ such that $\widetilde{u_\text{ss}}(k)\neq 0$. In particular, if $\widetilde{u_\text{ss}}(k)\neq 0$ for all $k\in\mathbb{Z}\setminus \{0\}$ then $\mathcal{W}$ is structurally identifiable from the single profile $u_\text{ss}$.
    \item \textbf{Non-identifiability of $(\mathcal{W},V)$ from a single profile.}
    Let $u_\text{ss}$ be a steady-state profile of Equation~\eqref{eq:model_SS}. Then the pair $(\mathcal{W},V)$ is not structurally identifiable from the single profile $u_\text{ss}$.
    \item \textbf{Identifiability of $(\mathcal{W},V)$ from two distinct profiles.} Let $u_{1,\text{ss}}$ and $u_{2,\text{ss}}$ be two distinct, non-trivial steady-state profiles of Equation~\eqref{eq:model_SS} for the same pair $(\mathcal{W},V)$. Then $(\widetilde{\mathcal{W}}(k),\widetilde V(k))$ is uniquely recoverable from the pair $(u_{1,\text{ss}},u_{2,\text{ss}})$ at every Fourier mode $k$ such that $\widetilde{u_{1,\text{ss}}}(k)\neq \widetilde{u_{2,\text{ss}}}(k)$. In particular, if $\widetilde{u_{1,\text{ss}}}(k)\neq \widetilde{u_{2,\text{ss}}}(k)$ for all $k\in \mathbb{Z} \setminus \{ 0 \}$, then $(\mathcal{W},V)$ is structurally identifiable from $(u_{1,\text{ss}},u_{2,\text{ss}})$.
\end{enumerate}
\end{theorem}

% ------------------------------------------------------------------------------------------------

Theorem~\ref{thm:identifiability} reveals a simple relationship between the number of observed steady-state profiles and the number of spatial components to be recovered. When only the interaction kernel $\mathcal{W}$ is unknown, a single (nontrivial) steady-state profile is sufficient for structural identifiability, provided that the relevant Fourier modes are present in the observed profile. In contrast, when both $\mathcal{W}$ and $V$ are unknown, a single profile is insufficient: the nonlocal interaction kernel and the external potential can compensate for one another, giving rise to structural non-identifiability. However, structural identifiability can be recovered once two distinct steady-state profiles are available, since the additional profile closes the Fourier system mode-by-mode and restores identifiability under the natural nondegeneracy condition $\widetilde{u_{1,\text{ss}}}(k)\neq \widetilde{u_{2,\text{ss}}}(k)$ for $k\in \mathbb{Z} \setminus \{ 0 \}$.

% ------------------------------------------------------------------------------------------------

\subsection{Proof of Theorem 3.2}

We now prove Theorem~\ref{thm:identifiability}. Since the stationary problem given in Equation~\eqref{eq:model_SS} is linear in its unknown spatial components, the argument follows the general PDE identifiability framework developed in~\cite{MR4914013}, combined here with the Fourier representation available on the torus. 

% ------------------------------------------------------------------------------------------------

\subsubsection*{(i) Identifiability of $\mathcal{W}$ from a single profile} 

Assume that $V$ is known, and let $u_\text{ss}(x)$ be a non-trivial steady-state profile solving Equation~\eqref{eq:model_SS} with interaction kernel $\mathcal{W}$, so that $\mathcal{O}_1(\mathcal{W}) = u_\text{ss}(x)$. To verify structural identifiability in the sense of Definition~\ref{def:structural_identifiability}, let $\mathcal{W}_1$ and $\mathcal{W}_2$ be two admissible kernels such that the same profile $u_\text{ss}(x)$ solves Equation~\eqref{eq:model_SS} for both $\mathcal{W}_1$ and $\mathcal{W}_2$:
\begin{equation*}
    \mathcal{O}_1(\mathcal{W}_1) = \mathcal{O}_1(\mathcal{W}_2) = u_\text{ss}.
\end{equation*}
We show that $\mathcal{W}_1$ and $\mathcal{W}_2$ must agree at every Fourier mode for which $\widetilde{u_\text{ss}}(k)\neq 0$.

Since $u_\text{ss}$ is a steady state for both kernels, we have
\begin{align*}
  0&=\partial_x\Bigl(u_\text{ss}\,\partial_x\bigl( \log u_\text{ss} + \mathcal{W}_1*u_\text{ss} + V\bigr)\Bigr),\\  
  0&=\partial_x\Bigl(u_\text{ss}\,\partial_x\bigl( \log u_\text{ss} + \mathcal{W}_2*u_\text{ss} + V\bigr)\Bigr), 
\end{align*}
over $\mathbb{T}$. Subtracting these relations and integrating over $\mathbb{T}$ yields
\begin{equation*}
    u_\text{ss}\partial_x\bigl((\mathcal{W}_1-\mathcal{W}_2)*u_\text{ss}\bigr)=c_0,
\end{equation*}
for some constant $c_0\in\mathbb{R}$. By Proposition~\ref{prop:wellposed_stationary}, we know that $u_\text{ss}(x)>0$ in $\mathbb{T}$, and therefore dividing by $u_\text{ss}(x)$, integrating once more over $\mathbb{T}$ and enforcing periodicity gives $c_0=0$. Hence, 
\begin{equation*}
    \partial_x\bigl((\mathcal{W}_1-\mathcal{W}_2)*u_\text{ss}\bigr)=0
\qquad \text{in } \mathbb{T},
\end{equation*}
so that
\begin{equation*}
    (\mathcal{W}_1-\mathcal{W}_2)*u_\text{ss} \equiv c_1,
\end{equation*}
for some constant $c_1 \in \mathbb{R}$. 

As the observation map $\mathcal{O}_1$ is invariant under addition of constants to the kernel (i.e., $\mathcal{O}_1(\mathcal{W}) = \mathcal{O}_1(\mathcal{W}+c)$ for any constant $c\in\mathbb{R}$), we may, without loss of generality, assume that $\mathcal{W}_1$ and $\mathcal{W}_2$ have equal mean. Hence, integrating over $\mathbb{T}$ then yields $c_1 = 0$, and therefore
\begin{equation*}
    (\mathcal{W}_1-\mathcal{W}_2)*u_\text{ss} \equiv 0.
\end{equation*}
Taking the $k^{\textup{th}}$ Fourier coefficient and using the convolution theorem, we obtain
\begin{equation*}
    \bigl[\widetilde{\mathcal{W}_1}(k)-\widetilde{\mathcal{W}_2}(k)\bigr]\widetilde{u_\text{ss}}(k)=0,
\end{equation*}
for all $k \in \mathbb{Z} \setminus \{ 0 \}$, where we note that in the special case $V \equiv 0$, we need consider only $k\geq1$ since the solution will be even because $\mathcal{W}$ is even. Consequently, for every mode $k$ such that $\widetilde{u_\text{ss}}(k)\neq 0$ we have
\begin{equation*}
    \widetilde{\mathcal{W}_1}(k)=\widetilde{\mathcal{W}_2}(k),
\end{equation*}
and so $\widetilde{\mathcal{W}} (k)$ is uniquely recoverable from the single profile $u_\text{ss}(x)$ at all Fourier modes present in the spectrum of $u_\text{ss}(x)$. In particular, if $\widetilde{u_\text{ss}}(k)\neq 0$ for every $k\in\mathbb{Z} \setminus \{ 0 \}$, then $\mathcal{W}$ is structurally identifiable.

% ------------------------------------------------------------------------------------------------

\subsubsection*{(ii) Non-identifiability of $(\mathcal{W},V)$ from a single profile} 

Let $u_\text{ss}(x)$ be a steady-state profile solving Equation~\eqref{eq:model_SS} for some admissible pair $(\mathcal{W},V)$, and assume that $\mathcal{O}_1(\mathcal{W},V)= u_\text{ss}$. To show that $(\mathcal{W},V)$ is not structurally identifiable, it suffices to construct $(\mathcal{W}_0, V_0) \neq (\mathcal{W},V)$ such that $u_\text{ss}$ is also a solution of Equation~\eqref{eq:model_SS} but with components $(\mathcal{W}_0, V_0)$. 

To this end, let $Q \in H^2(\mathbb{T}) \cap L^2_{\textup{per}} (\mathbb{T})$ be any non-trivial function with zero mean, and define
\begin{equation*}
    \mathcal{W}_0 := \mathcal{W}+Q, \qquad V_0 := V - Q*u_\text{ss},
\end{equation*}
so that the means of $\mathcal{W}$ and $\mathcal{W}_0$ are equal. As $Q$ is nontrivial, $(\mathcal{W}_0,V_0)\neq (\mathcal{W},V)$. Moreover,
\begin{equation*}
    \mathcal{W}_0*u_\text{ss}+V_0 = (\mathcal{W}+Q)*u_\text{ss} + V - Q*u_\text{ss} = \mathcal{W}*u_\text{ss}+V.
\end{equation*}
It follows immediately that
\begin{equation*}
    \partial_x\Bigl(u_\text{ss}\,\partial_x\bigl( \log u_\text{ss} + \mathcal{W}_0*u_\text{ss} + V_0\bigr)\Bigr)
=
\partial_x\Bigl(u_\text{ss}\,\partial_x\bigl( \log u_\text{ss} + \mathcal{W}*u_\text{ss} + V\bigr)\Bigr)
=0,
\end{equation*}
and therefore $u_\text{ss}$ is also a solution of Equation~\eqref{eq:model_SS} for the pair $(\mathcal{W}_0, V_0)$. Therefore, the observation map $\mathcal{O}_1 : (\mathcal{W},V) \mapsto u_\text{ss}$ is not injective, and so $(\mathcal{W},V)$ is not structurally identifiable from a single steady-state profile. 

% ------------------------------------------------------------------------------------------------

\subsubsection*{(iii) Identifiability of $(\mathcal{W},V)$ from two distinct profiles} 

Let $u_{1,\text{ss}}(x)$ and $u_{2,\text{ss}}(x)$ be two distinct, non-trivial steady-state profiles solving Equation~\eqref{eq:model_SS} for the same admissible pair $(\mathcal{W},V)$, with
\begin{equation*}
    \mathcal{O}_2(\mathcal{W},V) = (u_{1,\text{ss}}, u_{2,\text{ss}}).
\end{equation*}
To verify structural identifiability in the sense of Definition~\ref{def:structural_identifiability}, let $(\mathcal{W}_1,V_1)$ and $(\mathcal{W}_2,V_2)$ be two admissible pairs with
\begin{equation*}
    \mathcal{O}_2(\mathcal{W}_1,V_1)=\mathcal{O}_2(\mathcal{W}_2,V_2)=(u_{1,\text{ss}}(x),u_{2,\text{ss}}(x)).
\end{equation*}
Different from case (i), since $V(x)$ and hence the steady states $u_{1,\mathrm{ss}}(x)$ and $u_{2,\mathrm{ss}}(x)$ need not be even, we now work with the full complex Fourier series indexed by $k \in \mathbb{Z}$ so that the pair of modes $k$ and $-k$ together encode the corresponding cosine and sine components. We show that $\mathcal{W}_1$ and $\mathcal{W}_2$, as well as $V_1$ and $V_2$, must agree at every Fourier mode $k$ for which
\begin{equation*}
    \widetilde{u_{1,\text{ss}}}(k)\neq \widetilde{u_{2,\text{ss}}}(k).
\end{equation*}

Since $u_{1,\text{ss}}$ and $u_{2,\text{ss}}$ are steady states for both pairs $(\mathcal{W}_1,V_1)$ and $(\mathcal{W}_2,V_2)$, the same argument as in part (i) implies that, for each $j=1,2$, there exist constants $c_{1,j}, c_{2,j}\in\mathbb{R}$ such that
\begin{equation*}
  \log u_{j,\mathrm{ss}} + \mathcal{W}_1*u_{j,\mathrm{ss}} + V_1 = c_{1,j}
\quad \text{and} \quad
\log u_{j,\mathrm{ss}} + \mathcal{W}_2*u_{j,\mathrm{ss}} + V_2 = c_{2,j},  
\end{equation*}
over $\mathbb{T}$. Subtracting these relations gives
\begin{equation*}
   (\mathcal{W}_1-\mathcal{W}_2)*u_{j,\mathrm{ss}} + (V_1-V_2) = c_j,
\qquad j=1,2, 
\end{equation*}
for suitable constants $c_j\in\mathbb{R}$. Taking the $k^{\textup{th}}$ Fourier coefficient of both sides and using the convolution theorem, the constants $c_j$ vanish, and we obtain the equivalent relations
\begin{equation*}
    \bigl[\widetilde{\mathcal{W}_1}(k)-\widetilde{\mathcal{W}_2}(k)\bigr]\widetilde{u_{j,\mathrm{ss}}}(k) + \bigl[\widetilde{V_1}(k)-\widetilde{V_2}(k)\bigr] =0,
\qquad j=1,2, \quad k \in \mathbb{Z}\setminus \{ 0 \},
\end{equation*}
where we emphasise that we consider all nonzero wavenumbers since $V$ (and therefore $u_{1,\mathrm{ss}}$ and $u_{2,\mathrm{ss}}$) need not be even. In other words, for each wavenumber $k \in \mathbb{Z}\setminus \{ 0 \}$ we obtain the linear system:
\begin{equation*}
    \begin{pmatrix}
\widetilde{u_{1,\text{ss}}}(k) & 1 \\
\widetilde{u_{2,\text{ss}}}(k) & 1
\end{pmatrix}
\begin{pmatrix}
\widetilde{\mathcal{W}_1}(k)-\widetilde{\mathcal{W}_2}(k) \\
\widetilde{V_1}(k)-\widetilde{V_2}(k)
\end{pmatrix}
=
\begin{pmatrix}
0\\
0
\end{pmatrix}.
\end{equation*}
The determinant of this system is $\widetilde{u_{1,\text{ss}}}(k)-\widetilde{u_{2,\text{ss}}}(k)$; therefore, whenever $\widetilde{u_{1,\text{ss}}}(k)\neq \widetilde{u_{2,\text{ss}}}(k)$, the matrix is invertible, and hence
\begin{equation*}
    \widetilde{\mathcal{W}_1}(k)=\widetilde{\mathcal{W}_2}(k),
\qquad
\widetilde{V_1}(k)=\widetilde{V_2}(k).
\end{equation*}
Thus, $(\widetilde{\mathcal{W}}(k),\widetilde V(k))$ is uniquely recoverable from $(u_{1,\text{ss}},u_{2,\text{ss}})$ for every nonzero mode $k$ satisfying $\widetilde{u_{1,\text{ss}}}(k)\neq \widetilde{u_{2,\text{ss}}}(k)$. In particular, if
\begin{equation*}
    \widetilde{u_{1,\text{ss}}}(k)\neq \widetilde{u_{2,\text{ss}}}(k)
\quad \text{for all } k \in \mathbb{Z} \setminus \{ 0 \},
\end{equation*}
then $(\mathcal{W},V)$ is structurally identifiable from $(u_{1,\text{ss}},u_{2,\text{ss}})$. This completes the proof.

\medskip
The argument in part~\textup{(iii)} also suggests an immediate extension to more than two observed profiles. Indeed, if one has access to $J\geq2$ distinct steady-state profiles generated by the same pair $(\mathcal{W},V)$, then for each Fourier mode $k \in \mathbb{Z} \setminus \{ 0 \}$ one obtains an overdetermined linear system for the two unknown coefficients $\widetilde{\mathcal{W}}(k)$ and $\widetilde{V}(k)$. In particular, recovery at mode $k$ is possible as soon as there exists at least one pair of profiles $u_{i,\mathrm{ss}}$ and $u_{j,\mathrm{ss}}$ for which $\widetilde{u_{i,\mathrm{ss}}}(k)\neq \widetilde{u_{j,\mathrm{ss}}}(k)$.

We also note that the spectral nondegeneracy conditions appearing in Theorem~\ref{thm:identifiability} are not merely technical artefacts. In particular, the local bifurcation analysis presented in Supplementary Information~\ref{sec:appendix_analytical_details} shows that near a bifurcation point the emerging non-trivial branch is, to leading order, a single cosine mode. Thus, one expects steady-state profiles whose spectra are highly concentrated in a small number of modes, with the remaining coefficients vanishing or becoming very small. Such profiles are therefore intrinsically less informative for recovery, especially in the presence of numerical error or noise.

To conclude, we remark that related identifiability questions for nonlocal gradient-flow models have recently been studied in the time-dependent setting~\cite{MR4896516}, where recovery of interaction kernels from noisy discrete trajectory data using a sparse partial-inversion framework is considered. Their approach is designed for evolutionary data and relies on different analytical and computational tools than those used here. By contrast, our setting is stationary: rather than exploiting full, time-dependent trajectories, we work with multiple steady-state profiles featuring different amplitudes and stability properties, and obtain mode-by-mode identifiability information through the steady-state structure and Fourier analysis. In this sense, the two approaches are complementary, addressing different data regimes and different inverse formulations.

% ------------------------------------------------------------------------------------------------

\section{Computational recovery from data}\label{section:methods_udes}

For the PDE described in Equation~\eqref{eq:model_SS}, we now seek to recover the interaction kernel $W(x)$, the external potential $V(x)$, and (with a normalised interaction kernel) the scalar parameter $\kappa$, from steady-state data. To render this inverse problem finite-dimensional, we replace the unknown functions $W(x)$ and $V(x)$ with neural network approximations $\NN_W(x;\vec{\theta}_W)$ and $\NN_V(x;\vec{\theta}_V)$, yielding the following UPDE:\begin{equation}\label{eq:model_NN_version}
  0 = \partial_x ^2 u + \kappa\,\partial_x ( u \,\partial_x ( {\NN}_W (x; \vec{\theta}_W) *u) ) + \partial_x ( u \,\partial_x \, {\NN}_V(x; \vec{\theta}_V) ),
\end{equation}
depending on the neural network parameterisation given by $\vec{\theta}_W=(\theta_{W,1},\ldots, \theta_{W,N})$ and $\vec{\theta}_V=(\theta_{V,1}, \ldots,\theta_{V,M})$. Provided the neural networks are sufficiently expressive, there exists a parameterisation $\vec{\theta}_W^*$ and $\vec{\theta}_V^*$ for which $\NN_W(x; \vec{\theta}^*_W) \equiv W^*(x)\approx W(x)$ and $\NN_V(x; \vec{\theta}^*_V) \equiv V^*(x) \approx V(x)$ for all $x\in\mathbb{T}$. As discussed in Section~\ref{sec:identifiability}, using the normalisation of the kernel, one can then recover the parameter $\kappa$ directly from $W^{*}(x)$.

To perform inference, we require a cost functional whose minimisers correspond to solutions of Equation~\eqref{eq:model_SS}. Several choices are possible. A natural option is the \emph{PDE residual} loss
\begin{align}\label{eq:loss_function_PDE}
    R_{\textup{PDE}}(u) := \norm{\partial_x ( \partial_x u
  + \kappa\, u\, \partial_x [ W*u ]  +  u\, \partial_x V)}_X,
\end{align}
where $X$ is an appropriate Banach space, taken here to be $L^2(\mathbb{T})$. This residual directly measures the extent to which a candidate solution fails to satisfy the steady-state PDE. However, its evaluation requires differentiation of the observed solution and therefore additional regularity assumptions. However, as outlined previously, a second option is to instead exploit the fixed-point formulation developed in Section~\ref{section:methods_nlPDE}. Since steady states coincide with fixed points of the nonlinear operator $\mathcal{T}$, we define the \emph{fixed-point residual} loss
\begin{align}\label{eq:loss_function_FP}
R_{\textup{FP}}(u) := \norm{\mathcal{T}u-u}_X.
\end{align}
This residual vanishes precisely at equilibria of Equation~\eqref{eq:model} and avoids the need to differentiate. Moreover, because the numerical scheme used to generate the synthetic data is based on the fixed-point map $\mathcal{T}$~\cite{carrillosalmaniwvillares2025}, the resulting cost functional remains fully consistent with the underlying PDE and its boundary conditions. Other residual constructions are also possible, including weak-form residuals based on the variational formulation of Equation~\eqref{eq:model_SS}. However, for the reasons outlined above, all results presented in the main text use the fixed-point residual. 

Given the neural network approximations of $W(x)$ and $V(x)$, the inverse problem is therefore reduced to the finite-dimensional optimisation problem of identifying the parameter vector $(\kappa^*,\vec{\theta}_W^*,\vec{\theta}_V^*)$ that minimises the fixed-point residual defined in Equation~\eqref{eq:loss_function_FP} for given steady-state data. Throughout this work, we employ small, fully connected feedforward neural networks, primarily using the softplus activation function---all architectural details are provided in  Section~\ref{sup_section:detailed_methods_nn_arch}. Optimisation is performed using a two-stage procedure. We first run the Adam optimiser for a relatively large number of iterations (typically between $10^4$ and $2\times10^6$, depending on the complexity of the inference problem), before switching to an LBFGS optimiser to refine the solution so that it converges to a nearby local minimum~\cite{liu_limited_1989,kingma_adam_2017}. This combination of optimisers has been widely adopted in the UDE literature~\cite{Rackauckas2020,philipps_current_2025} and it proved effective across all examples considered here. To mitigate issues relating to convergence to local minima, we employ an ensemble multi-start optimisation strategy, again following established practice in UDE inference~\cite{philipps_current_2025}. Independent optimisation runs are performed from multiple initialisations, allowing the best-performing solutions to be identified while also exploiting parallel computation. 

We present results for a range of different problems in Section~\ref{sec:function_recovery_and_identifiability}. To demonstrate that our conclusions are not specific to the choice of loss function,  Figure~\ref{fig:fig_1_sup_alt_loss} reproduces selected results using the PDE-residual loss given in Equation~\eqref{eq:loss_function_PDE}. In addition, although we primarily employ neural networks as universal function approximators, the methodology is not restricted to this choice: any finite-dimensional representation capable of approximating continuous functions may be used. For the aggregation--diffusion equation, a natural alternative is a truncated Fourier expansion, with Fourier coefficients playing the role of trainable parameters. We implement this approach in Section~\ref{sup_section:fouier_expansion_ufa} and reproduce selected results  (Figure~\ref{fig:fig_1_sup_fouier_ufa}).

% ------------------------------------------------------------------------------------------------

\section{Functional recovery and practical identifiability for unknown functions}\label{sec:function_recovery_and_identifiability}

\begin{figure}[htbp]
    \centering
    \includegraphics[width=0.72\linewidth]{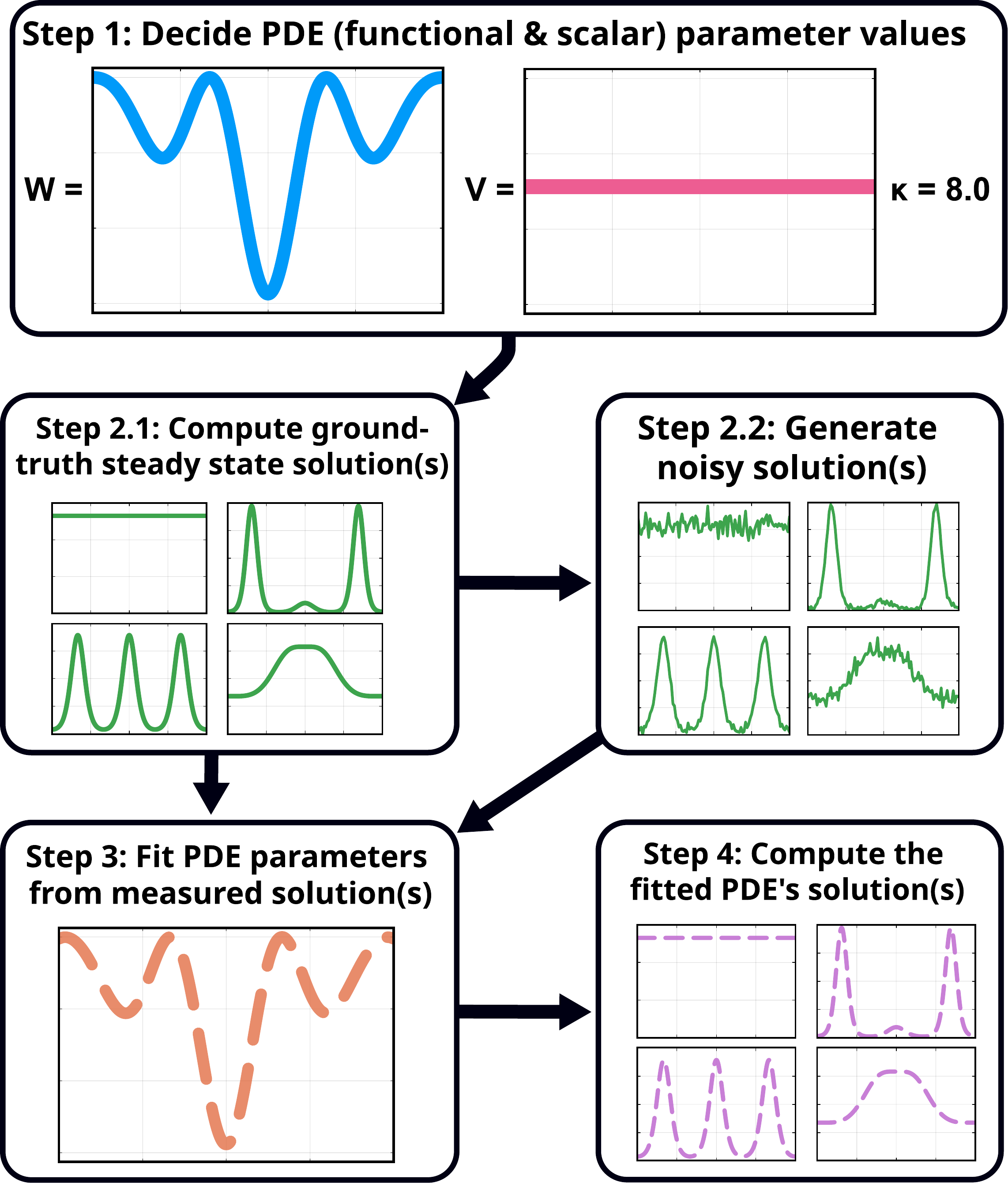}
    \caption{\textbf{Diagram of the standard workflow for analysing the feasibility of PDE recovery from data.} Step 1: specify the functional forms for $W$ and $V$, and the value of $\kappa$, for the PDE we wish to analyse. Details for all PDEs analysed in this work can be found in Table~\ref{tab:PDE_list}. Step 2.1: From the PDE, compute steady-state solution profiles using the approach described in~\cite{carrillosalmaniwvillares2025}. Step 2.2: use the approach described in Supplementary Section~\ref{sup_section:detailed_methods_noisy_data} to generate downsampled, noisy, solution profiles. Step 3: assume that some subset of the parameters is unknown (in Section~\ref{sec:results_normal_fits} and Section~\ref{sec:results_noisy_fits} only $W$ is assumed to be unknown, while Section~\ref{sec:results_WVk_fits} considers multiple unknown parameters). Next, attempt to recover these from measured solution profiles using the approach described in Section~\ref{section:methods_udes}. The success of the fitting process can be evaluated by comparing the estimated parameters to the true ones (as chosen in Step 1). Step 4: The fitting process can be further evaluated by computing the steady-state solutions for the fitted PDE, which then can be compared to true ones computed in Step 2.1.}
    \label{fig:workflow_flowchart}
\end{figure}

Having established in Section~\ref{def:structural_identifiability} the conditions under which functional recovery is theoretically possible, that is, when the unknown functions are structurally identifiable, we now apply the recovery framework introduced in Section~\ref{section:methods_udes} to a series of practical examples. Following the workflow shown in Figure~\ref{fig:workflow_flowchart}, we investigate the circumstances under which functional recovery succeeds and characterise the principal modes by which it can fail.

% ------------------------------------------------------------------------------------------------

% ------------------------------------------------------------------------------------------------

As part of this analysis, we examine \emph{practical identifiability}: given a non-ideal dataset, we assess whether the optimisation procedure can reliably recover a single PDE parameterisation that provides a good fit to the observations. While well-established methods exist for assessing the practical identifiability of scalar parameters~\cite{kreutz_profile_2013}, analogous methodologies for functional parameters remain comparatively underdeveloped. Throughout this work, we therefore adopt the ensemble-plot approach introduced in~\cite{loman_funcident_2025} (see, for example, Figure~\ref{fig:bif_analysis}). The optimisation is repeated from multiple random initialisations, and the ensemble of inferred functions whose loss values lie below a prescribed threshold (typically taken to be twice the minimum loss) is plotted. If these functions collapse onto a common profile, this provides evidence that the unknown function is practically identifiable from the available data.

% ------------------------------------------------------------------------------------------------

\subsection{The interaction kernel \texorpdfstring{$W$}{W} can be recovered from steady state solutions}\label{sec:results_normal_fits}

\begin{figure}[htbp]
    \centering
    \includegraphics[width=0.85\linewidth]{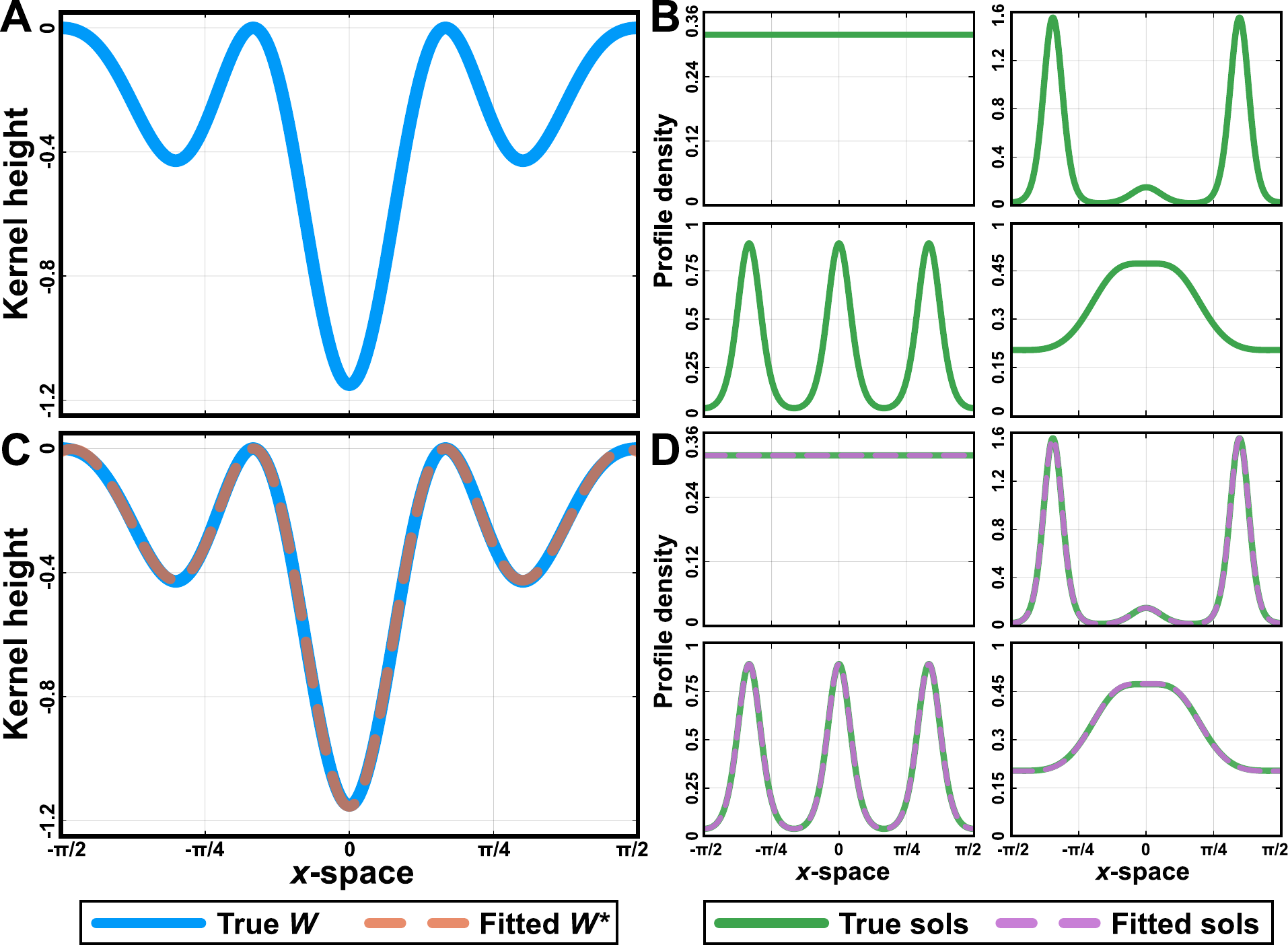}
    \caption{\textbf{The interaction kernel $W$ can be recovered from steady-state solution profiles.} (A) For the nonlocal PDE given in Equation~\eqref{eq:model} we set the potential $V \equiv 0$, the interaction strength parameter $\kappa = 8$, and use a multi-modal interaction kernel $W$ (described in Supplementary Section~\ref{sup_section:pde_list}, depicted with a solid blue line). (B) From the fully-determined PDE, we compute the four steady-state solutions (solid green lines). (C) The interaction kernel, $W$, can be recovered using the UDE-based approach outlined in Section~\ref{section:methods_udes} and Figure~\ref{fig:workflow_flowchart}. We confirm that the fitted $W^*$ (dashed red line) replicates the true $W$ (solid blue line). (D) Finally, we compute the solutions of the PDE instance $(W^*,V,\kappa)$ (dashed purple lines), confirming these provide an excellent match to the ground-truth solutions (solid green lines).}
    \label{fig:fit_W_to_sols}
\end{figure}

We begin with the simplest setting, in which only a single functional component, the interaction kernel $W(x)$, is unknown, while the external potential $V(x)$ and interaction strength $\kappa$ are assumed to be known. We first specify the ground-truth functions $W(x)$ and $V(x)$, together with the scalar parameter $\kappa$ (Figure~\ref{fig:fit_W_to_sols}A), and compute the corresponding steady-state solution profiles of Equation~\eqref{eq:model_SS} using the fixed-point formulation defined in Equation~\eqref{eq:fixed_point_map_methods} (Figure~\ref{fig:fit_W_to_sols}B). We then replace $W$ with a neural network within the UPDE framework (Section~\ref{section:methods_udes}) and fit the resulting model to the complete set of noise-free steady-state profiles. The chosen network architecture is sufficiently expressive to represent the ground-truth interaction kernel (Figure~\ref{fig:fig_1_sup_direct_fit}), and optimisation converges reliably (Figure~\ref{fig:fig_1_sup_loss_traces}).

The inferred interaction kernel $W^*$ is visually indistinguishable from the ground truth (Figure~\ref{fig:fit_W_to_sols}C), demonstrating that accurate functional recovery is possible in this idealised setting. Moreover, the ensemble analysis indicates that no alternative interaction kernels achieve comparably good fits, providing evidence that the recovered function is also practically identifiable (Supplementary Figure~\ref{fig:fig_1_sup_practical_identifiability}).

% ------------------------------------------------------------------------------------------------

% ------------------------------------------------------------------------------------------------

We emphasise that recovering an interaction kernel with a small loss value, or even one that is visually indistinguishable from the ground truth, does not necessarily imply that it reproduces the same set of steady-state solutions. This follows from the analysis in~\cite[Remark 4.7]{Carrillo2020LongTime}, where a family of kernels ${W_s(x)}_{s\geq1}$ converges uniformly to a limiting kernel $-w_1(x)$ as $s\to\infty$ (here, $w_k$ denotes the basis element defined in Equation~\eqref{eq:ONB_L2s}). Despite this uniform convergence, every kernel $W_s$ possesses infinitely many bifurcation points, whereas the limiting kernel $-w_1$ has exactly one. Thus, kernels that are arbitrarily close in function space can nevertheless exhibit qualitatively different bifurcation structures and, consequently, different sets of steady-state solutions. We illustrate this phenomenon numerically in Figure~\ref{fig:fig_1_sup_bad_sols}. For the example considered in Figure~\ref{fig:fit_W_to_sols}, however, this issue does not arise: the recovered interaction kernel $W^*$ reproduces essentially the same steady-state solution set as the ground-truth kernel $W$ (Figure~\ref{fig:fit_W_to_sols}D).

We next investigate whether the complete set of steady-state solutions is required to recover the interaction kernel. Motivated by the structural identifiability result of Theorem~\ref{thm:identifiability}, we repeat the inference procedure using only a single spatially varying steady-state profile from Figure~\ref{fig:fit_W_to_sols}B (again assuming noise-free data). In each case, the recovered interaction kernel satisfies $W^*\approx W$ and reproduces essentially the same steady-state solution set as the ground-truth kernel (Figure~\ref{fig:fig_1_sup_single_sols}). These results demonstrate that, in agreement with the theoretical analysis, a single informative steady-state profile is sufficient for accurate functional recovery in this setting. Finally, to demonstrate that these conclusions are not specific to the example in Figure~\ref{fig:fit_W_to_sols}, we repeat the analysis for piecewise linear (Figure~\ref{fig:fig_1_sup_alt_Ws}A,B), smooth (Figure~\ref{fig:fig_1_sup_alt_Ws}C), and mixed piecewise linear/smooth (Figure~\ref{fig:fig_1_sup_alt_Ws}D) interaction kernels. In every case, the inference procedure accurately recovers the interaction kernel and reproduces the corresponding steady-state solution set.

% ------------------------------------------------------------------------------------------------

\subsection{The interaction kernel \texorpdfstring{$W$}{W} can be recovered from noisy solution profiles}\label{sec:results_noisy_fits}

\begin{figure}[htbp]
    \centering
    \includegraphics[width=0.90\linewidth]{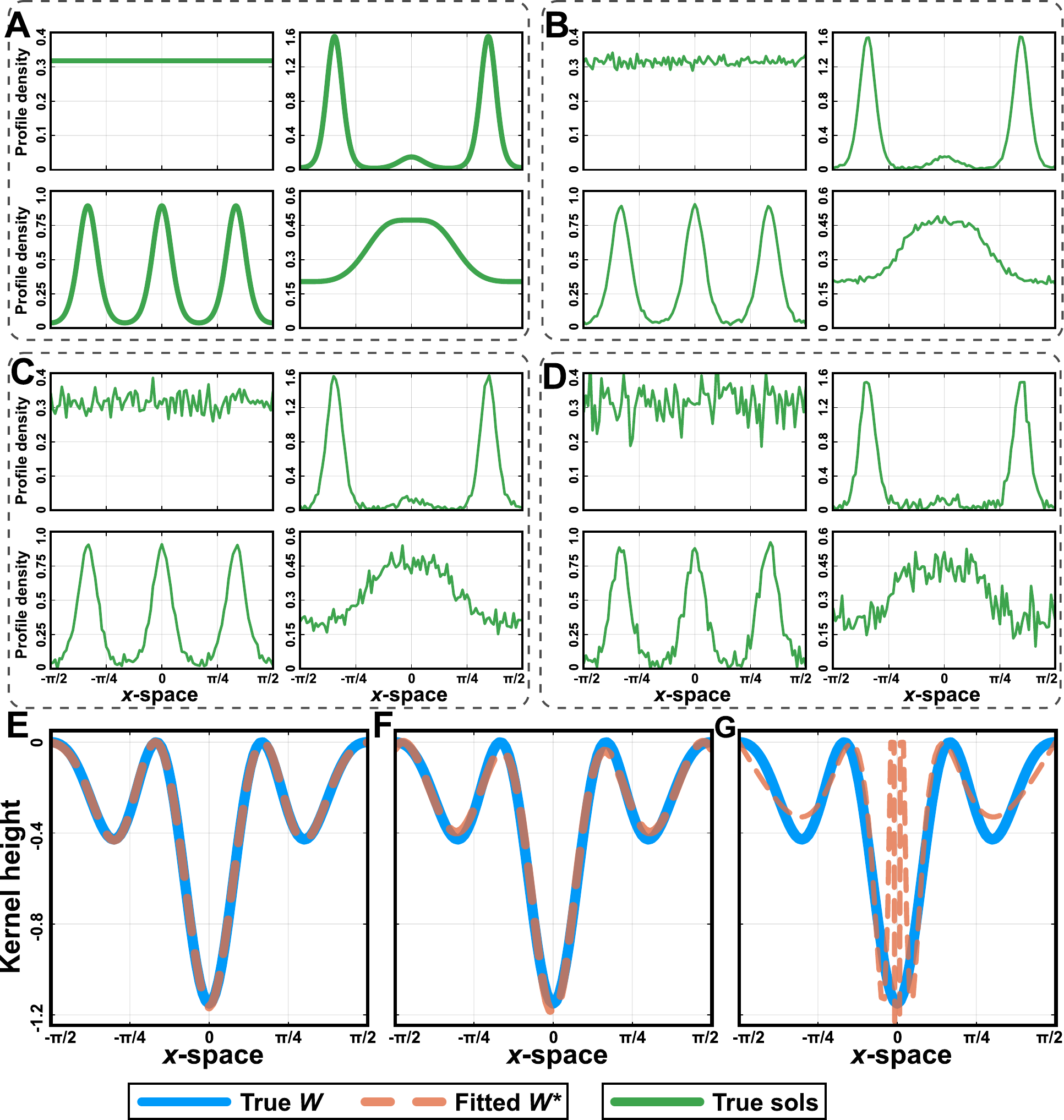}
    \caption{\textbf{The interaction kernel $W$ can be recovered from noisy steady-state solutions.} (A) Solutions of the PDE described in Figure~\ref{fig:fit_W_to_sols}. (B-D) We downsample the solutions in A to $100$ datapoints, and also add low (B), medium (C), and high (D) levels of noise (Section \ref{sup_section:detailed_methods_noisy_data}). (E-G) Using the optimisation procedure from Figure~\ref{fig:workflow_flowchart}, we attempt to recover $W$ using the noisy solution profiles. For low (E) and medium (F) noise levels, the fitted $W^*$ (dashed red lines) is very close to the ground truth $W$ (solid blue lines). However, for the high noise levels (G), we can no longer accurately recover $W$.}
    \label{fig:noisy_ss_fit}
\end{figure}

In the previous section, we assumed complete knowledge of the steady-state solution set, sampled with arbitrarily fine spatial resolution and without measurement noise. To assess the applicability of our framework to experimental data, we now consider sparse and noisy observations. Starting from the ground-truth PDE parameters and steady-state profiles used in Figure~\ref{fig:fit_W_to_sols}, we downsample each profile to $100$ spatial samples over the domain $[-\pi/2,\pi/2)$ and add Gaussian measurement noise at varying amplitudes (Figure~\ref{fig:noisy_ss_fit}A--D and Section~\ref{sup_section:detailed_methods_noisy_data}). We then repeat the inference procedure using datasets with low, medium, and high noise levels. 

% ------------------------------------------------------------------------------------------------

% ------------------------------------------------------------------------------------------------

The interaction kernel can be accurately recovered from data with low and me\-dium noise levels (Figure~\ref{fig:noisy_ss_fit}E,F), but accuracy deteriorates substantially in the high-noise regime (Figure~\ref{fig:noisy_ss_fit}G). Moreover, for the low- and medium-noise datasets, the recovered interaction kernels reproduce essentially the same steady-state solution sets as the ground-truth kernel (Figure~\ref{fig:fig_2_sup_noisy_fit_sols}). To assess robustness, we repeat the analysis using independently generated noise realisations with the same statistical properties, obtaining consistent results (Figure~\ref{fig:fig_2_sup_repeats}). In all cases, the recovered interaction kernels remain practically identifiable according to the ensemble analysis (Figure~\ref{fig:fig_2_sup_practical_identifiability}). Finally, we note that the ability to recover the interaction kernel depends not only on the quantity and quality of the data, but also on the kernel itself. For a different ground-truth interaction kernel, recovery is no longer possible under comparable noise levels (Figure~\ref{fig:fig_2_sup_alt_W}), illustrating that recoverability depends on both the observational data and the underlying PDE.

% ------------------------------------------------------------------------------------------------

\begin{figure}[!t]
    \centering
    \includegraphics[width=\textwidth]{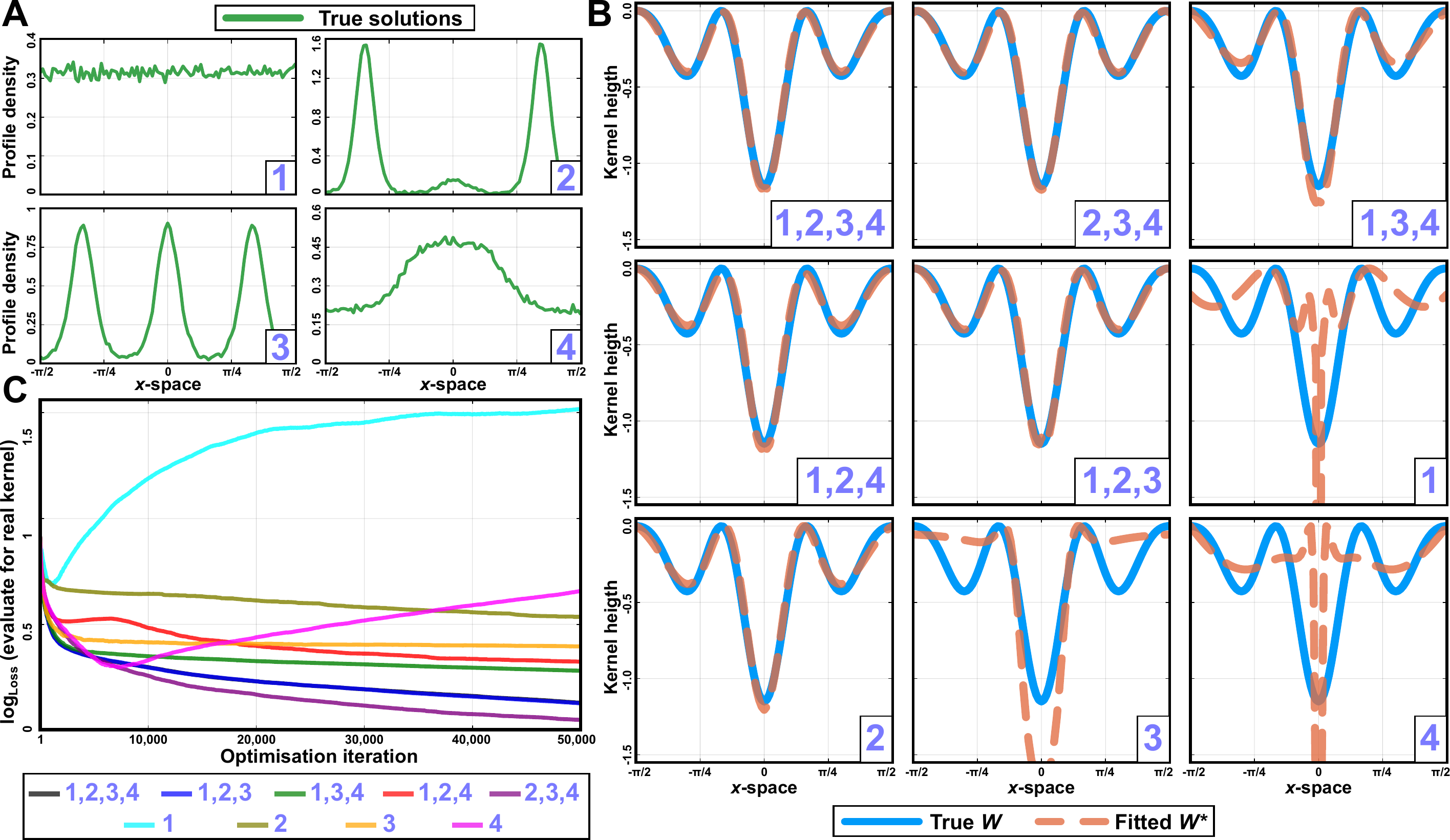}
    \caption{\textbf{Each steady-state solution contains different information.} (A) The four low-noise steady-state solution profiles from Figure~\ref{fig:noisy_ss_fit}B. (B) Recovery of the interaction kernel using different subsets of the solution profiles. The inference procedure is repeated using either all four profiles, all combinations of three profiles, or each profile individually. Recovery fails when only the first or fourth profile is used. The constant first profile is uninformative, since it is a steady-state solution for a broad class of PDEs. By contrast, although the unimodal fourth profile is sufficient for recovery in the noise-free, densely sampled setting (Figure~\ref{fig:fig_1_sup_single_sols}), it no longer contains enough information under sparse sampling and measurement noise. Using only the three-peaked profile recovers some qualitative features of the ground-truth interaction kernel, but not the kernel itself. While using single profiles is not necessarily sufficient for accurate recovery in this setting, combining multiple profiles is. In particular, all combinations containing the second profile yield the most accurate recovered kernels. (C) Convergence of the inferred interaction kernel during optimisation for each subset of solution profiles. At each optimisation step, the current best estimate $W^*$ is compared with the ground-truth kernel $W$. Runs using either the second or third profile converge to the correct kernel, with those including the third but not the second typically converging slightly faster. By contrast, runs that exclude both profiles still converge, but to interaction kernels that differ from the ground truth. An equivalent plot with the optimisation trajectories normalised by runtime is shown in Figure~\ref{fig:fig_3_sup_loss_trajs_normalised}.}
    \label{fig:partial_noisy_ss_fit}
\end{figure}

% ------------------------------------------------------------------------------------------------

Under controlled experimental conditions, it may be possible to observe multiple distinct steady-state solution profiles. In many applications, however, only stable solutions are accessible, and often only a single steady state can be observed. Even when multiple steady states exist, obtaining them experimentally may be costly or time-consuming. It is therefore important to determine how many steady-state profiles are required to recover the interaction kernel from sparse, noisy data. Our analysis of idealised, noise-free data showed that a single steady-state profile can be sufficient for functional recovery (Figure~\ref{fig:fig_1_sup_single_sols}). Whether this remains true in the presence of sparse sampling and measurement noise is less clear. To investigate this question, we repeat the inference procedure using different subsets of the noisy solution profiles shown in Figure~\ref{fig:noisy_ss_fit}B (Figure~\ref{fig:partial_noisy_ss_fit}A). When only a single profile is available, recovery succeeds for some profiles but fails for others (Figure~\ref{fig:partial_noisy_ss_fit}B). By contrast, recovery is successful for every combination of multiple profiles considered, although the convergence rate depends on the particular subset. For example, combining the three least informative profiles still permits recovery, despite each of them performing poorly in isolation. However, this combination converges more slowly than any other three-profile combination (Figure~\ref{fig:partial_noisy_ss_fit}C). These observations suggest that different steady-state profiles contain different amounts of information about the underlying interaction kernel. A natural hypothesis is that this information content is related to the richness of the Fourier spectrum of the observed profile. Both theoretical analysis and numerical experiments support this interpretation (Section~\ref{sec:appendix_analytical_details} and Figures~\ref{fig:fig_3_sup_spectrum_1},\ref{fig:fig_3_sup_spectrum_2}), although a rigorous proof remains an open problem.

% ------------------------------------------------------------------------------------------------

\subsection{All PDE components can be simultaneously recovered from solution profiles}\label{sec:results_WVk_fits}

\begin{figure}[htbp]
    \centering
    \includegraphics[width=0.85\linewidth]{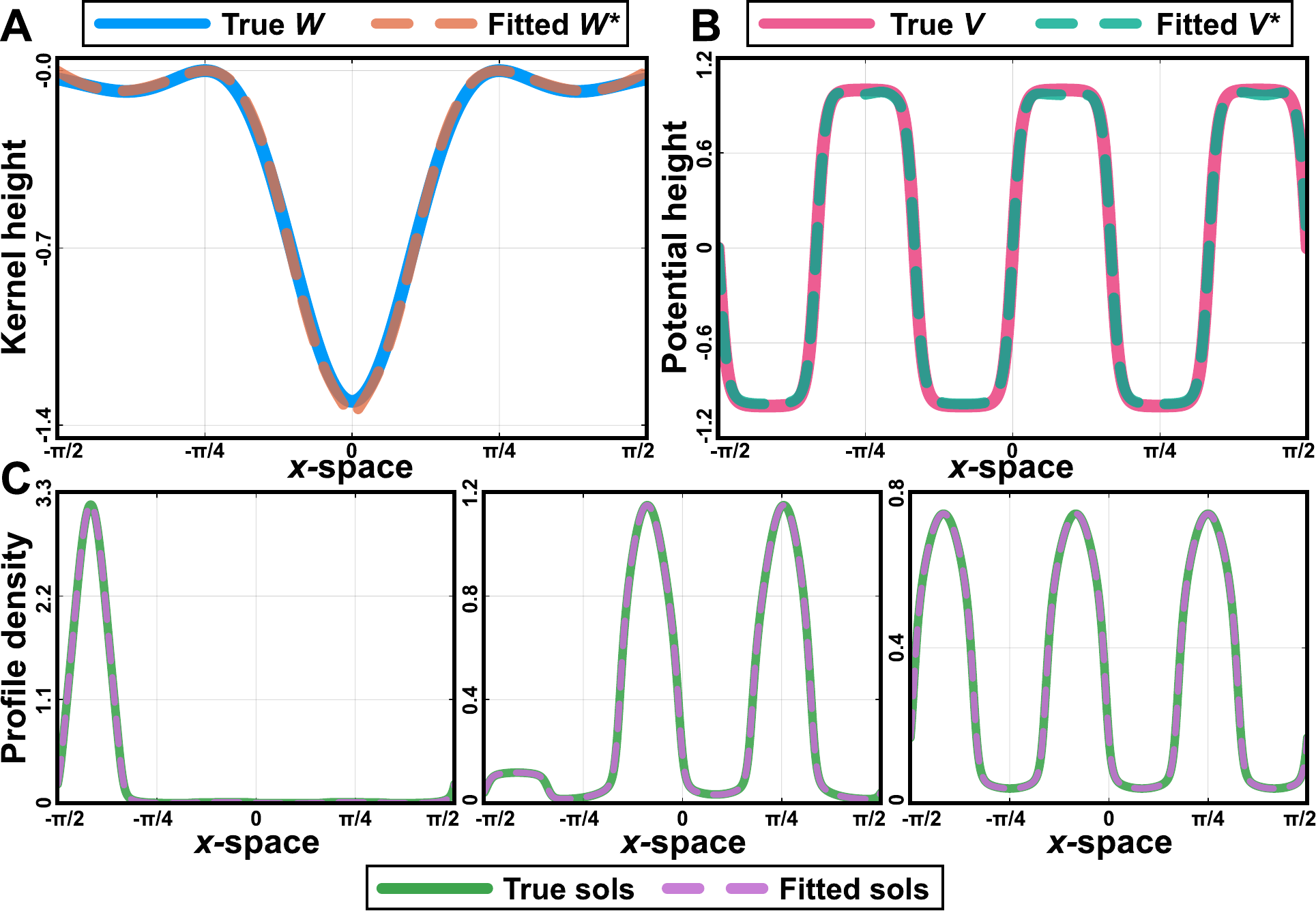}
    \caption{\textbf{The full PDE can be recovered from its solutions.} We set the interaction kernel $W(x)$ (A, solid blue line), potential function $V(x)$ (B, solid pink line), and interaction strength parameter $\kappa = 5.0$---exact functional forms are provided in Supplementary Section~\ref{sup_section:pde_list}. From this PDE, we generate the full set of three steady-state solutions (C, solid green lines). We confirm that the fitted PDE recovers the correct $W$ (A, dashed red line), $V$ (B, dashed teal line) and $\kappa$ (fitted value $4.87$, true value $5.0$). We also confirm that the recovered PDE regenerates the ground-truth solution set (C, dashed purple lines).}
    \label{fig:fit_WVk_to_sols}
\end{figure}

In the previous sections, we assumed that only the interaction kernel $W(x)$ was unknown. In many applications, however, several model components must be inferred simultaneously. We therefore consider the more challenging setting in which the interaction kernel $W(x)$, the external potential $V(x)$, and the interaction strength $\kappa$ are all unknown. We first specify ground-truth functions $W(x)$ and $V(x)$, together with the scalar parameter $\kappa$, and compute the corresponding steady-state solution profiles using Equation~\eqref{eq:model_SS}. We then infer all three model components simultaneously, representing $W(x)$ and $V(x)$ with neural networks while treating $\kappa$ as a trainable scalar parameter. The resulting optimisation accurately recovers all three quantities (Figure~\ref{fig:fit_WVk_to_sols}).

% ------------------------------------------------------------------------------------------------

% ------------------------------------------------------------------------------------------------

As discussed in Section~\ref{sec:identifiability}, simultaneous recovery of multiple model components may be limited by identifiability. In the present setting, however, the availability of multiple steady-state profiles is sufficient to ensure structural identifiability. Consistent with this theoretical prediction, the ensemble analysis demonstrates that the recovered solution is also practically identifiable: optimisation from multiple initialisations consistently yields $(W^*,V^*,\kappa^*)\approx(W,V,\kappa)$ (Figure~\ref{fig:fig_4_sup_practical_identifiability}). To demonstrate the generality of the approach, we repeat the analysis for four additional aggregation--diffusion models, confirming that simultaneous recovery of multiple functional components and scalar parameters is consistently achievable (Figure~\ref{fig:fig_4_sup_additional_nPDEs}).

We next investigate how many steady-state profiles are required for simultaneous recovery of $W$ and $V$. Using the same ground-truth model as in Figure~\ref{fig:fit_WVk_to_sols}, we repeat the inference procedure using either one or two steady-state profiles. Recovery succeeds when two profiles are available, but fails when only a single profile is used (Figure~\ref{fig:fig_4_sup_partial_sols}). This behaviour agrees precisely with Theorem~\ref{thm:identifiability}, which shows that the pair $(W,V)$ is structurally identifiable from two sufficiently distinct steady-state profiles, but structurally non-identifiable from a single profile.

Finally, we consider the setting in which the available steady-state profiles are generated using different values of the interaction strength $\kappa$. In this setting, multiple solution profiles, each associated with a known (and potentially different) value of $\kappa$, are used to infer the interaction kernel $W(x)$ and external potential $V(x)$. We hypothesise that the diversity of the selected solution profiles, rather than simply their number, should determine recovery performance. In particular, two nearly identical solutions obtained from nearby values of $\kappa$ on the same bifurcation branch are expected to provide little additional information beyond that contained in either solution alone. By contrast, solutions that are well separated in $\kappa$, or that lie on different bifurcation branches, should be substantially more informative.

\begin{figure}[!t]
    \centering
    \includegraphics[width=0.85\linewidth]{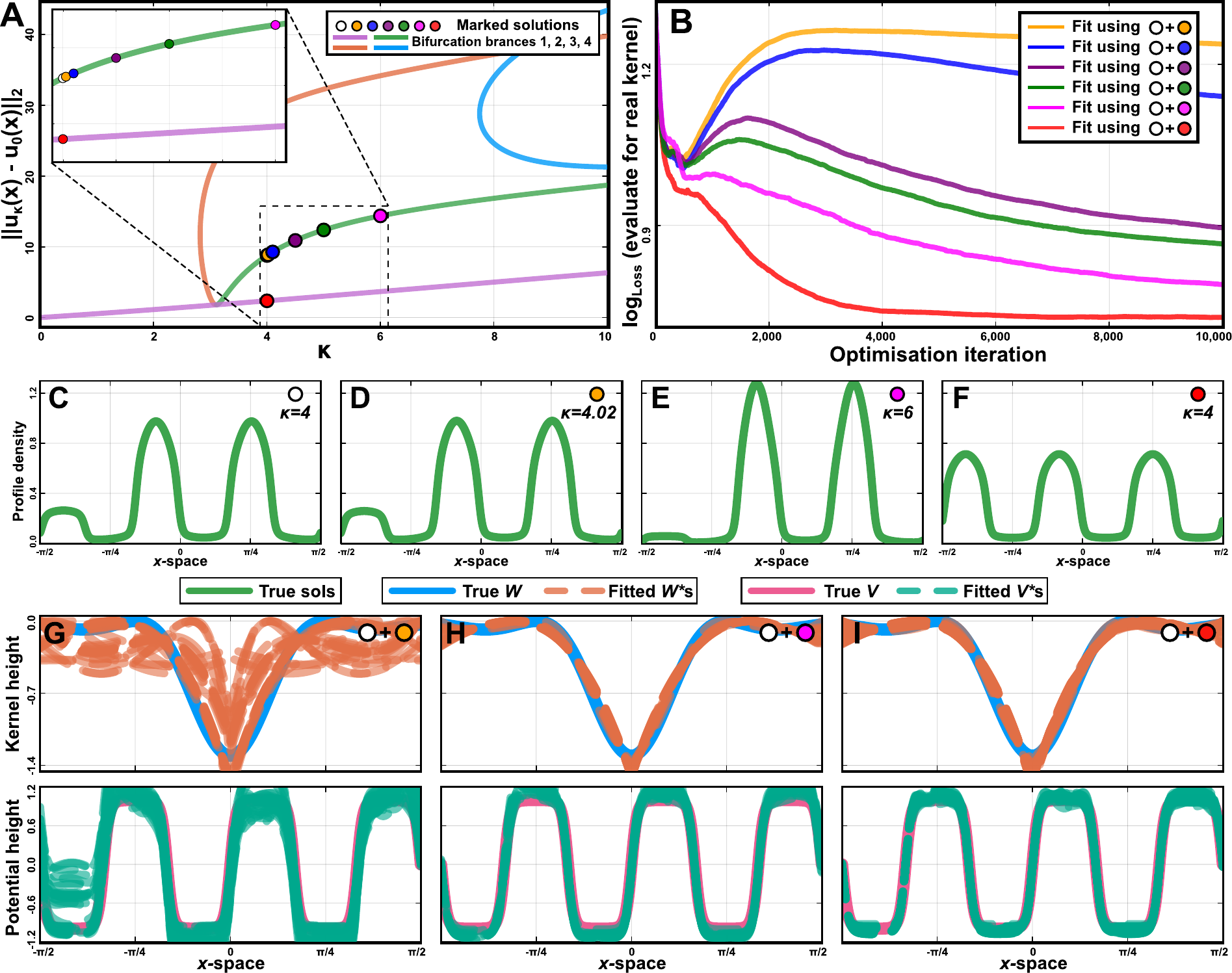}
    \caption{\textbf{Accurate function recovery using solutions from the same bifurcation branch is possible but requires sufficient separation in $\kappa$.} (A) Bifurcation diagram for the aggregation--diffusion model used in Figure~\ref{fig:fit_WVk_to_sols}, showing how the steady-state solution branches vary with the interaction strength $\kappa$. The vertical axis shows $\int (u(x)-1/\pi)^2\dx$. Dots indicate the solution profiles used for inference (profiles are shown in panels C--F and Figure~\ref{fig:fig_5_sols}). (B) Recovery accuracy when inferring $W(x)$ and $V(x)$ from pairs of steady-state solution profiles. A fixed reference solution (white dot in panel A) is paired with each coloured solution, and the mean reconstruction error, $\int (W^*(x)-W(x))^2\dx+\int (V^*(x)-V(x))^2\dx$, averaged over 400 independent optimisation runs, is plotted throughout the optimisation. For pairs of solutions on the same bifurcation branch, recovery improves as their separation in $\kappa$ increases. By contrast, solutions drawn from different branches permit accurate recovery even when they correspond to the same value of $\kappa$. (C--F) Representative steady-state solution profiles corresponding to the highlighted points in panel A (identified by the coloured markers in the upper-right corners). (G--I) Ensemble plots of the recovered interaction kernels $W^*(x)$ and external potentials $V^*(x)$. Panels G and H use pairs of solutions from the same bifurcation branch, corresponding to the minimum and maximum $\kappa$ separations, respectively, while panel I uses solutions from different branches. Practical non-identifiability is observed in panel G, whereas accurate recovery is achieved in panels H and I. Ensemble plots for the remaining solution pairs are provided in Figure~\ref{fig:fig_5_dual_fits}.}
    \label{fig:bif_analysis}
\end{figure}

The results shown in Figure~\ref{fig:bif_analysis} support this hypothesis. Recovery using pairs of steady-state profiles on the same bifurcation branch is unsuccessful when the corresponding values of $\kappa$ are close, but improves steadily as their separation in $\kappa$ increases. Indeed, sufficiently separated solutions from the same branch achieve recovery performance comparable to that obtained using solutions from different branches. These results demonstrate that the information gained from an additional steady-state profile depends on the novel information it provides beyond that already contained in the available observations. More broadly, they highlight that careful selection of complementary steady-state observations can substantially improve functional recovery.

% ------------------------------------------------------------------------------------------------

% ------------------------------------------------------------------------------------------------

\pagebreak

\section{Discussion}

In this work, we have shown that unknown functional components of PDE models can be recovered directly from data by embedding neural networks within a mechanistic modelling framework. Using nonlocal aggregation--diffusion equations as a case study, we demonstrated that interaction kernels, external potentials, and scalar parameters can be inferred simultaneously using standard optimisation techniques while preserving the underlying PDE structure. Once inferred, the resulting UPDE can be analysed and simulated in the same way as a conventional PDE model, providing a practical route from incomplete mechanistic models to predictive mathematical descriptions.

A central conclusion of this work is that successful functional inference depends not only on the expressive power of the neural network or the optimisation algorithm, but fundamentally on the information contained in the available observations. Through a combination of theoretical analysis and numerical experiments, we identified the principal factors governing recoverability: the structure of the governing PDE, the number and diversity of observed steady states, the spatial resolution and noise level of the data, and the intrinsic properties of the unknown functions. Together, these factors determine both structural identifiability, which characterises whether recovery is theoretically possible, and practical identifiability, which determines whether it can be achieved reliably in the presence of realistic data limitations.

An important insight emerging from this work is that functional inference admits several qualitatively distinct outcomes, which should not be conflated. The ideal case is one in which the unknown function is recovered accurately and consequently reproduces the correct solution landscape. However, successful functional recovery is neither necessary nor sufficient for accurate prediction. We observed examples in which a recovered function closely matched the ground truth yet numerical solution of the UPDE failed to reproduce the full set of steady-state solutions, illustrating that seemingly minor discrepancies in functional form can alter qualitative PDE behaviour. Conversely, an imperfectly recovered function may nevertheless generate accurate solution-level predictions over the range of observations considered. Finally, inference may fail either because the optimisation procedure converges to an incorrect local minimum or because the inverse problem is structurally non-identifiable, in which case multiple distinct model parameterisations give rise to the same observations. These failure modes arise for fundamentally different reasons and therefore require different remedies: optimisation failures motivate improved numerical algorithms, whereas structural non-identifiability can only be overcome through the addition of more informative observations, stronger prior assumptions, or reformulation of the inverse problem itself.

Our results also have direct implications for experimental design. The informational content of steady-state observations is highly non-uniform: some equilibria contain sufficient information for accurate recovery, whereas others remain largely uninformative, even under idealised conditions. Moreover, the information gained from an additional observation depends on how much novel information it contributes beyond that already contained in the available data. Pairs of sufficiently distinct steady states can resolve ambiguities that cannot be overcome by repeated observations of similar equilibria, and observations from different regions of the solution landscape often provide substantially more complementary information than nearby states on the same branch. These findings suggest that experimental design should prioritise informative and complementary observations rather than simply increasing the number of measurements. More broadly, they reinforce the conclusion that recoverability is governed by the interplay between model structure and observational data, rather than by either in isolation.

Although our primary focus has been on methodological and theoretical aspects of functional inference, our approach and results have broad practical implications. In many applications, qualitative knowledge of governing mechanisms is available, while the associated functional forms remain unknown. Rather than prescribing these functions heuristically, the UPDE framework provides a systematic means of recovering them directly from observations while retaining mechanistic interpretability. Once inferred, the resulting model can be used for forward simulation, scenario analysis, or quantitative assessment of interventions, thereby closing the loop between mechanistic modelling, data assimilation, and prediction.

Several important questions remain open. Throughout this work, we deliberately restricted attention to inference from steady-state data. Time-dependent observations generally contain substantially richer information and may resolve ambiguities that are unavoidable when only equilibria are available. Extending the present framework to incorporate transient dynamics therefore represents an important direction for future research. Likewise, incorporating prior information about unknown functions---for example symmetry, positivity, monotonicity, or smoothness---through constrained neural network architectures or Bayesian priors may substantially improve robustness in challenging inference problems. Finally, although our numerical experiments indicate that combining the fixed-point residual with neural networks provides particularly robust performance, a more complete theoretical and computational comparison of alternative loss functions and universal function approximators remains an important topic for future investigation.

Although our numerical examples focused on a class of aggregation--diffusion equations, the underlying framework is considerably more general. Unknown functional components arise throughout mathematical biology, physics, engineering, finance, and many other disciplines in the form of spatially varying coefficients, interaction kernels, constitutive laws, and nonlinear response functions. More broadly, our results establish that functional inference for PDE models is both theoretically tractable and computationally practical, while clarifying the conditions under which recovery succeeds and, equally importantly, when it cannot. We hope that this perspective will prove valuable not only for empirical applications, but also for future work on inverse problems, identifiability, and the mathematical analysis of complex PDE systems.

% ------------------------------------------------------------------------------------------------

\section*{Acknowledgements}

TEL and REB are supported by a grant from the Simons Foundation (MP-SIP-00001828). JAC was partially supported by the Advanced Grant Nonlocal-CPD (Nonlocal PDEs for Complex Particle Dynamics: Phase Transitions, Patterns and Synchronization) of the European Research Council Executive Agency (ERC) under the European Union Horizon 2020 research and innovation programme (grant agreement No.~883363). JAC acknowledges support by the EPSRC grant EP/V051121/1. JAC was also partially supported by the “Maria de Maeztu” Excellence Unit IMAG, reference CEX2020-001105-M, funded by MCIN/AEI/10.13039/501100011033/. YS was partially supported by the Natural Sciences and Engineering Research Council of Canada through an NSERC Postdoctoral Fellowship (PDF-578181-2023), by the Atlantic Association for Research in the Mathematical Sciences through an AARMS Postdoctoral Fellowship, and by The Leverhulme Trust through a Leverhulme Early Career Fellowship (ECF-2025-084). For the purpose of open access, the authors have applied a CC BY public copyright licence to any author-accepted manuscript arising from this submission.

% ------------------------------------------------------------------------------------------------

\section*{Code availability}

All scripts for regenerating all figures (as well as all other numerical analyses) can be found here: \url{https://github.com/TorkelE/NonlocalPDEKernelFitting_2026}.

% ------------------------------------------------------------------------------------------------

\bibliographystyle{abbrv}
\bibliography{references} 

\pagebreak

% ------------------------------------------------------------------------------------------------

\appendix
%\pagebreak
\clearpage
% Reset counters
\setcounter{page}{1}
\setcounter{section}{0}
\setcounter{equation}{0}
\setcounter{figure}{0}
\setcounter{table}{0}
% Page numbering
\renewcommand{\thepage}{S\arabic{page}}
% Section numbering
\renewcommand{\thesection}{S\arabic{section}}
% Equation numbering
\renewcommand{\theequation}{S\arabic{equation}}
\renewcommand{\thefigure}{S\arabic{figure}}
\renewcommand{\thetable}{S\arabic{section}.\arabic{table}}
% Make equations follow section numbering if desired
\numberwithin{equation}{section}
% Remove "Appendix" prefix from section headings
\renewcommand{\appendixname}{}
\makeatletter
\renewcommand{\@seccntformat}[1]{\csname the#1\endcsname.\quad}
\makeatother
% Title
\begin{center}
    {\Large\bfseries Learning functional components of partial differential equation models from data using neural networks - Supplementary Information}\\
\date{}
\end{center}
% ------------------------------------------------------------------------------------------------

% ------------------------------------------------------------------------------------------------

\section{Analytical properties of aggregation--diffusion equations} \label{sec:appendix_analytical_details}

In this section, we compile some known properties of the aggregation--diffusion equation, Equation~\eqref{eq:model}, and its stationary counterpart, Equation~\eqref{eq:model_SS}. We state all results for the one-dimensional case only to avoid additional technicalities; however, we note that many aspects of the analyses carry over to the higher-dimensional cases under appropriate modification. 

Consider the domain $\mathbb{T} = (-L/2, L/2]$ for some $L>0$ fixed. We denote by $L^2(\mathbb{T})$ the space of Lebesgue measurable functions with finite $L^2$-norm:
\begin{align*}
    L^2(\mathbb{T}) := \left\{ f \text{ (Lebesgue) measurable}: \int_\mathbb{T} \as{f}^2 \dx < \infty \right\}.
\end{align*}
We then denote by $L^2_{\textup{per}} (\mathbb{T}) \subset L^2(\mathbb{T})$ the (closed) subspace of periodic functions belonging to $L^2(\mathbb{T})$, and by $L^2_s(\mathbb{T})$ the closed subspace of $L^2_{\textup{per}}(\mathbb{T})$ consisting of even functions. We denote by $H^k(\mathbb{T})$, $k \in \mathbb{N}$, those functions having weak derivatives up to and including order $k$, each of which belongs to $L^2(\mathbb{T})$. Finally, we denote by $\{ w_k \}_{k \geq 0}$ the orthonormal basis for $L^2_s(\mathbb{T})$ given by
\begin{align}\label{eq:ONB_L2s}
    w_k(x) := \begin{cases}
        \frac{1}{\sqrt{L}}, \qquad\qquad\qquad\quad k = 0; \cr 
        \sqrt{\tfrac{2}{L}} \cos \left(2 \pi k x / L  \right), \quad k \geq 1.
    \end{cases}
\end{align}
For simplicity, we will assume that $W$ is even in addition to $W,\, V \in H^2(\mathbb{T}) \cap L^2_{\textup{per}} (\mathbb{T})$. 
We note that these conditions can be weakened significantly, see, e.g.,~\cite{carrillo2024wellposedness,carrillo2025longtimebehaviourbifurcationanalysis}.

% ------------------------------------------------------------------------------------------------

\subsection{Well-posedness} 

We begin with the well-posedness (existence and uniqueness) of the time-dependent problem given in Equation~\eqref{eq:model}, its stationary counterpart in Equation~\eqref{eq:model_SS} (existence and smoothness), as well as the positivity and mass-preserving property of solutions. This follows from the proof of well-posedness found in, e.g.,~\cite{chazelle2017wellposedness,Carrillo2020LongTime}. Note that minor modification is required to accommodate $V$, but assuming sufficient regularity (e.g., $H^2$) leaves the core approach unchanged.

\begin{proposition}\label{prop:wellposed}
    Suppose $W, V \in H^2(\mathbb{T})$. Then, for initial data $u_0 \in H^{4} (\mathbb{T})$, there exists a unique classical solution solving Equation~\eqref{eq:model} such that $u(x,\cdot) \in C^2(\mathbb{T})$ for all $t \geq 0$. Moreover, $u(\cdot,t)$ is strictly positive for all $t>0$ and satisfies
    \begin{align}
        \int_\mathbb{T} u(y,t) \dy = \int_\mathbb{T} u_0 (y) \dy \quad \forall t \geq 0,
    \end{align}
    i.e., the mass is preserved for all $t>0$.
\end{proposition}
We also note the following existence, smoothness, and positivity result for the stationary problem whose statement for $V \equiv 0$ can be found in, e.g.,~\cite{Carrillo2020LongTime}. The same approach applies for nontrivial $V$. 

\begin{proposition}\label{prop:wellposed_stationary}
    Suppose $W,V \in H^2(\mathbb{T})$. Then, there exists a weak solution $u \in H^1(\mathbb{T})$ such that $\int_\mathbb{T} u \, \dx = 1$ solving the stationary problem given in Equation~\eqref{eq:model_SS}. Moreover, any such weak solution is smooth and strictly positive in $\mathbb{T}$, i.e., $u \in H^k(\mathbb{T})$ for any $k \geq 1$ and $u > 0$ everywhere in $\mathbb{T}$.
\end{proposition}

Since the proof of this result uses an equivalence between steady-state solutions and fixed points of a nonlinear map $\mathcal{T}$, we forego the details until after the map $\mathcal{T}$ has been introduced (see Proposition \ref{prop:fixedpoint_iff_SS}).

% ------------------------------------------------------------------------------------------------

\subsection{The aggregation--diffusion model as a gradient flow}

To better understand the energy structure of the model and the influence on its steady states, it is useful to rewrite Equation~\eqref{eq:model} as a gradient-flow:
\begin{align}\label{eq:model_onedimension_gradflow}
  \partial_t u = \partial_x\, \Big(u\,\partial_x\big( \log u + \kappa W * u +  V\big)\Big)
  = \partial_x \Big(u\,\partial_x \mathcal{F}^\prime [u]\Big),
\end{align}
where $\cdot^\prime$ denotes the first variation of the free energy functional:
\begin{align}\label{eq:energy}
  \mathcal{F}[u]
  := \int_{\mathbb{T}} u \left(  \log u - 1 + \tfrac{1}{2} \kappa W * u +  V\right)\dx .
\end{align}
In this sense, the model is a continuity equation of the form $\partial_t u + \partial_x J = 0$ with flux $J := - u \partial_x \mathcal{F}^\prime [u]$. Assuming that the initial data is a probability density function, from Proposition \ref{prop:wellposed} we have that $\norm{u(\cdot,t)}_{L^1(\mathbb{T})} = 1$ for all $t>0$. A formal computation that can be made rigorous then yields 
\begin{align}
  \mathcal{F}^\prime [u] := \lim_{\varepsilon \to 0} \frac{\mathcal{F}[u + \varepsilon \eta] - \mathcal{F}[ u]}{\varepsilon}
  = \log u - 1 + \kappa W\ast u + V,
\end{align}
where $\eta$ is a mean-zero variation. Hence, we multiply Equation~\eqref{eq:model_onedimension_gradflow} by $\mathcal{F}^\prime[u]$ and integrate by parts to conclude that along smooth trajectories of Equation~\eqref{eq:model_onedimension_gradflow} there holds
\begin{align}\label{eq:Ediss}
  \frac{\rm{d}}{{\rm d} t}\mathcal{F}[u(t)]
  + \int_{\mathbb{T}} u\,\bigl|\partial_x\bigl( \log u + \kappa W\ast u +  V\bigr)\bigr|^2 \dx
  = 0.
\end{align}
In particular, ${\rm d} \mathcal{F} / {\rm d} t \leq 0$ and so we conclude that $\mathcal{F}[u]$ is decreasing along solution trajectories for all time. Moreover, since we are on a bounded domain, it can be shown that the free energy $\mathcal{F}(u)$ is bounded below. Consequently, persistent oscillatory behaviour cannot occur in the long-time limit, and only those solutions that remain constant over time are possible in this long-time limit. It is therefore natural to study steady states of Equation~\eqref{eq:model_onedimension_gradflow} through critical points of the free energy functional $\mathcal{F}$, or equivalently, through fixed points of a nonlinear map $\mathcal{T}$, which we identify now.

% ------------------------------------------------------------------------------------------------

\subsection{Steady states via a fixed-point map} 

With the only possibility in the long-time limit being $\partial_t u = 0$, stationary states $u = u_\text{ss}(x)$ satisfy
\begin{align}\label{eq:stat}
0 = \partial_x\Big(u_\text{ss}\,\partial_x\big( \log u_\text{ss} + \kappa W \ast u_\text{ss} +  V\big)\Big) , \quad x \in \mathbb{T}.
\end{align}
To ensure this problem is well-defined, we prescribe the mass condition $\int_\mathbb{T} u_\text{ss}(x) \dx = 1$ so that we seek steady states that are probability density functions. This is natural when the time-dependent problem is equipped with initial data $u_0(\cdot)$ which is itself a probability density function.

Similar to how we obtained Equation~\eqref{eq:Ediss} in the time-dependent case, we multiply Equation~\eqref{eq:stat} by $ \log u_\text{ss} + \kappa W \ast u_\text{ss} + V$, integrate by parts, and apply the periodic boundary conditions to obtain
$$
\int_\mathbb{T} u_\text{ss} \magg{\partial_x ( \log u_\text{ss} + \kappa  W \ast u_\text{ss} +  V)}\, \dx = 0.
$$
By Proposition~\ref{prop:wellposed_stationary}, any steady state is smooth and satisfies $u_\text{ss} > 0$ in $\mathbb{T}$, and so there must hold
$$
\partial_x ( \log u_\text{ss} +  \kappa W \ast u_\text{ss} +  V) = 0 \quad \text{ everywhere in } \mathbb{T},
$$
and so $ \log u_\text{ss} +  \kappa W \ast u_\text{ss} +  V = c_0$ over $\mathbb{T}$, for some $c_0 \in \mathbb{R}$ to be determined. In the case of linear diffusion, as chosen in this work, we may rearrange and solve directly for this constant to deduce that $u_\text{ss}$ must satisfy
\begin{align}
    u(x) &=  \dfrac{\exp \left( - [ \kappa W \ast u(x) +  V(x)]  \right)}{Z(u)},\label{eq:FP_1.1} \\
    Z(u) &:= \int_\mathbb{T}\exp \left( -[ \kappa W \ast u(y) + V(y)]  \right) \dy. \label{eq:FP_1.2}
\end{align}
This motivates the fixed-point formulation via the map $\mathcal{T} : L^2(\mathbb{T}) \mapsto L^2(\mathbb{T})$ defined by
\begin{align}\label{eq:Tmap}
  \mathcal{T}u := \frac{\exp\big(-[\kappa W \ast u + V]\big)}{Z(u)},
\end{align}
with $Z = Z(u)$ as defined above. Notice that Equations~\eqref{eq:FP_1.1}-\eqref{eq:FP_1.2} are equivalent to $\mathcal T u = u$. On the other hand, if a function $u>0$ satisfies $\mathcal{T}u = u$, direct computation shows that $u$ necessarily satisfies Equation~\eqref{eq:stat} with $\int_{\mathbb{T}} u\, \dx = 1$. The arguments above provide the following proposition.

\begin{proposition}\label{prop:fixedpoint_iff_SS}
    A function $u \in H^1(\mathbb{T})$ is a nonnegative weak solution to the problem given in Equation~\eqref{eq:stat} if and only if it is a fixed point of the nonlinear map $\mathcal{T}$ given by Equation~\eqref{eq:Tmap}. 
\end{proposition}
The proof of Propositions~\ref{prop:wellposed_stationary}-\ref{prop:fixedpoint_iff_SS} when $V \equiv 0$ can be found in, e.g.,~\cite{Carrillo2020LongTime}; in that case, they use a fixed-point argument when $W \in H^1(\mathbb T)$, which yields sufficient compactness properties. Assuming sufficient regularity on $V$, e.g., $V \in H^2(\mathbb{T})$, an identical approach applies when nontrivial $V$ is included. We omit further details.

In addition to the Lipschitz continuity of $\mathcal{T}$, in \cite{carrillo2025longtimebehaviourbifurcationanalysis} it is shown that when $V\equiv0$, the map $\mathcal{T}$ is Fr{\'e}chet differentiable, with Lipschitz continuous derivative. Forgoing the technical details, the map $\mathcal{T}$ possesses the necessary ingredients to apply an efficient Newton-Krylov-type algorithm to identify its fixed points. We refer to \cite{carrillosalmaniwvillares2025} for full details.

% ------------------------------------------------------------------------------------------------

\subsection{Analytical description of steady states} 

When at least one of $W(x)$ or $V(x)$ is identically zero over $\mathbb{T}$, the solution structure at steady state can be understood analytically for some special cases. To demonstrate this in a logically sensible way, we consider these two cases as guidance for future numerical explorations: first, when $W(x) \equiv 0$ so that there is no nonlocal self-interaction; second, when $V(x) \equiv \text{constant}$ so that there is no influence from the external potential.

\subsection*{Case I: $W(x) \equiv 0$} In this case, there is no self-interaction through the nonlocal kernel $W$, and the problem reduces to a linear advection--diffusion equation. Given any $V \in H^2(\mathbb{T}) \cap L^2_{\textup{per}} (\mathbb{T})$, the unique steady state is given explicitly by
\begin{align}
    u_V(x) = \frac{\exp ( - V )}{\int_\mathbb{T} \exp (- V) \dx}.
\end{align}
Notice that when $V(x) \equiv \text{constant}$, the unique steady state is precisely $L^{-1}$; otherwise, the unique steady state is always nontrivial. 

\subsection*{Case II: $V(x) \equiv \text{constant}$} When $V$ is constant, the external potential does not influence the dynamics; on the other hand, so long as the kernel $W * \cdot$ is nontrivial, the problem becomes nonlinear. This has several implications for the possible solution structure of Equation~\eqref{eq:stat}. First, the problem becomes translationally invariant in the sense that for any steady-state solution $u(x)$, $u(x + \xi)$, $\xi \in \mathbb{R}$, also as a steady-state solution. Furthermore, any constant profile $u_\infty \equiv c$ is a valid steady-state solution to Equation~\eqref{eq:stat}; under the unit mass constraint, this implies that there is always a unique homogeneous (also called the \textit{trivial}) steady-state solution given by $u_\infty = L^{-1}$, a valid solution for any $\kappa \geq 0$. 

Fixing the trivial state $u_\infty = L^{-1}$ allows one to study in detail the associated bifurcation structure of Equation~\eqref{eq:stat} in the absence of an external potential $V(x)$. To this end, we treat $\kappa \geq 0$ as a bifurcation parameter. Due to translational invariance, we can then restrict the analysis to $L^2_s(\mathbb{T}) \subset L^2_{\textup{per}}(\mathbb{T})$, the (closed) subspace of even, $L$-periodic functions. We seek solutions that bifurcate from the homogeneous solution $u_\infty = L^{-1}$ in terms of the strength parameter $\kappa$. The key intuition, up to some technical criteria, is the following result (see \cite[Theorem 1.2]{Carrillo2020LongTime}): \textit{any Fourier mode $k \in \mathbb{N} \setminus \{ 0 \}$ such that $\widetilde W(k) < 0$ leads to a bifurcation point $(\kappa^*_k, L^{-1})$ from the homogeneous state taking the following form near $\kappa = \kappa_k^*$:
$$
u_{k,\mathrm{ss}}(x) = L^{-1} + s \cos ( 2 \pi k x / L ) + o(s), \quad \as{s} \ll 1 , \quad \kappa_k^* := -\frac{ \sqrt{2L}}{\widetilde W(k)} > 0.
$$
Moreover, any Fourier mode such that $\widetilde W(k) \geq 0$ does not lead to a bifurcation point.} 

This alone allows one to quickly identify which kernels will lead to nontrivial steady states for our study; cases without nontrivial steady states leave little work to be done. More importantly, since the bifurcation value $\kappa^*_k$ has an analytical expression, we know not only a minimal number of nontrivial solutions to expect\footnote{The minimal number of nontrivial states to expect is one nontrivial solution for each wavenumber such that $\widetilde W(k) < 0$. In practice, we observe several additional higher-order bifurcations for every such wavenumber, and these cannot be explored through analytical theory alone.}, but also where to search for them in $\kappa$-space. Furthermore, \cite{carrillo2025longtimebehaviourbifurcationanalysis} has recently expanded upon this bifurcation analysis to determine the branch direction and stability properties of the emergent branch, depending on whether it is sub- or supercritical.

From Cases \textbf{I}-\textbf{II} described above, we can gain some understanding of the expected solution behaviour for combinations of $\kappa W$ and $V$ by (heuristically) interpolating between these two edge cases.

% ------------------------------------------------------------------------------------------------

\section{Detailed methodology}\label{sup_section:detailed_methods}

% ------------------------------------------------------------------------------------------------

\subsection{Fourier mode expansion as universal function approximator}\label{sup_section:fouier_expansion_ufa}

\begin{figure}[htbp]
    \centering
    \includegraphics[width=\textwidth]{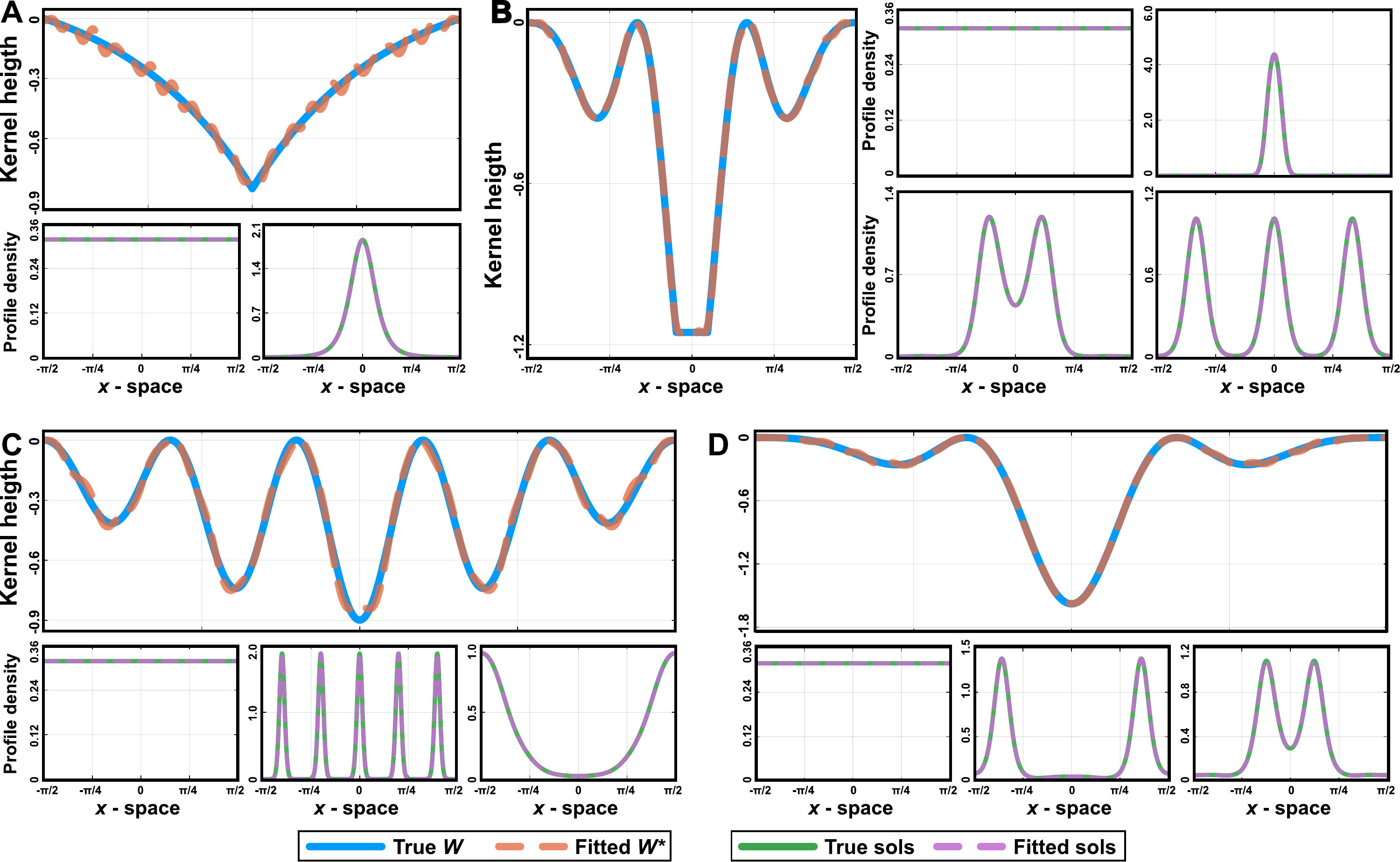}
    \caption{\textbf{A Fourier mode expansion can be used as a universal function approximator.} Throughout this paper, we rely on neural network function approximators to find the functional parameters $V$ and $W$. However, other approaches are possible. Here, for four different $W$ (in all cases, $V\equiv0$), we perform the workflow described in Figure~\ref{fig:workflow_flowchart} to recover $W$ from the solution profiles. Instead of using neural networks to approximate the unknown functions, we use Fourier mode expansion (detailed in Section~\ref{sup_section:fouier_expansion_ufa}). (A-D) In all four cases, the Fourier mode expansion can successfully recover $W$ from the steady state solutions, and the fitted PDE yields the ground-truth solution profiles.}
    \label{fig:fig_1_sup_fouier_ufa}
\end{figure}

\begin{figure}[htbp]
    \centering
    \includegraphics[width=0.86\linewidth]{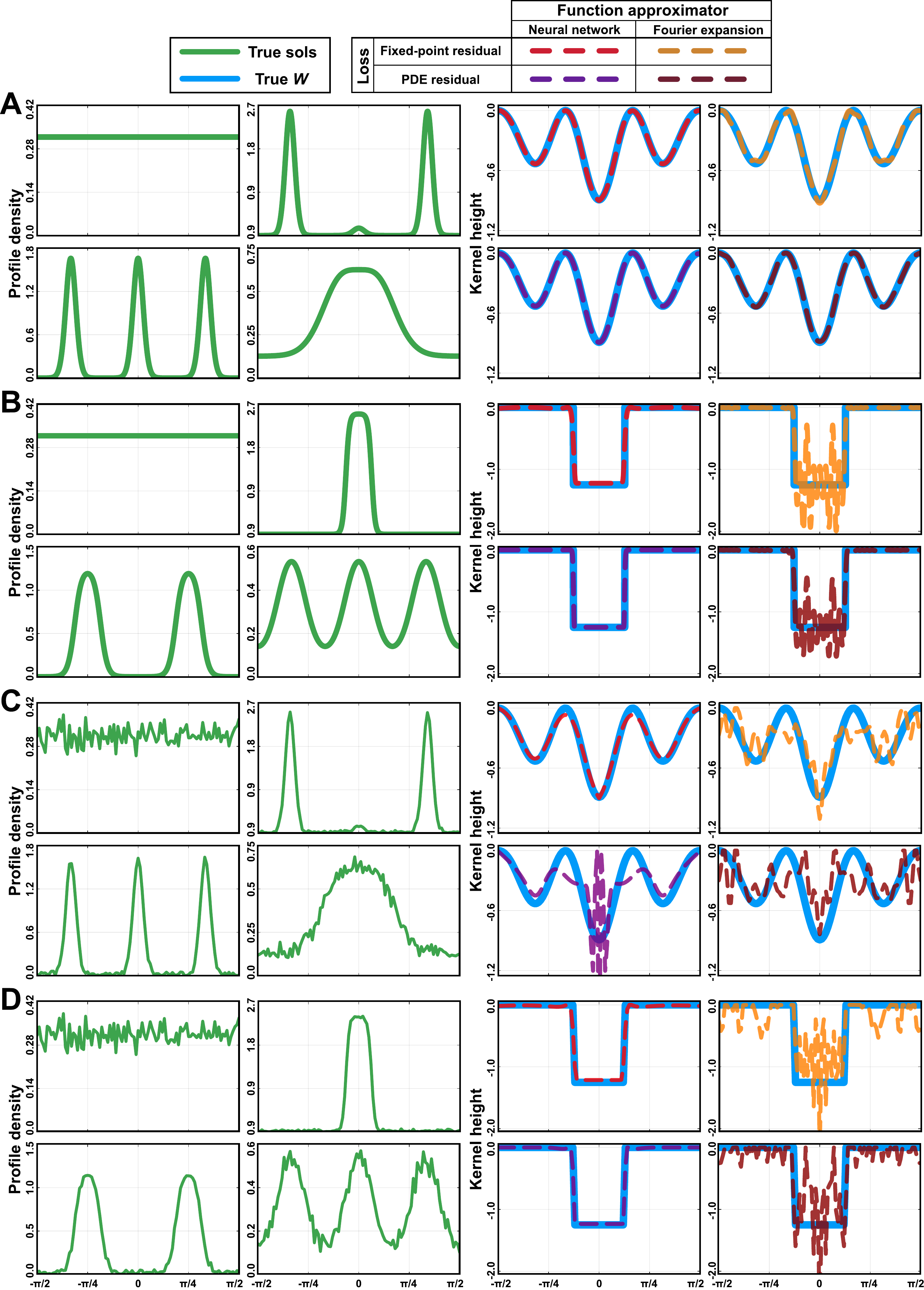}
    \caption{\textbf{Combining a neural network function approximator with the fixed-point residual loss function offers the most robust inference approach.} In Section~\ref{section:methods_udes}, we introduce two loss functions (the PDE residual in Equation~\eqref{eq:loss_function_PDE} and the fixed-point residual in Equation~\eqref{eq:loss_function_FP}) and two function approximators (neural networks and Fourier mode expansions). Here, we briefly evaluate their abilities to recover functions from PDEs with either a smooth kernel $W$ and no noise (A), a piecewise linear kernel $W$ and no noise (B), a smooth kernel $W$ and noise (C), or a piecewise linear kernel $W$ and noise (D). For each case, we show the four PDE solution profiles (green lines) on the left and the four approaches' recovered kernels $W^*$ (coloured dashed line) compared to the true kernels (solid blue lines) on the right. Across the four fitting approaches, we use the same number of parameters in the function approximator (ranging from $15$ to $40$), and run the optimiser for the same number of iterations. We note that only combining the fixed-point residual loss function with the neural network function approximator robustly recovers the kernels (red dashed lines) across all cases. How generally this holds across different PDEs and measurements, and the extent to which the chosen fitting approach affects convergence speed, remains unanswered.}
    \label{fig:fig_1_sup_fitting_approaches}
\end{figure}

In Figures~\ref{fig:fig_1_sup_fouier_ufa} and~\ref{fig:fig_1_sup_fitting_approaches} we use Fourier mode expansions, rather than neural networks, to approximate the unknown functional parameter $W$. Specifically, we use a truncated cosine expansion
\begin{equation} %\label{eq:fouier_ufa}
  W^*(x; \vec{\theta}_W) = a_0 + \sum_{k=1}^K a_k \cos\left(\frac{2\pi kx}{L}\right),
\end{equation}
with $\vec{\theta}_W = (a_0,a_1,\dots,a_K)$. This directly imposes the constraint of symmetry, and the constraints $W\leq0$, $\max_{x} \{ W \} = 0$ are imposed in a similar manner as for the neural network-based approach (Section~\ref{sup_section:detailed_methods_nn_arch}). This yields the following parameterised PDE:
\begin{equation} %\label{eq:fouier_ufa_pde}
  0
  = \partial_x ^2 u
  + \kappa \partial_x ( u \partial_x (W^*(x; \vec{\theta}_W) \ast u) ) + \partial_x ( u \partial_x V ),
\end{equation}
where $V$ is assumed to be known. A similar approach can be used to approximate $V$, now using a truncated Fourier series:
\begin{equation} %\label{eq:fouier_ufa}
  V^*(x; \vec{\theta}_V) = a_0 + \sum_{k=1}^K \left[a_k \cos\left(\frac{2\pi kx}{L}\right) + b_k \sin\left(\frac{2\pi kx}{L}\right)\right],
\end{equation}
with $\vec{\theta}_V = (a_0,a_1,\dots,a_K,b_1,\dots,b_K)$. Different from $W^*$, the approximation $V^*$ must also include sine terms as it is assumed periodic, but not necessarily even.

In practice, when recovering $W$ we use a similar number of Fourier mode terms as the number of parameters in our fitted neural networks (numbers vary between problems, but are all approximately between $15$ and $40$). In Figures~\ref{fig:fig_1_sup_fitting_approaches}, when we apply the two approaches to the same PDE, we ensure that these numbers are identical so that sensible comparisons can be made.

% ------------------------------------------------------------------------------------------------

\subsection{Neural network architecture}\label{sup_section:detailed_methods_nn_arch}

\begin{figure}[htbp]
    \centering
    \includegraphics[width=\textwidth]{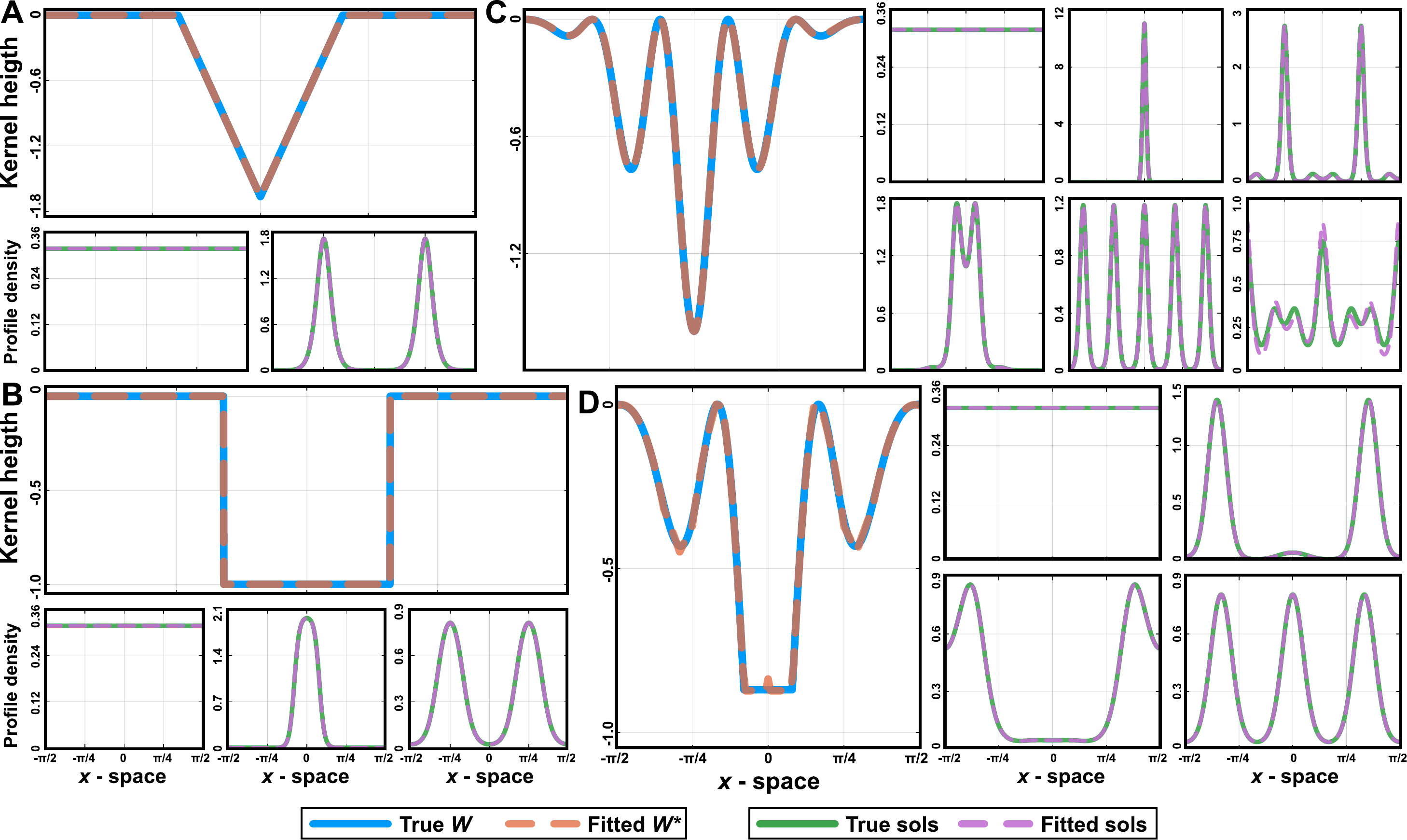}
    \caption{\textbf{A wide range of potential $W$ can be recovered from the solutions.} (A-D) To confirm the general nature of our results, we repeat them for a range of interaction kernels $W$. In each case, we confirm that the fitted kernel $W^*$ (dashed red lines) follows the true kernels (solid blue lines) and that the solutions generated by the fitted kernels (dashed purple lines) are also solutions of the ground-truth kernels (solid green lines). The kernels include piecewise linear ones (A and B), a continuous one with multiple modes (C) and one with both piecewise linear and continuous elements (D).}
    \label{fig:fig_1_sup_alt_Ws}
\end{figure}

\begin{figure}[htbp]
    \centering
    \includegraphics[width=0.85\linewidth]{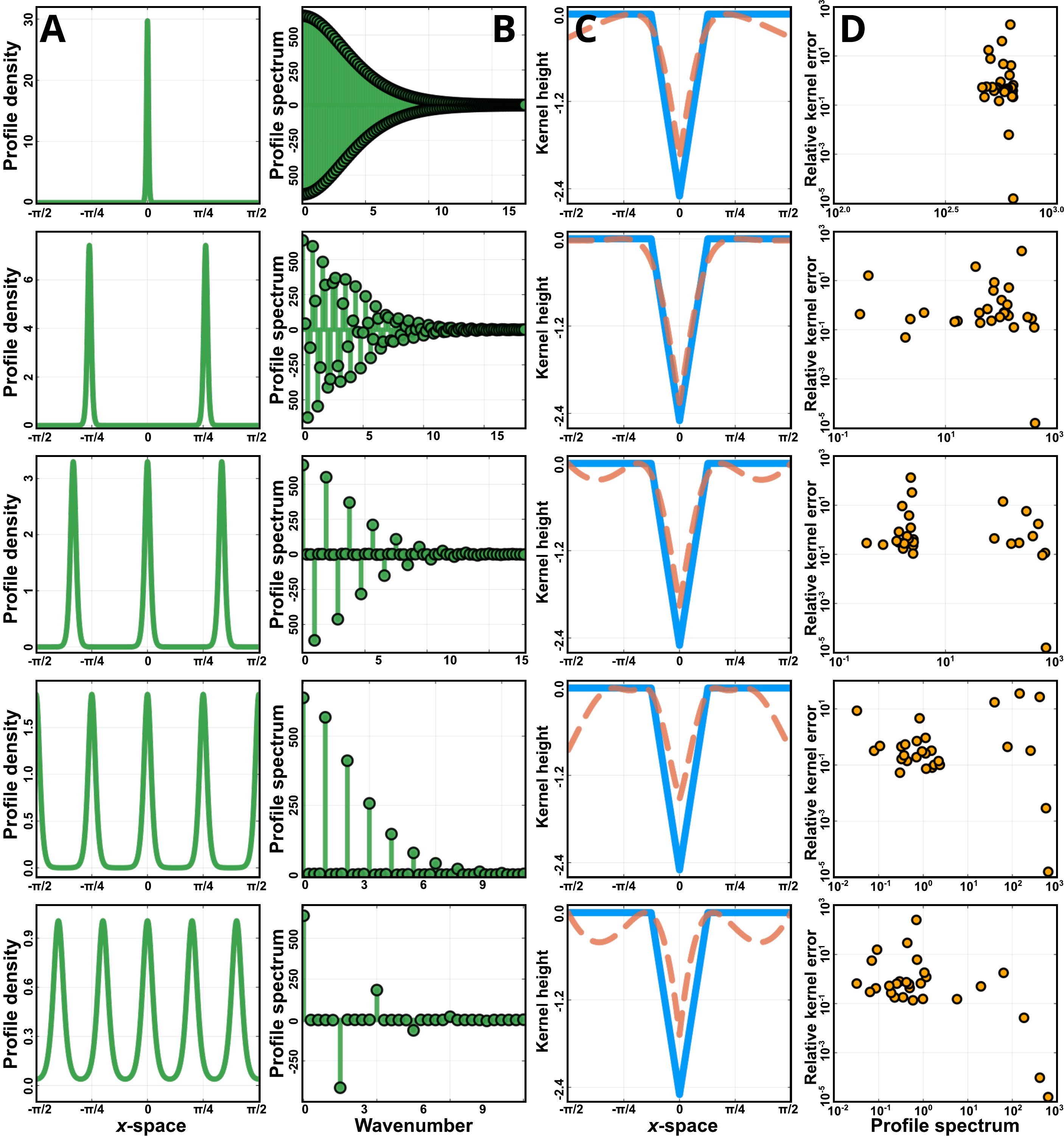}
    \caption{\textbf{Analysis of the impact of the spectral density on fitting performance (Case 2).}  We consider a nonlocal PDE with a triangle kernel $W$ (column C, blue curves), constant $V$, and $\kappa = 19$. It generates five (non-trivial) steady state solutions (A). From every single solution, we check the solution's spectrum (column B), and then attempt to recover a fitted $W^*$ from the solution (column C). In theory, a solution with a rich spectrum should generate a better fit, however, the results are inconclusive. (Column D) For each case, for each wavenumber, we plot the absolute value of that wavenumber for the solution in A (x-axis) compared to the relative error between the true and fitted kernels. In theory, this should generate a negative slope, as we get better fits for more prominent wavenumbers. This phenomenon holds.}
    \label{fig:fig_3_sup_spectrum_2}
\end{figure}

\begin{figure}[htbp]
    \centering
    \includegraphics[width=0.98\linewidth]{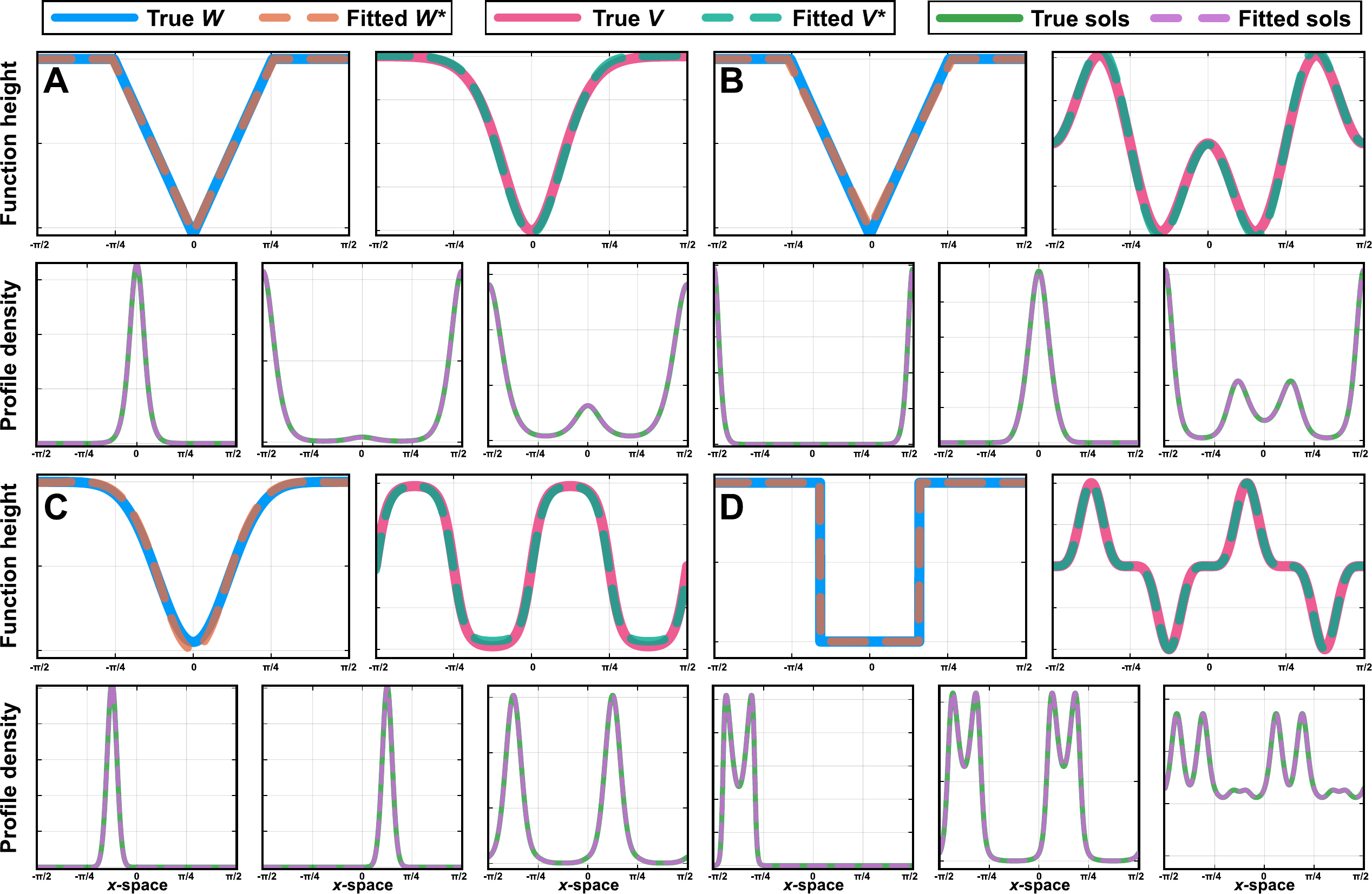}
    \caption{\textbf{A wide range of potential $W$ and $V$ combinations can be recovered from the solutions.} (A-D) To confirm the general nature of our results, we repeat them for a range of interaction kernels $W$, potential functions $V$, and values of $\kappa$. In each case, we confirm that the fitted functions (dashed lines) follow the true functions (solid lines) and that the fitted functions generate the same solutions (dashed purple lines) as the ground truth functions (solid green lines). Finally, we note that the fitted values $\kappa^*$ ($6.1$, $10.9$, $17.8$, and $20.1$, for A, B, C, and D, respectively) align well with the true values ($6.0$, $10.0$, $20.0$, and $20.0$).}
    \label{fig:fig_4_sup_additional_nPDEs}
\end{figure}

All neural networks used throughout this work are fully connected feedforward neural networks. The width and depth are typically between three and four, but vary from case to case. The activation functions are softplus, except in Figures~\ref{fig:fig_1_sup_alt_Ws}A,B,~\ref{fig:fig_3_sup_spectrum_2}, and~\ref{fig:fig_4_sup_additional_nPDEs}A,B,D, where ReLU is used, and Figure~\ref{fig:fig_1_sup_alt_Ws}D, where a mix of ReLU and softplus is used. In these cases, ReLU should better capture the sought piecewise constant components. Notably, both softplus and ReLU are nonnegative. By implementing these in the output layer, we can ensure that the unmodified neural network is nonnegative, which is a basis for follow-up transformations.

We then apply transformations on fitted functions to ensure they conform to the required forms for either $W$ or $V$ as imposed by the original PDE. For the kernels $W$, it is assumed that they are: (i) symmetric so that the original problem remains a gradient-flow with fixed-point structure; and (ii) $W\leq0$ and $\max_{x} \{ W \} = 0$ to fix the degree of freedom offered by invariance under vertical shifts of the kernel. Here, for an unmodified fitted neural network $\NN_{pre,W}(x, \vec{\theta}_W)$ (which is nonnegative), we let $W^*(x, \vec{\theta}_W) = \min(\NN_{pre,W}(x, \vec{\theta}_W))-\NN_{pre,W}(|x|, \vec{\theta}_W)$, which produces a function of the desired form. For the external potential $V$ we have only the condition that $\int_\mathbb{T}V \dx = 0$, again due to invariance with respect to vertical shifts of $V$, and so we apply the transformation $V^*(x, \vec{\theta}_V) = \NN_{pre,V}(x, \vec{\theta}_V) - L^{-1}\int_\mathbb{T}\NN_{pre,V}(x, \vec{\theta}_V)\dx$.

% ------------------------------------------------------------------------------------------------

\subsection{Noisy data generation}\label{sup_section:detailed_methods_noisy_data}

All downsampled solution profiles are sampled at a grid of $N$ evenly distributed datapoints. While different noise models can be used, for simplicity, we will, for each datapoint, draw a normally distributed value $\mathcal{N}(x,\sigma^2)$, where $x$ is the ground truth value, and the standard deviation $\sigma$ is determined by the noise level for that solution set. If the resulting data value is negative, it is set to zero. Sampling densities ($N$) and noise levels ($\sigma$) for each figure are listed in Section~\ref{sup_section:pde_list}.

% ------------------------------------------------------------------------------------------------

\section{List of studied PDE instances}\label{sup_section:pde_list}

In this section, we provide Table~\ref{tab:PDE_list} that lists the functions $W(x)$, $V(x)$, and parameter $\kappa$ used for the PDEs studied in each main article figure. We also provide Table~\ref{tab:data_list}, listing the sample number ($N$) and noise magnitude ($\sigma$) used for each noisy solution profile set. Tables~\ref{tab:PDE_list_sup} and~\ref{tab:data_list_sup} provide similar information, but for PDEs and noisy solution profiles studied in the supplementary figures.

% \raggedbottom

% ------------------------------------------------------------------------------------------------

\begin{table}[!htbp]
    \centering
    % \footnotesize
    \begin{tabular}{|c||c|c|c|}
    \hline
    Figure & $W(x)$ & $V(x)$ & $\kappa$ \\ 
    \hline\hline 
    Figure~\ref{fig:fit_W_to_sols} & $W_{\textup{mm}}(x; 3, 1)$  & $V_{\textup{const}}(x)$ & $8$ \\ \hline
    Figure~\ref{fig:noisy_ss_fit} & $W_{\textup{mm}}(x; 3, 1)$  & $V_{\textup{const}}(x)$ & $8$ \\ \hline
    Figure~\ref{fig:partial_noisy_ss_fit} & $W_{\textup{mm}}(x; 3, 1)$  & $V_{\textup{const}}(x)$ & $8$ \\ 
    \hline
    Figure~\ref{fig:fit_WVk_to_sols} & $W_{\textup{mm}}(x; 2, 1.5)$  & $V_{\textup{plat}}(x;2, 1.5)$ & $5$ \\ 
    \hline
    Figure~\ref{fig:bif_analysis} & $W_{\textup{mm}}(x; 2, 1.5)$  & $V_{\textup{plat}}(x;2, 1.5)$ & $4.0$, $4.02$, $4.1$, $4.5$, $5.0$, $6.0$ \\ 
    \hline
    \end{tabular}
    \medskip
    \caption{\textbf{Specific instances of functional forms.} Similar information for the supplementary figures is provided in Supplementary Table~\ref{tab:PDE_list_sup}. General function formulae are found in Table~\ref{tab:functional_form_formulas}.} 
    \label{tab:PDE_list}
\end{table}

\bigskip

% ------------------------------------------------------------------------------------------------

\begin{table}[!htbp]
\centering
% \small
\renewcommand{\arraystretch}{1.4}

\begin{tabularx}{\textwidth}{
    >{$}l<{$}@{\;\:=\;}X
    >{$}l<{$}@{\;\:=\;}X
}
% \toprule

W_{\mathrm{mm}}(x;n,d)
&
$-e^{-d x^{2}}\cos^{2}\!\left(\dfrac{nx}{2}\right)$
&
W_{\mathrm{tri}}(x;w)
&
$\displaystyle
\begin{cases}
|x|-w, & |x|\le w,\\
0, & |x|>w,
\end{cases}$
\\

\addlinespace

W_{\mathrm{th}}(x;w)
&
$\displaystyle
\begin{cases}
-1, & |x|\le w,\\
0, & |x|>w,
\end{cases}$
&
W_{\mathrm{exp}}(x)
&
$e^{-L/2}-e^{-|x|}$
\\

\addlinespace

V_{\mathrm{const}}(x)
&
$0$
&
V_{\mathrm{plat}}(x;a,n)
&
$\tanh\!\big(a\sin(2nx)\big)$
\\

\addlinespace

V_{\mathrm{sink}}(x;w)
&
$-\cos\!\left(
\dfrac{4}{\pi}
\dfrac{\tanh(wx)}
{\tanh(w\pi/2)}
\right)$
&
V_{\mathrm{wave}}(x;n,d)
&
$-\cos(2nx)+\dfrac{\cos(6nx)}{d}$
\\

\addlinespace

\multicolumn{4}{c}{
$V_{\mathrm{mount}}(x;m,n)\;=\;\sin^{\,1+2m}(2nx)$
}
\\

% \bottomrule

\end{tabularx}
\medskip
\caption{\textbf{General functional forms.}
Definitions of the functional forms used as PDE inputs. Each form contains up to two parameters controlling features such as amplitude, width, or frequency. Particular parameter choices used throughout the manuscript are listed in Table~\ref{tab:PDE_list}. All functional forms are subsequently normalised to satisfy the conditions described in Section~\ref{sup_section:detailed_methods_nn_arch}.}
\label{tab:functional_form_formulas}

\end{table}

\begin{table}[!htbp]
    \centering
    \begin{tabular}{|c||cc|}  \hline
         Figure & $N$ & $\sigma$ \\ \hline \hline 
         Figure~\ref{fig:noisy_ss_fit}B,E & $100$ & $0.01$ \\ \hline
         Figure~\ref{fig:noisy_ss_fit}C,F & $100$ & $0.025$ \\ \hline
         Figure~\ref{fig:noisy_ss_fit}D,G & $100$ & $0.05$ \\ \hline
         Figure~\ref{fig:partial_noisy_ss_fit} & $100$ & $0.01$  \\ \hline
    \end{tabular}
    \medskip
    \caption{\textbf{Data parameters for relevant figures.} For each figure where downsampled noisy data is used, this table lists the number of data samples ($N$) and noise magnitude (i.e. standard deviation of the normal distribution, $\sigma$). Similar information for supplementary figures is provided in Supplementary Table~\ref{tab:data_list_sup}. Data generation details are described in Supplementary Section~\ref{sup_section:detailed_methods_noisy_data}.}
    \label{tab:data_list}
\end{table}

\clearpage

% ------------------------------------------------------------------------------------------------

\section{List of supplementary figure PDE instances}

% ------------------------------------------------------------------------------------------------

\begin{table}[!htbp]
    \centering
    \footnotesize
    \begin{adjustbox}{max width=\textwidth}
    \begin{tabular}{|c||c|c|c|}
    \hline
    Figure & $W(x)$ & $V(x)$ & $\kappa$ \\
    \hline\hline
    Figure~\ref{fig:fig_1_sup_alt_loss}A & $W_{\textup{tri}}(x; 0.6)$  & $V_{\textup{const}}(x)$ & $10$ \\ 
    \hline
    Figure~\ref{fig:fig_1_sup_alt_loss}B & $W_{\textup{th}}(x; 0.5)$  & $V_{\textup{const}}(x)$ & $6$ \\ 
    \hline
    Figure~\ref{fig:fig_1_sup_alt_loss}C & $W_{\textup{mm}}(x; 5, 2)$  & $V_{\textup{const}}(x)$ & $10$ \\ 
    \hline
    Figure~\ref{fig:fig_1_sup_alt_loss}D & 
    $\displaystyle
        \begin{cases}
        W_{\textup{th}}(x; 0.5), & |x| \leq 0.25, \\
        W_{\textup{mm}}(x; 3, 1.5), & |x| > 0.25
        \end{cases}
        $ & $V_{\textup{const}}(x)$ & $20$ \\ 
        \hline
        Figure~\ref{fig:fig_1_sup_fouier_ufa}A & $W_{\textup{exp}}(x)$  & $V_{\textup{const}}(x)$ & $10$ \\ \hline
        Figure~\ref{fig:fig_1_sup_fouier_ufa}B & 
    $\displaystyle
        \begin{cases}
        W_{\textup{th}}(x; 0.5), & |x| \leq 0.15, \\
        W_{\textup{mm}}(x; 3, 1.5), & |x| > 0.15
        \end{cases}
    $ & $V_{\textup{const}}(x)$ & $10$ \\ 
    \hline
    Figure~\ref{fig:fig_1_sup_fouier_ufa}C & $W_{\textup{mm}}(x; 5, 0.5)$  & $V_{\textup{const}}(x)$ & $20$ \\ 
    \hline
    Figure~\ref{fig:fig_1_sup_fouier_ufa}D & $W_{\textup{mm}}(x; 3, 2)$  & $V_{\textup{const}}(x)$ & $6$ \\ 
    \hline
    Figure~\ref{fig:fig_1_sup_fitting_approaches}A,C & $W_{\textup{mm}}(x; 3, 0.5)$  & $V_{\textup{const}}(x)$ & $16$ \\ 
    \hline
    Figure~\ref{fig:fig_1_sup_fitting_approaches}B,D & $W_{th}(x;0.4)$  & $V_{\textup{const}}(x)$ & $12$ \\ 
    \hline
    Figure~\ref{fig:fig_1_sup_direct_fit} & $W_{\textup{mm}}(x; 3, 1)$  & $V_{\textup{const}}(x)$ & $8$ \\ 
    \hline
    Figure~\ref{fig:fig_1_sup_loss_traces} & $W_{\textup{mm}}(x; 3, 1)$  & $V_{\textup{const}}(x)$ & $8$ \\ 
    \hline
    Figure~\ref{fig:fig_1_sup_practical_identifiability} & $W_{\textup{mm}}(x; 3, 1)$  & $V_{\textup{const}}(x)$ & $8$ \\ 
    \hline
    Figure~\ref{fig:fig_1_sup_bad_sols} & $W_{\textup{mm}}(x; 3, 1)$  & $V_{\textup{const}}(x)$ & $8$ \\ 
    \hline
    Figure~\ref{fig:fig_1_sup_single_sols} & $W_{\textup{mm}}(x; 3, 1)$  & $V_{\textup{const}}(x)$ & $8$ \\ 
    \hline
    Figure~\ref{fig:fig_1_sup_alt_Ws}A & $W_{\textup{tri}}(x; 0.6)$  & $V_{\textup{const}}(x)$ & $10$ \\ 
    \hline
    Figure~\ref{fig:fig_1_sup_alt_Ws}B & $W_{\textup{th}}(x; 0.5)$  & $V_{\textup{const}}(x)$ & $10$ \\ \hline
    Figure~\ref{fig:fig_1_sup_alt_Ws}C & $W_{\textup{mm}}(x; 5, 2)$  & $V_{\textup{const}}(x)$ & $10$ \\ \hline
    Figure~\ref{fig:fig_1_sup_alt_Ws}D & 
    $\displaystyle
    \begin{cases}
    W_{\textup{th}}(x; 0.5), & |x| \leq 0.25,\\
    W_{\textup{mm}}(x; 3, 1.5), & |x| > 0.25
    \end{cases}
    $ & $V_{\textup{const}}(x)$ & $10$ \\ \hline
    Figure~\ref{fig:fig_2_sup_noisy_fit_sols} & $W_{\textup{mm}}(x; 3, 1)$  & $V_{\textup{const}}(x)$ & $8$ \\ \hline
    Figure~\ref{fig:fig_2_sup_repeats} & $W_{\textup{mm}}(x; 3, 1)$  & $V_{\textup{const}}(x)$ & $8$ \\ \hline
    Figure~\ref{fig:fig_2_sup_practical_identifiability} & $W_{\textup{mm}}(x; 3, 1)$  & $V_{\textup{const}}(x)$ & $8$ \\ \hline
    Figure~\ref{fig:fig_2_sup_alt_W} & $W_{\textup{mm}}(x; 1,3)$  & $V_{\textup{const}}(x)$ & $30$ \\ \hline
    Figure~\ref{fig:fig_3_sup_spectrum_1} & $W_{\textup{mm}}(x; 3,2)$  & $V_{\textup{const}}(x)$ & $8$ \\ \hline
    Figure~\ref{fig:fig_3_sup_spectrum_2} & $W_{\textup{tri}}(x; 0.4)$  & $V_{\textup{const}}(x)$ & $19$ \\ \hline
    Figure~\ref{fig:fig_4_sup_practical_identifiability} & $W_{\textup{mm}}(x; 2, 1.5)$  & $V_{\textup{plat}}(x; 2, 1.5)$ & $5$ \\ \hline
    Figure~\ref{fig:fig_4_sup_additional_nPDEs}A & $W_{\textup{tri}}(x; 0.8)$  & $V_{\textup{sink}}(x; 1.5)$ & $6$ \\ \hline
    Figure~\ref{fig:fig_4_sup_additional_nPDEs}B & $W_{\textup{tri}}(x; 0.8)$  & $V_{\textup{wave}}(x; 1, 1.0)$ & $10$ \\ \hline
    Figure~\ref{fig:fig_4_sup_additional_nPDEs}C & $W_{\textup{mm}}(x; 1, 3.0)$  & $V_{\textup{plat}}(x; 2.0, 2)$ & $20$ \\ \hline
    Figure~\ref{fig:fig_4_sup_additional_nPDEs}D & $W_{\textup{th}}(x; 0.5)$  & $V_{\textup{mount}}(x; 2, 2)$ & $20$ \\ \hline
    Figure~\ref{fig:fig_4_sup_partial_sols} & $W_{\textup{mm}}(x; 2, 1.5)$  & $V_{\textup{plat}}(x; 2, 1.5)$ & $5$ \\ \hline
    Figure~\ref{fig:fig_5_sols} & $W_{\textup{mm}}(x; 2, 1.5)$  & $V_{\textup{plat}}(x; 2, 1.5)$ & $4.0$, $4.02$, $4.1$, $4.5$, $5.0$, $6.0$ \\ \hline
    Figure~\ref{fig:fig_5_dual_fits} & $W_{\textup{mm}}(x; 2, 1.5)$  & $V_{\textup{plat}}(x; 2, 1.5)$ & $4.0$, $4.02$, $4.1$, $4.5$, $5.0$, $6.0$ \\ 
    \hline
    \end{tabular}
    \end{adjustbox}
    \medskip
    \caption{\textbf{Specific instances of functional forms.} Functions used throughout the supplementary information figures. General function formulae are found in Table~\ref{tab:functional_form_formulas}.} 
    \label{tab:PDE_list_sup}
\end{table}

\enlargethispage{0.5cm}

% ------------------------------------------------------------------------------------------------

\begin{table}[!htbp]
    \centering
    \footnotesize
    \begin{tabular}{|c||cc|}  \hline
         Figure & $N$ & $\sigma$ \\ \hline \hline
         Figure~\ref{fig:fig_1_sup_fitting_approaches}C & $100$ & $0.025$ \\ \hline
         Figure~\ref{fig:fig_1_sup_fitting_approaches}D & $100$ & $0.025$ \\ \hline
         Figure~\ref{fig:fig_2_sup_noisy_fit_sols} & $100$ & $0.025$ \\ \hline
         Figure~\ref{fig:fig_2_sup_repeats} & $100$ & $0.025$ \\ \hline
         Figure~\ref{fig:fig_2_sup_practical_identifiability} & $100$ & $0.025$ \\ \hline
         Figure~\ref{fig:fig_2_sup_alt_W} & $100$ & $0.025$ \\ 
         \hline
    \end{tabular}
    \medskip
    \caption{\textbf{Data parameters for relevant figures.} For each supplementary figure where downsampled noisy data is used, this table lists the number of data samples ($N$) and noise magnitude ($\sigma$). This is an extension of Table~\ref{tab:data_list}, which lists the corresponding information for the main article figures. Data generation details are described in Supplementary Section~\ref{sup_section:detailed_methods_noisy_data}.}
    \label{tab:data_list_sup}
\end{table}

\FloatBarrier

% ------------------------------------------------------------------------------------------------

\section{Result summaries}\label{sup_section:result_sumarries}

In this section, we provide Table~\ref{tab:conditions_tested}, which lists various conditions under which we attempt to recover functional forms, and Table~\ref{tab:results}, which lists various outcomes encountered during the fitting procedure. In both tables, we reference figures where the corresponding condition/outcome is used/encountered.

% ------------------------------------------------------------------------------------------------

\begin{table}[!h]
    \centering
    \resizebox{\textwidth}{!}{
    \begin{tabular}{|l||l|}  \hline
         Condition & Relevant Figures \\ \hline \hline 
         Recovery using single solution & Figure~\ref{fig:partial_noisy_ss_fit}, Figures~\ref{fig:fig_1_sup_single_sols}, \ref{fig:fig_3_sup_spectrum_1}, \ref{fig:fig_3_sup_spectrum_2}, \ref{fig:fig_4_sup_partial_sols}. \\ \hline
         Recovery using partial set of solutions & Figures~\ref{fig:partial_noisy_ss_fit}, \ref{fig:bif_analysis}. Figures~\ref{fig:fig_4_sup_partial_sols}, \ref{fig:fig_5_dual_fits}.   \\ \hline
         Sparse sampling with measurement noise & Figures \ref{fig:noisy_ss_fit}, \ref{fig:partial_noisy_ss_fit}, Figures~\ref{fig:fig_2_sup_noisy_fit_sols}, \ref{fig:fig_2_sup_repeats}, \ref{fig:fig_2_sup_practical_identifiability}, \ref{fig:fig_2_sup_alt_W}. \\ \hline
         Recovery of multiple functional parameters &  Figures \ref{fig:fit_WVk_to_sols}, \ref{fig:bif_analysis}, Figures~\ref{fig:fig_4_sup_practical_identifiability}, \ref{fig:fig_4_sup_additional_nPDEs}, \ref{fig:fig_4_sup_partial_sols}, \ref{fig:fig_5_dual_fits}. \\ \hline
         Recovery from multiple solutions on the same bifurcation branch & Figure~\ref{fig:bif_analysis}, Figure~\ref{fig:fig_5_dual_fits}.  \\ \hline
         Recovery using PDE-residual loss function & Figure~\ref{fig:fig_1_sup_alt_loss}. \\ \hline
         Recovery using solution with varying spectral content & Figures \ref{fig:fig_3_sup_spectrum_1} and \ref{fig:fig_3_sup_spectrum_2}. \\ \hline
    \end{tabular}}
    \caption{\textbf{Recovery conditions investigated.} The conditions are spread across PDEs with different $W$, $V$, and $\kappa$, the details of which are listed in Tables~\ref{tab:PDE_list},\ref{tab:PDE_list_sup}.}
    \label{tab:conditions_tested}
\end{table}

% ------------------------------------------------------------------------------------------------

\begin{table}[!htbp]
    \centering
    \footnotesize
    \resizebox{\textwidth}{!}{
    \begin{tabular}{|p{0.42\textwidth}||p{0.43\textwidth}|} \hline
         Outcome & Relevant Figures \\ \hline \hline 
         Success & Figures \ref{fig:fit_W_to_sols}, \ref{fig:noisy_ss_fit}E,F, \ref{fig:partial_noisy_ss_fit}, \ref{fig:fit_WVk_to_sols}, \ref{fig:bif_analysis}H,I, Figures \ref{fig:fig_1_sup_single_sols}A,E, \ref{fig:fig_1_sup_alt_Ws}A-D, \ref{fig:fig_1_sup_alt_loss}A-D, \ref{fig:fig_2_sup_repeats}A,C, \ref{fig:fig_2_sup_alt_W}A, \ref{fig:fig_4_sup_additional_nPDEs}A-D,\ref{fig:fig_4_sup_partial_sols}D-F, \ref{fig:fig_5_dual_fits}C-F. \\ \hline
         Successful recovery of functions, erroneous solution profile predictions & Figures \ref{fig:fig_1_sup_bad_sols}, \ref{fig:fig_1_sup_single_sols}C.  \\ \hline
         Unsuccessful recovery of function, correct solution profile predictions & Figure~\ref{fig:fig_2_sup_alt_W}C. \\ \hline
         Unsuccessful recovery and predictions & Figures \ref{fig:noisy_ss_fit}G, \ref{fig:partial_noisy_ss_fit}. \\ \hline
         Unsuccessful recovery due to practical non-identifiability  &  Figure~\ref{fig:bif_analysis}G, Figure~\ref{fig:fig_5_dual_fits}A,B.  \\ \hline
         Unsuccessful recovery due to (demonstrated) structural non-identifiability  &  Figure~\ref{fig:fig_4_sup_partial_sols}A-C. \\ \hline
    \end{tabular}}
    \caption{\textbf{Encountered outcomes of functional recovery.} Throughout our analysis, we note both when functional recovery is possible, but also note multiple (partial or full) modes of failure. Here, we list each such potential case, and the figures in which it is illustrated.}
    \label{tab:results}
\end{table}

% ------------------------------------------------------------------------------------------------

\clearpage

\section{Supplementary figures}

% ------------------------------------------------------------------------------------------------

\begin{figure}[!h]
    \centering
    \includegraphics[width=\textwidth]{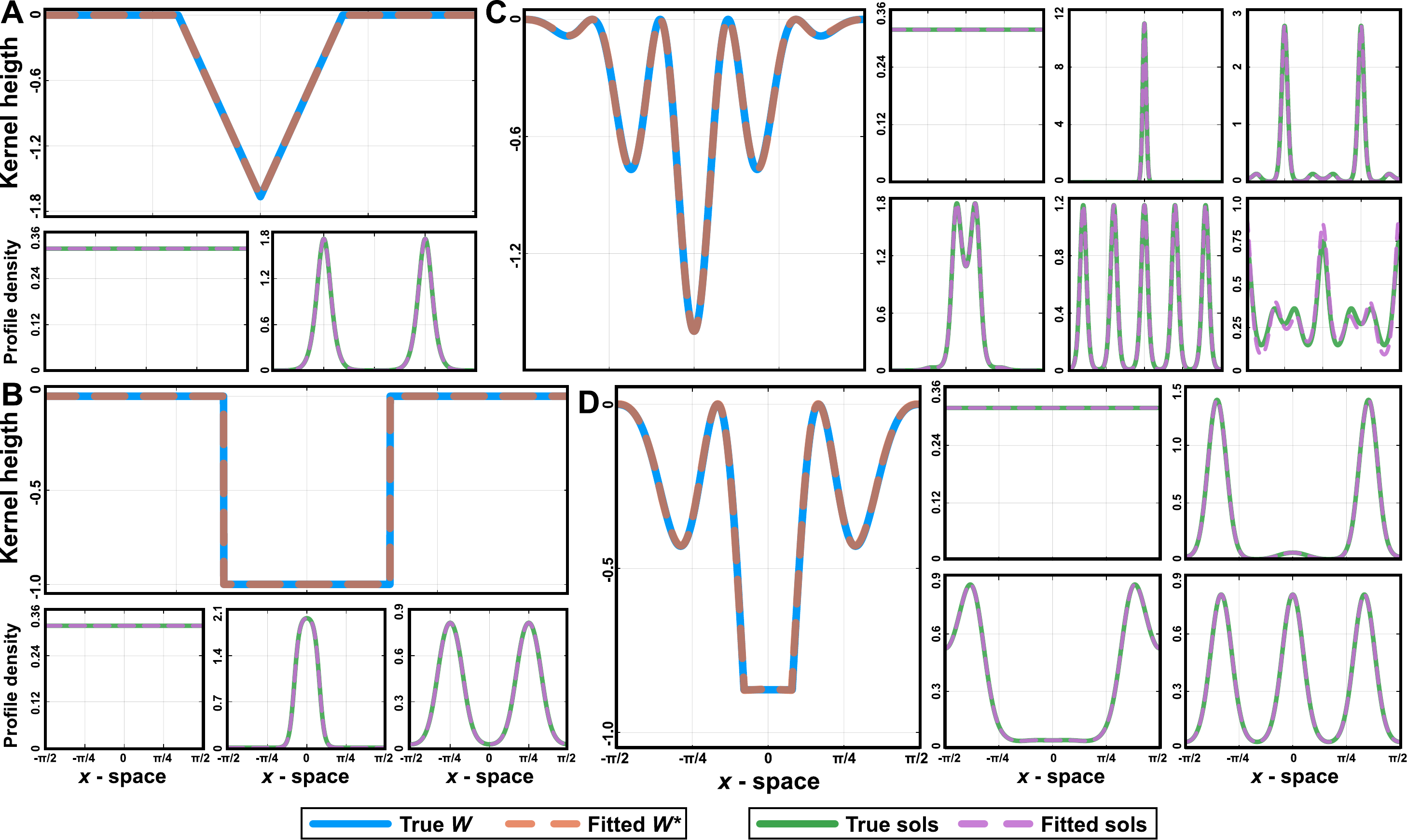}
    \caption{\textbf{Functional parameters can be recovered using an alternative loss function.} To show that our approach is not restricted to the fixed-point residual loss function presented in Section~\ref{section:methods_udes} (Equation~\eqref{eq:loss_function_FP}), we repeat the analysis using the PDE residual loss function presented in the same section (Equation~\eqref{eq:loss_function_PDE}). (A-D) For four different parameterisations of the PDE, we carry out the procedure described in Figure~\ref{fig:workflow_flowchart}. In each case, using the PDE residual, the interaction kernel is correctly recovered, and the fitted PDE yields the ground-truth solution profiles.}
    \label{fig:fig_1_sup_alt_loss}
\end{figure}

% ------------------------------------------------------------------------------------------------

% ------------------------------------------------------------------------------------------------

% Figure 1 %

% ------------------------------------------------------------------------------------------------

% ------------------------------------------------------------------------------------------------

\begin{figure}[htbp]
    \centering
    \includegraphics[width=0.5\linewidth]{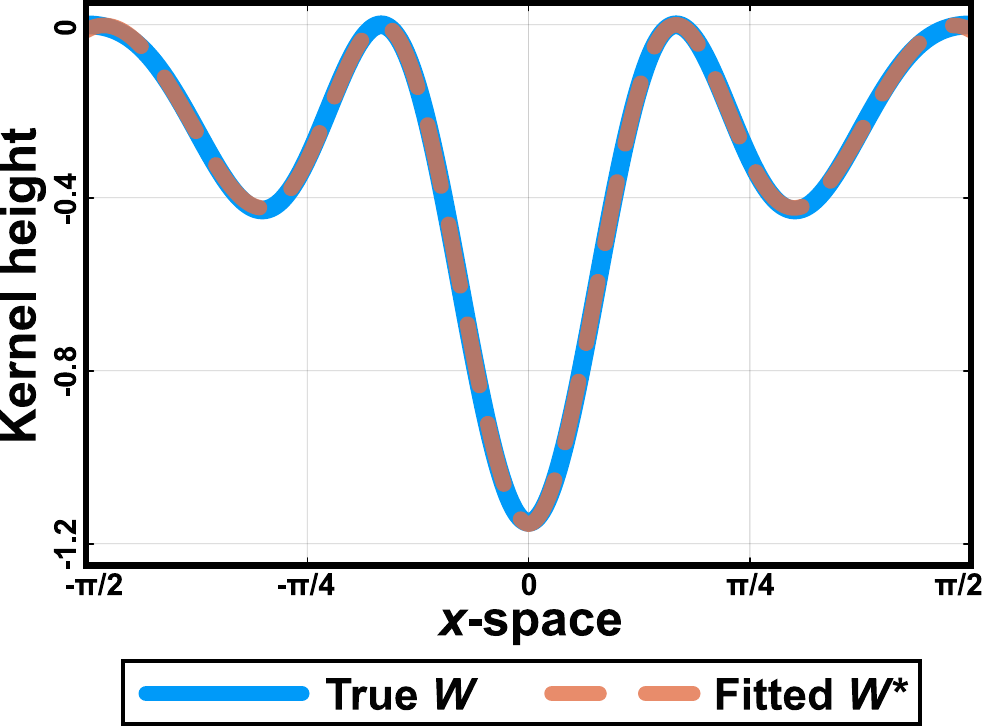}
    \caption{\textbf{The selected neural network architecture can be successfully fitted directly to the true $W$.} Since the true $W$ is known, we can perform a pre-fit optimisation run to confirm that the neural-network architecture used (Supplementary Section \ref{sup_section:detailed_methods_nn_arch}) can represent the $W$ we attempt to fit. Here, we perform a direct fit of the neural network (dashed red line) to the ground truth $W$ (solid blue line). Since the fit is successful, we carry on with (the more computationally intensive) fitting of the neural network to the solution profiles (Figure~\ref{fig:fit_W_to_sols}). Generally, while larger neural networks can fit more general functions, they do so at a greater computational cost (and at an increased risk of overfitting). For real-world applications (where the true $W$ is not known), this kind of pre-fitting check is not possible. However, it is still possible to analyse how the complexity of the neural network affects the fitted function. Previous UDE research has suggested that neural network architecture has only minor effects on results~\cite{philipps_current_2025}.}
    \label{fig:fig_1_sup_direct_fit}
\end{figure}

% ------------------------------------------------------------------------------------------------

\begin{figure}[htbp]
    \centering
    \includegraphics[width=0.5\linewidth]{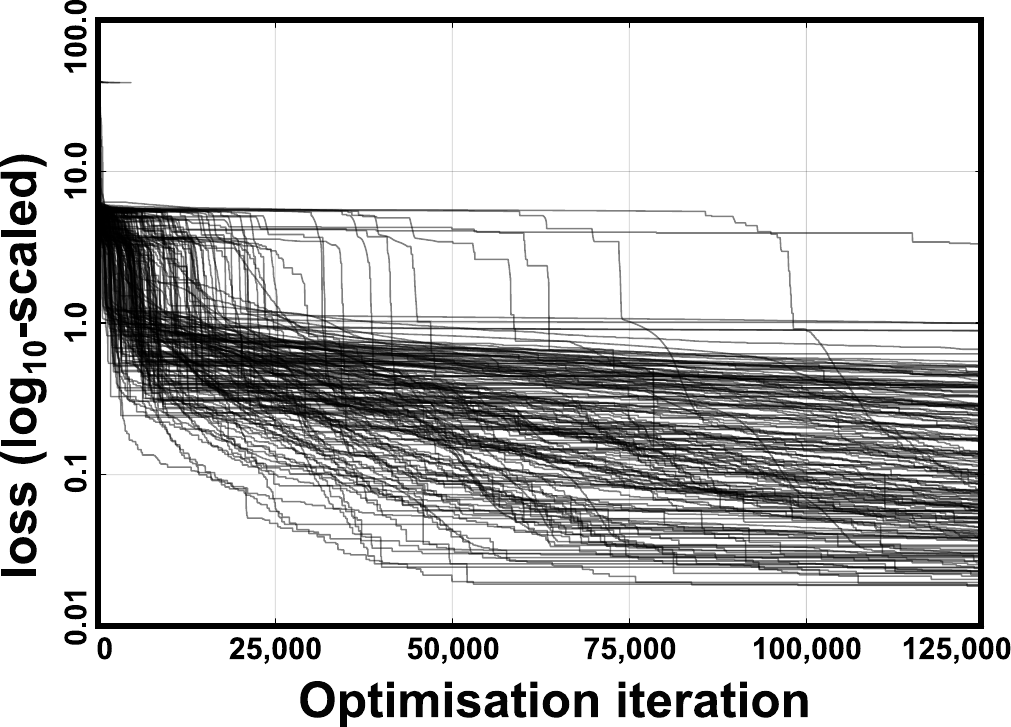}
    \caption{\textbf{Convergence can be confirmed by plotting loss traces.} For each individual optimisation run performed in Figure~\ref{fig:fit_W_to_sols} and for each iteration (also called \textit{epoch}) of the optimisation procedure, we plot the best achieved loss function evaluation so far. Since the curves are flattening out towards the end of the optimisation, we determine that the optimisation has converged. Generally, loss trace progression plots like this one can be used to determine convergence. Finally, we note that, due to the nature of the used Adam optimiser, the actual loss function evaluation at each iteration might be higher than the plotted value.}
    \label{fig:fig_1_sup_loss_traces}
\end{figure}

% ------------------------------------------------------------------------------------------------

\begin{figure}[htbp]
    \centering
    \includegraphics[width=0.66\linewidth]{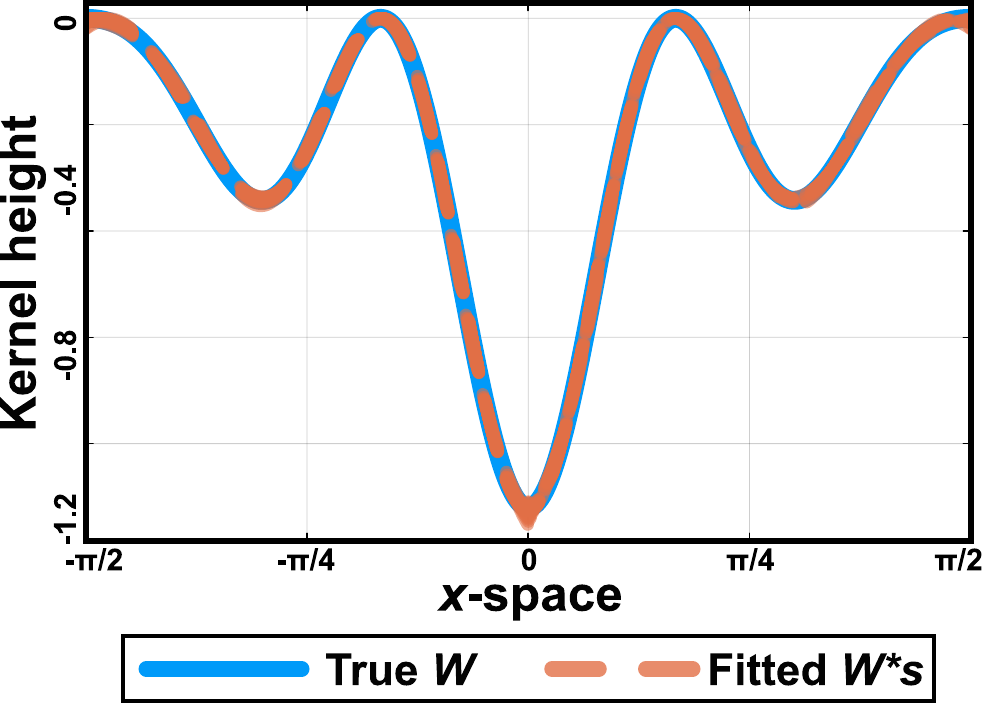}
    \caption{\textbf{Ensemble plots of fitted $W^*$ functions confirm practical identifiability.} In Figure~\ref{fig:fit_W_to_sols}, we confirmed that the fitted $W^*$ aligned with the ground truth $W$. We also confirmed that $W^*$ and $W$ yield the same steady-state solutions. Here, we confirm that $W$ is \textit{practically identifiable} (Section~\ref{sec:function_recovery_and_identifiability}) using an ensemble plot approach~ \cite{loman_funcident_2025}. The plot shows the ensemble of fitted functions $W^*$ achieving a loss function value no more than twice that of the best-fitting $W^*$ (dashed red lines). That they all converge to the same function (which is identical to the ground truth $W$, solid blue line) suggests practical identifiability.}
    \label{fig:fig_1_sup_practical_identifiability}
\end{figure}

% ------------------------------------------------------------------------------------------------

\begin{figure}[htbp]
    \centering
    \includegraphics[width=\textwidth]{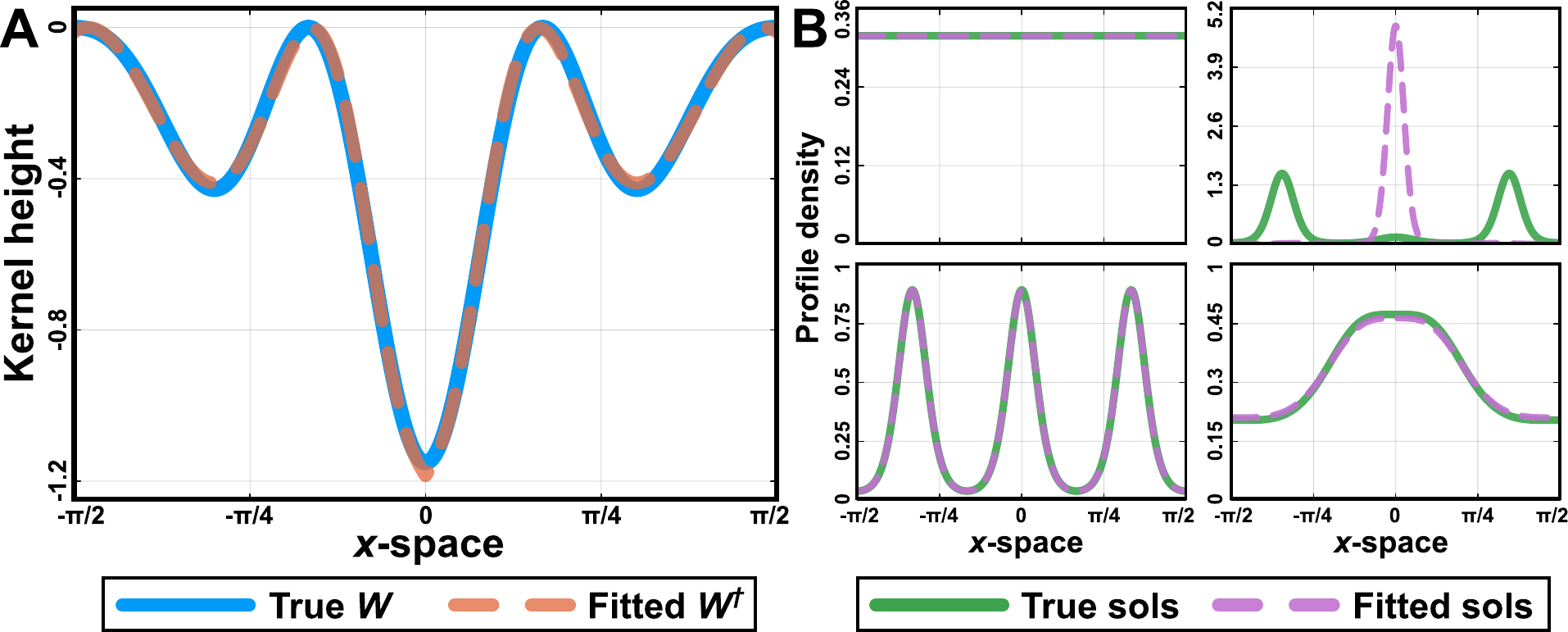}
    \caption{\textbf{A slightly perturbed $W$ can yield a different solution set.} In Figure~\ref{fig:fit_W_to_sols} we showed how the optimally fitting $W^*$ regenerated the solution set of the ground truth $W$. In this figure, we have an alternative function $W^\dagger$ (A, red dashed line), so that $W^\dagger \approx W$ (A, solid blue line). However, while the two kernels are similar, the solution profiles generated by $W^\dagger$ (B, dashed purple lines) differ from those generated by $W$ (B, solid green lines). This suggests that, especially for applications to empirical data, one has to carefully consider the trustworthiness of steady state solutions generated by fitted PDEs.}
    \label{fig:fig_1_sup_bad_sols}
\end{figure}

% ------------------------------------------------------------------------------------------------

\begin{figure}[htbp]
    \centering
    \includegraphics[width=\textwidth]{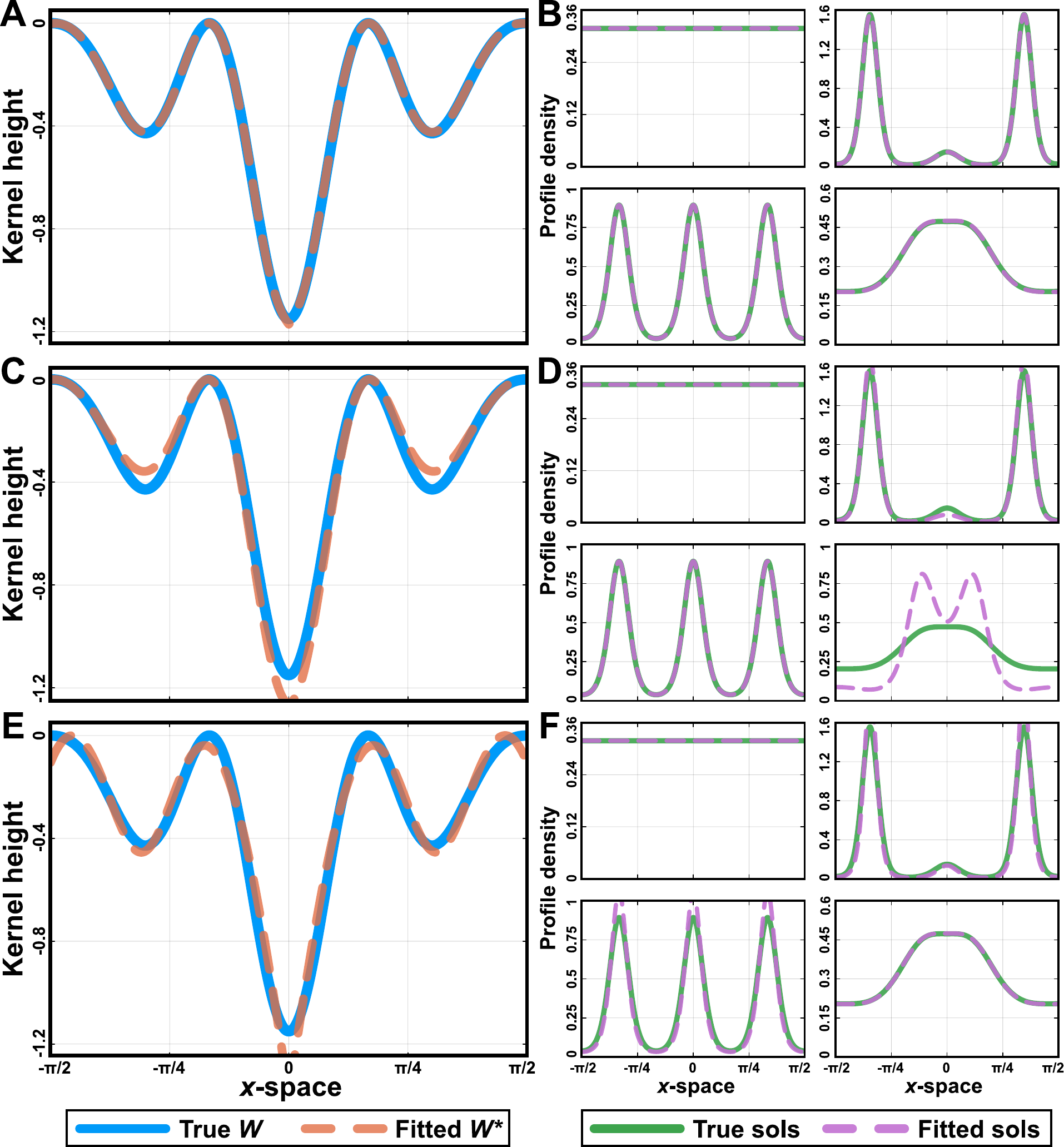}
    \caption{\textbf{The correct $W$ can be recovered from a single solution.} In Figure~\ref{fig:fit_W_to_sols}, we used all four solutions to recover $W$. Here, we perform the same fit three times, but in each case only using a single solution profile (omitting the trivial constant solution case). The double-peak solution is used in A,B, the triple-peak solution in C,D, and the bump in E,F. (A,C,E) While there are slight differences, in each case the fitted $W^*$ (dashed red lines) is a good approximation of the ground truth $W$ (solid blue lines). (B,D,F) In almost all cases, the fitted $W^*$s regenerate the ground truth solution profiles. The exception is the $W^*$ fitted to the triple-peak solution, where only three of the four solutions are regenerated.}
    \label{fig:fig_1_sup_single_sols}
\end{figure}

% ------------------------------------------------------------------------------------------------

% ------------------------------------------------------------------------------------------------

% Figure 2 %

% ------------------------------------------------------------------------------------------------

\begin{figure}[htbp]
    \centering
    \includegraphics[width=0.85\linewidth]{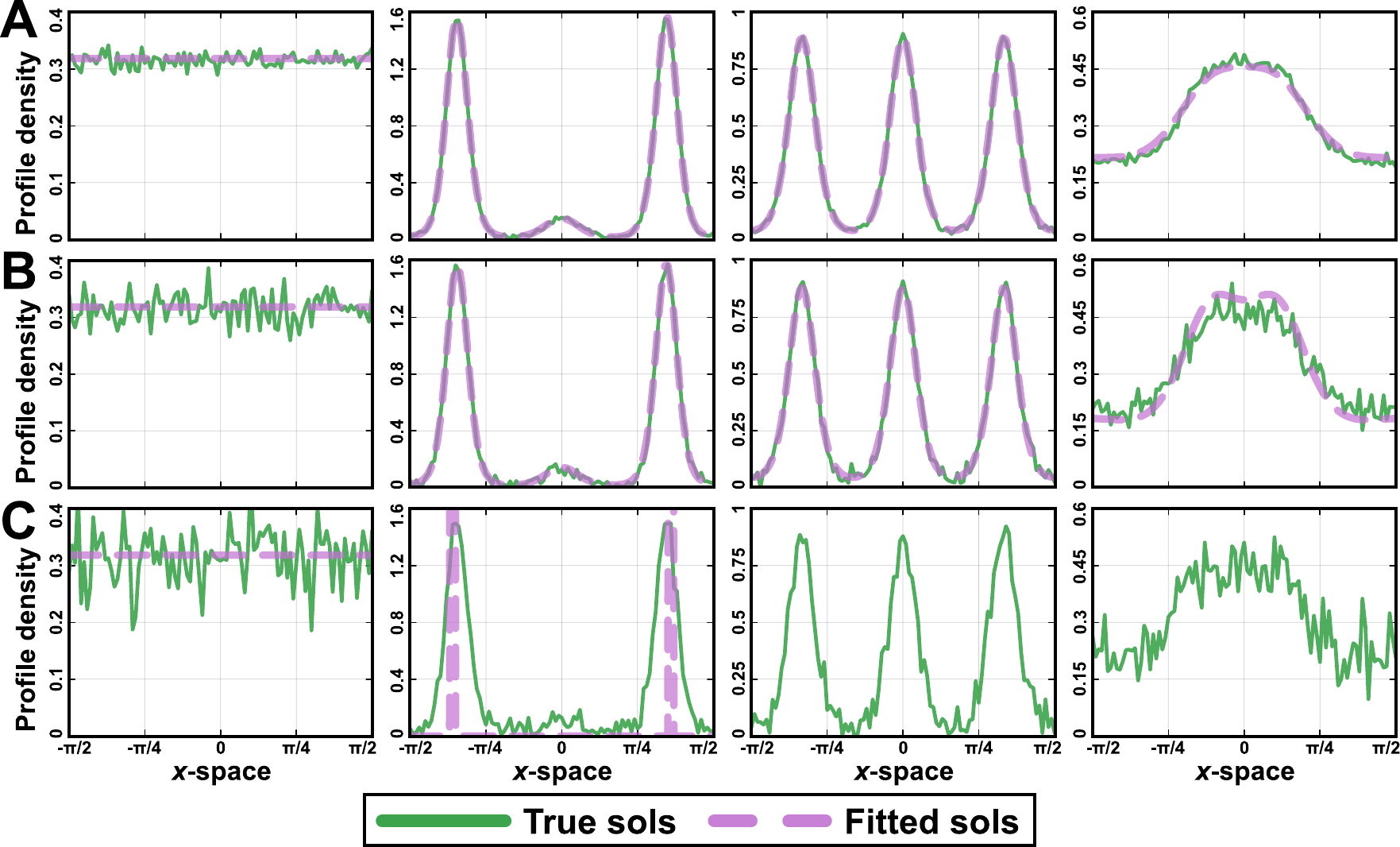}
    \caption{\textbf{PDEs fitted from noisy data retain the full solution sets}. (A-C) The solution profiles generated by the $W^*$s fitted to the solution profiles in Figure~\ref{fig:noisy_ss_fit} (dashed purple lines) compared to the noisy solutions to which they were fitted (solid green lines). (A,B) For the low noise (A) and medium noise (B) cases, the fitted $W^*$s generate (smooth) solution profiles similar to those to which they were fitted. (C) For the high-noise case, not only are we unable to recover the ground truth $W$, but the fitted $W^*$ fails to regenerate the solution profiles.}
    \label{fig:fig_2_sup_noisy_fit_sols}
\end{figure}

% ------------------------------------------------------------------------------------------------

\begin{figure}[htbp]
    \centering
    \includegraphics[width=0.85\linewidth]{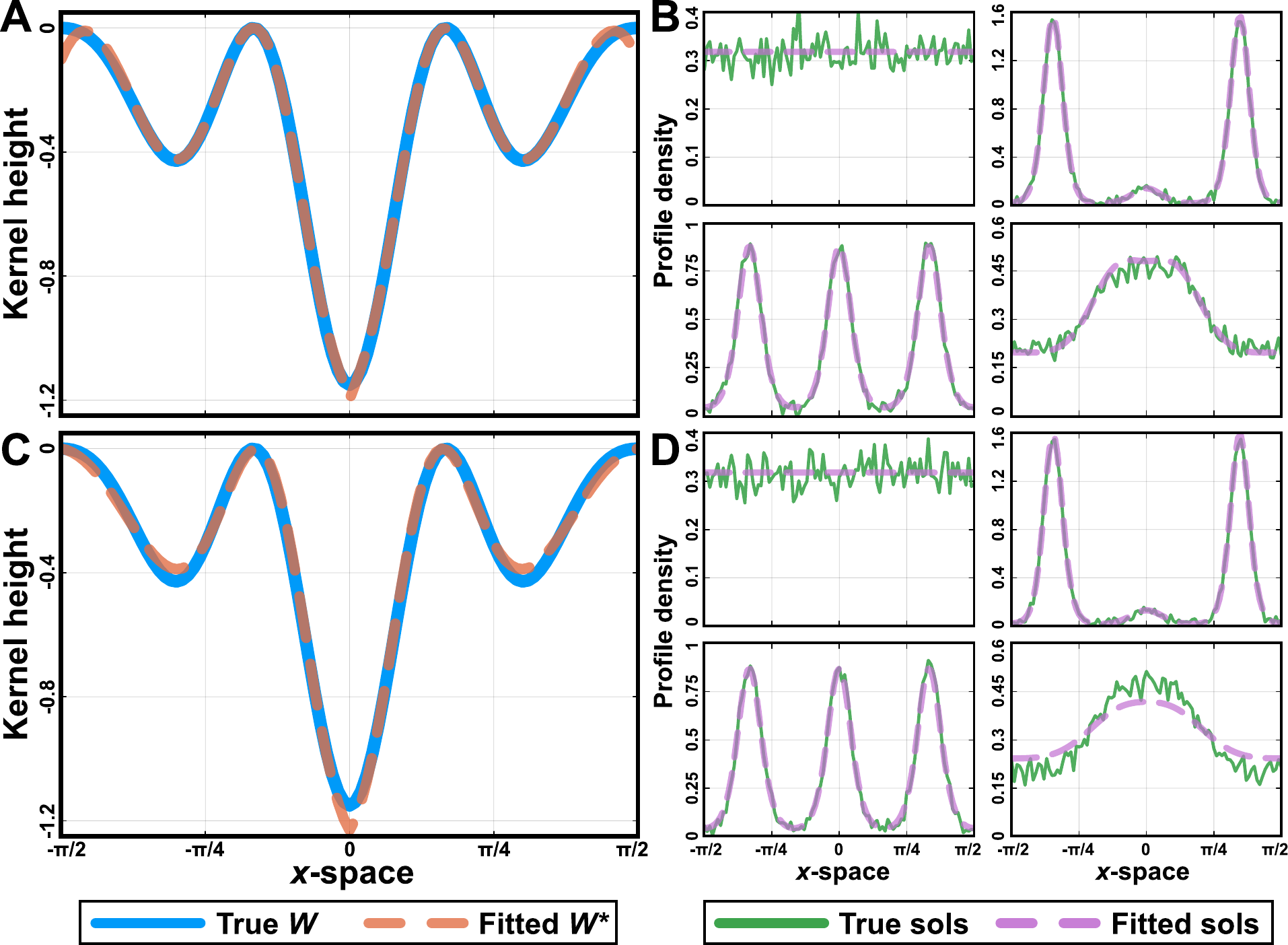}
    \caption{\textbf{The fits to noisy data succeed in numerical repeats.} For the medium noise case in Figure~\ref{fig:noisy_ss_fit}C, we generate two new noisy steady state solution sets. These use new measurements, but the same sampling sparsity and noise distributions as in Figure~\ref{fig:noisy_ss_fit}C. (A,C) In both cases, the fitted $W^*$ (dashed red lines) follows the ground truth $W$ (solid blue lines). (B,D) Furthermore, the fitted $W^*$s solutions (dashed purple lines) correspond to the ground truth solutions (solid green lines).}
    \label{fig:fig_2_sup_repeats}
\end{figure}

% ------------------------------------------------------------------------------------------------

\begin{figure}[htbp]
    \centering
    \includegraphics[width=0.85\linewidth]{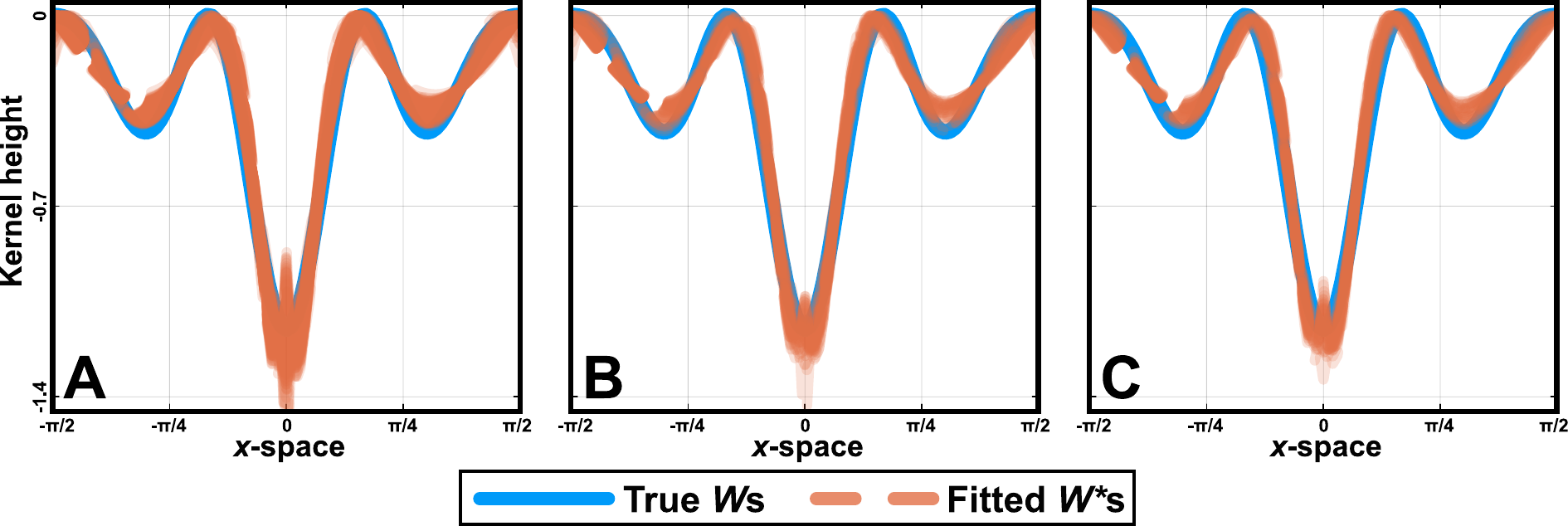}
    \caption{\textbf{$W$ functions recovered from noisy data are practically identifiable}. To ensure that interaction kernels fitted from the noisy solutions are identifiable, we perform ensemble plots. In each case, the ensemble (all functions yielding a loss value $<1.1 L_{opt}$, where $L_{opt}$ is the best found loss function value) is plotted. (A) Ensemble plot for the fit in Figure~\ref{fig:noisy_ss_fit}F. (B) Ensemble plot for the fit in Figure~\ref{fig:fig_2_sup_repeats}A. (C) Ensemble plot for the fit in Figure~\ref{fig:fig_2_sup_repeats}C. In all cases, the ensemble of plotted $W^*$ (dashed red lines) aligns with one another and the ground truth $W$ (solid blue lines), suggesting practical identifiability. }
    \label{fig:fig_2_sup_practical_identifiability}
\end{figure}

% ------------------------------------------------------------------------------------------------

\begin{figure}[htbp]
    \centering
    \includegraphics[width=0.85\linewidth]{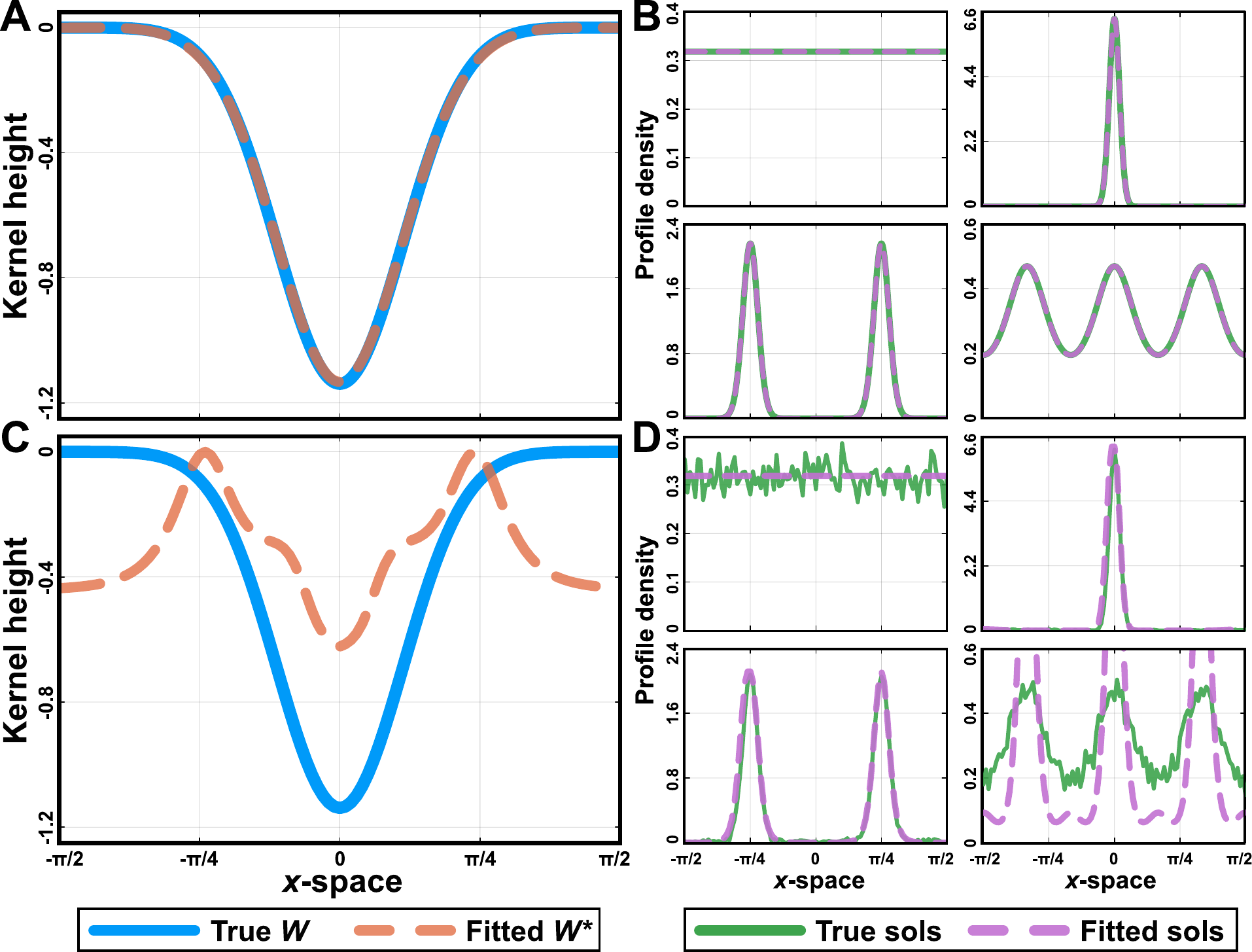}
    \caption{\textbf{Our ability to recover the true $W$ is highly dependent on the PDE}. We generate a new nonlocal PDE with potential $V \equiv 0$, interaction strength parameter $\kappa = 30$, and a unimodal interaction kernel $W$ (A, solid blue line). (B) The new PDE, just like the ones used in Figure~\ref{fig:noisy_ss_fit}, generates four solutions (B, solid green lines). First, we confirm that the ground truth $W$ can be recovered from the non-noisy solutions (A, dashed red line), and that the fitted $W^*$ regenerates the ground truth solutions (B, dashed purple lines). Next, we generate a sparsely sampled, noisy, solution set (using the same sampling frequency and noise distributions as in Figure~\ref{fig:noisy_ss_fit}C) (D, solid green lines). The $W^*$ that we fit from this noisy solution set (C, dashed red line) does not follow the ground truth kernel (C, solid blue line). It is noteworthy (but not necessarily unexpected) that we could fit a more complicated, multi-modal, $W$ to solutions with similar noise levels (Figure~\ref{fig:noisy_ss_fit}F). Finally, we note that while we could not recover the correct $W$, $W^*$ still generates a solution set (D, dashed purple lines) very similar to the ground truth one (D, solid green lines).
    }
    \label{fig:fig_2_sup_alt_W}
\end{figure}

% ------------------------------------------------------------------------------------------------

% Figure 3 %

% ------------------------------------------------------------------------------------------------

\begin{figure}[htbp]
    \centering
    \includegraphics[width=0.85\linewidth]{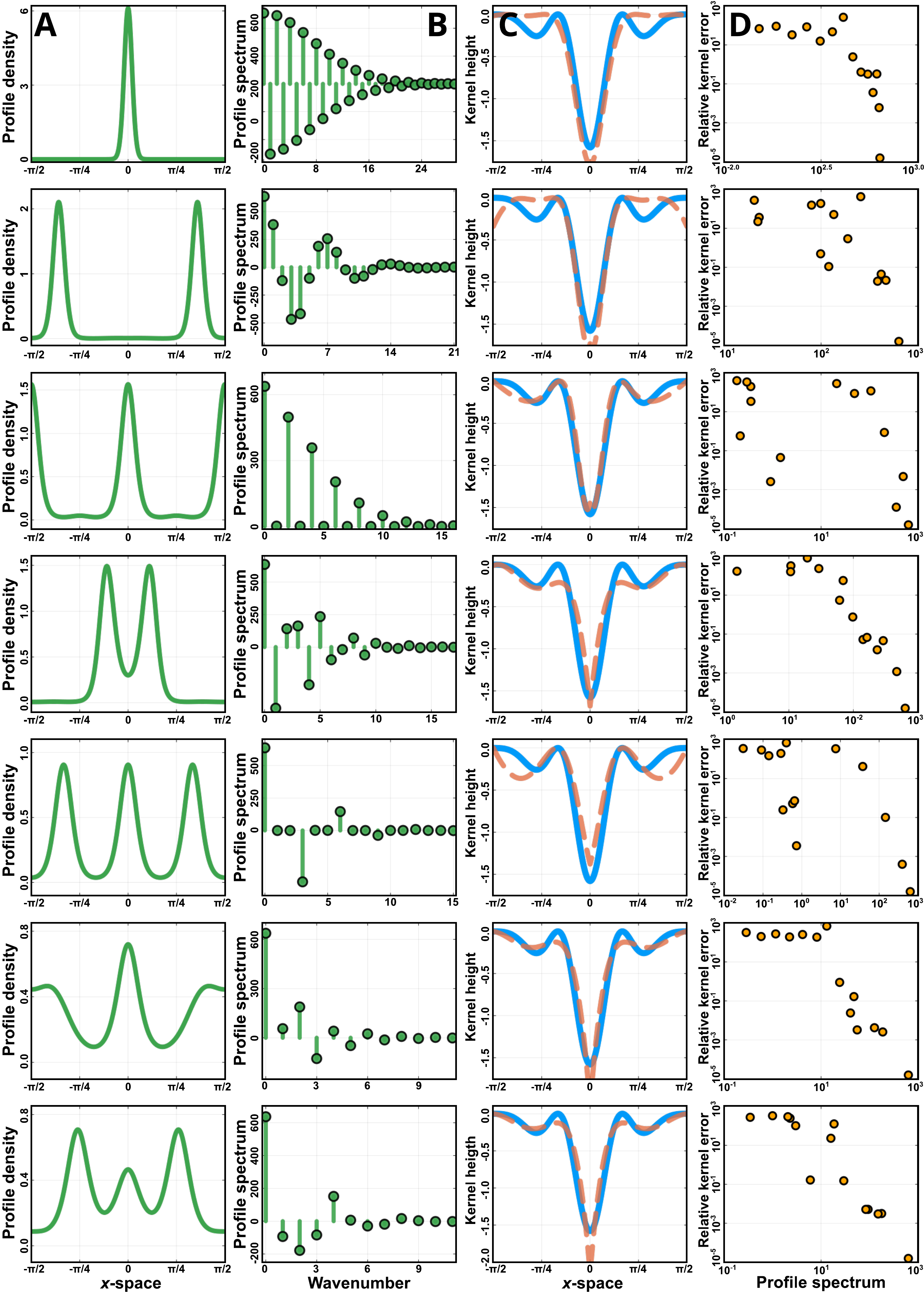}
    \caption{\textbf{Analysis of the impact of the spectral density on fitting performance (Case 1).} We consider a nonlocal PDE with a multimodal $W$ (column C, blue curves), constant $V$, and $\kappa = 8$. It generates seven (non-trivial) steady state solutions (A). From every single solution, we check the solution's spectrum (column B), and then attempt to recover a fitted $W^*$ from the solution (column C). In theory, a solution with a rich spectrum should generate a better fit, however, the results are inconclusive. (Column D) For each case, for each wavenumber, we plot the absolute value of that wavenumber for the solution in A (x-axis) compared to the relative error between the true and fitted kernels. In theory, this should generate a negative slope, as we get better fits for more prominent wavenumbers. This phenomenon holds.}
    \label{fig:fig_3_sup_spectrum_1}
\end{figure}

% ------------------------------------------------------------------------------------------------

% ------------------------------------------------------------------------------------------------

\begin{figure}[htbp]
    \centering
    \includegraphics[width=0.5\linewidth]{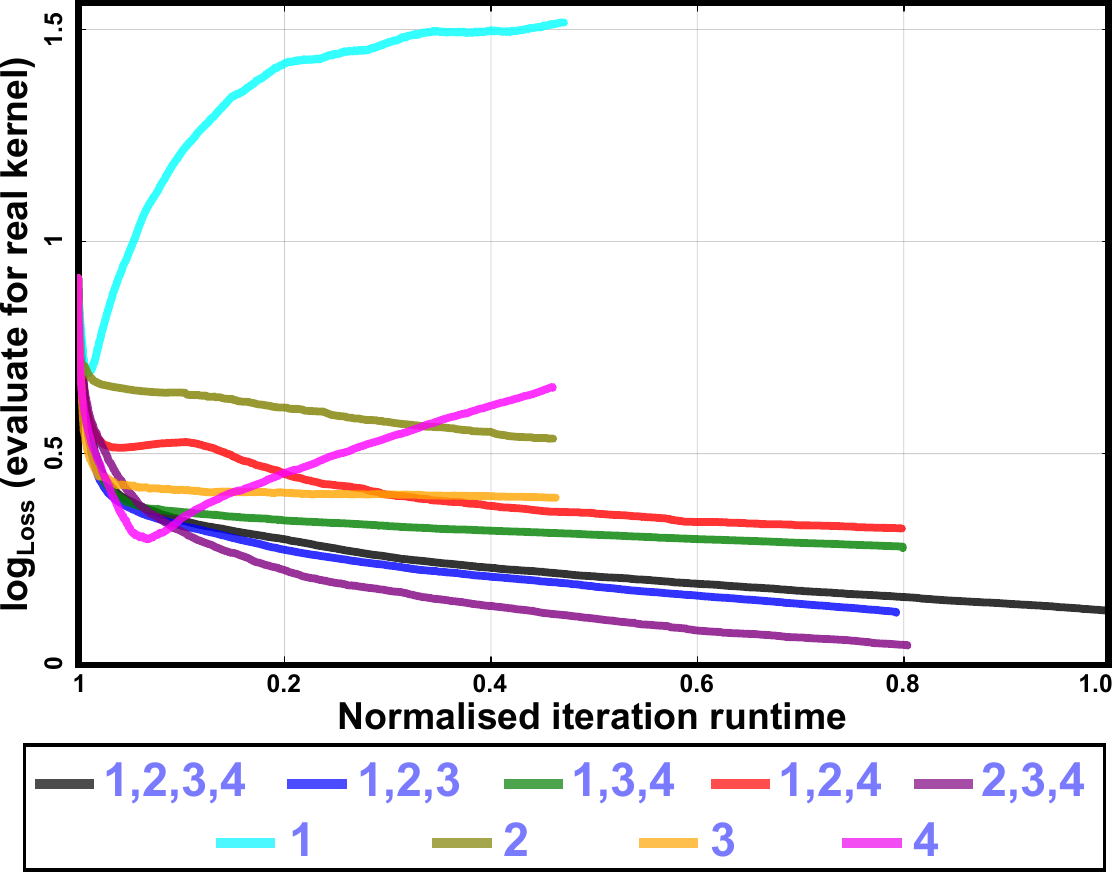}
    \caption{\textbf{Figure \ref{fig:partial_noisy_ss_fit}C loss function progression trajectories normalised by runtime.} In Figure \ref{fig:partial_noisy_ss_fit}C, we plotted the convergence trajectories of the loss function for nine different optimisation runs (each fitting to a different set of solution profiles). That plot, however, does not consider that each optimisation run has a different runtime. Here, optimisation runs that fit against a larger number of solution profiles will take longer to evaluate the same number of optimisation iterations. This plot shows the same optimisation runs as in \ref{fig:partial_noisy_ss_fit}C, but normalises each progression trajectory by the corresponding optimisation run's total runtime. I.e. if $t_i$ is the runtime of the $i$th optimisation run, and $t_{max}$ is the maximum runtime across the nine cases, then the $i$th progression trajectory is normalised to be plotted on the interval $(0, t_i/t_{max})$. When doing so, we see the same pattern as in Figure \ref{fig:partial_noisy_ss_fit}C, however, when considering performance only, convergence is comparably faster for the optimisation runs with fewer solution profiles.}
    \label{fig:fig_3_sup_loss_trajs_normalised}
\end{figure}

% ------------------------------------------------------------------------------------------------

% Figure 4 %

% ------------------------------------------------------------------------------------------------

\begin{figure}[htbp]
    \centering
    \includegraphics[width=\linewidth]{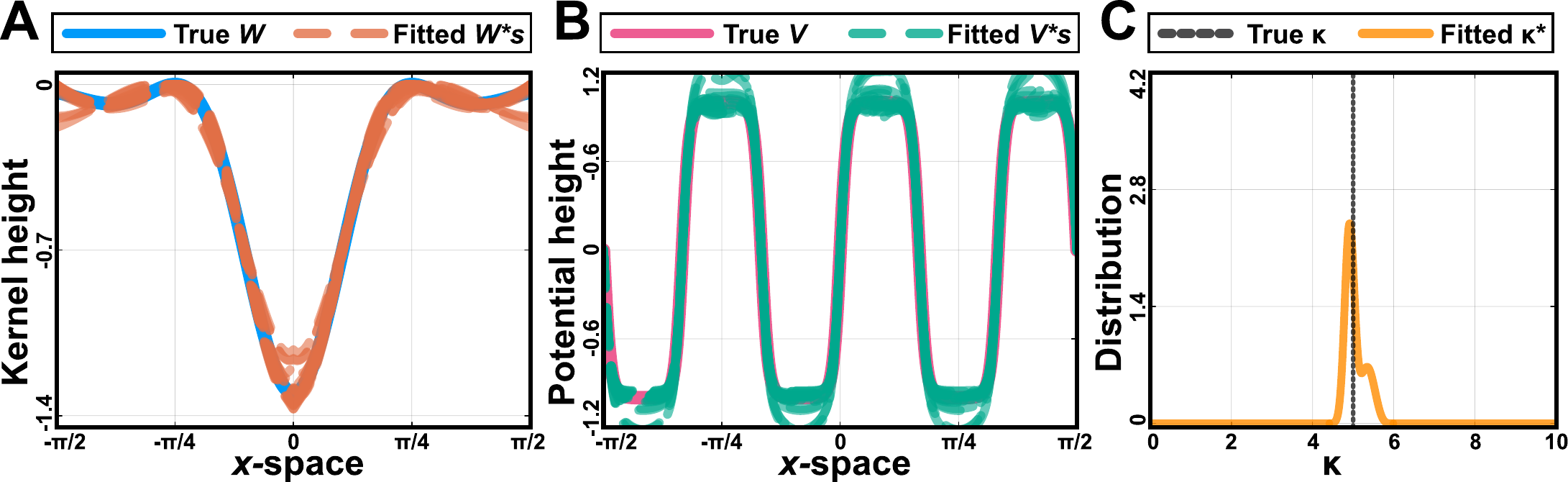}
    \caption{\textbf{Ensemble plots of fitted PDE components confirm identifiability.} In Figure~\ref{fig:fit_WVk_to_sols}, when fitting $(W^*,V^*,\kappa^*)$ to the solutions, we perform an ensemble fit with multiple random initial guesses ($N=800$). Here, we consider the set of optimisation runs achieving a loss function value no more than three times that of the optimal one. (A) The ensemble of well-fitting interaction kernels $W^*$ (dashed red lines) all converge to the same functional form, which also follows the ground truth $W$ (solid blue line). (B) The ensemble of well-fitting potential functions $V^*$ (dashed teal lines) all converge to the same functional form, which also follows the ground truth $V$ (solid pink line). (C) The distribution of well-fitting $\kappa^*$ values (yellow line) all aggregate around the true value (vertical dotted black line). We note that parameter identifiability could have been assessed more comprehensively by using profile likelihood analysis~\cite{kreutz_profile_2013}, however, we here use a distribution plot for simplicity. Taken together, these results show that not only can we recover the correct PDE components from the solution, but we can do so in an identifiable manner.}
    \label{fig:fig_4_sup_practical_identifiability}
\end{figure}

% ------------------------------------------------------------------------------------------------

% ------------------------------------------------------------------------------------------------

\begin{figure}[htbp]
    \centering   \includegraphics[width=0.85\linewidth]{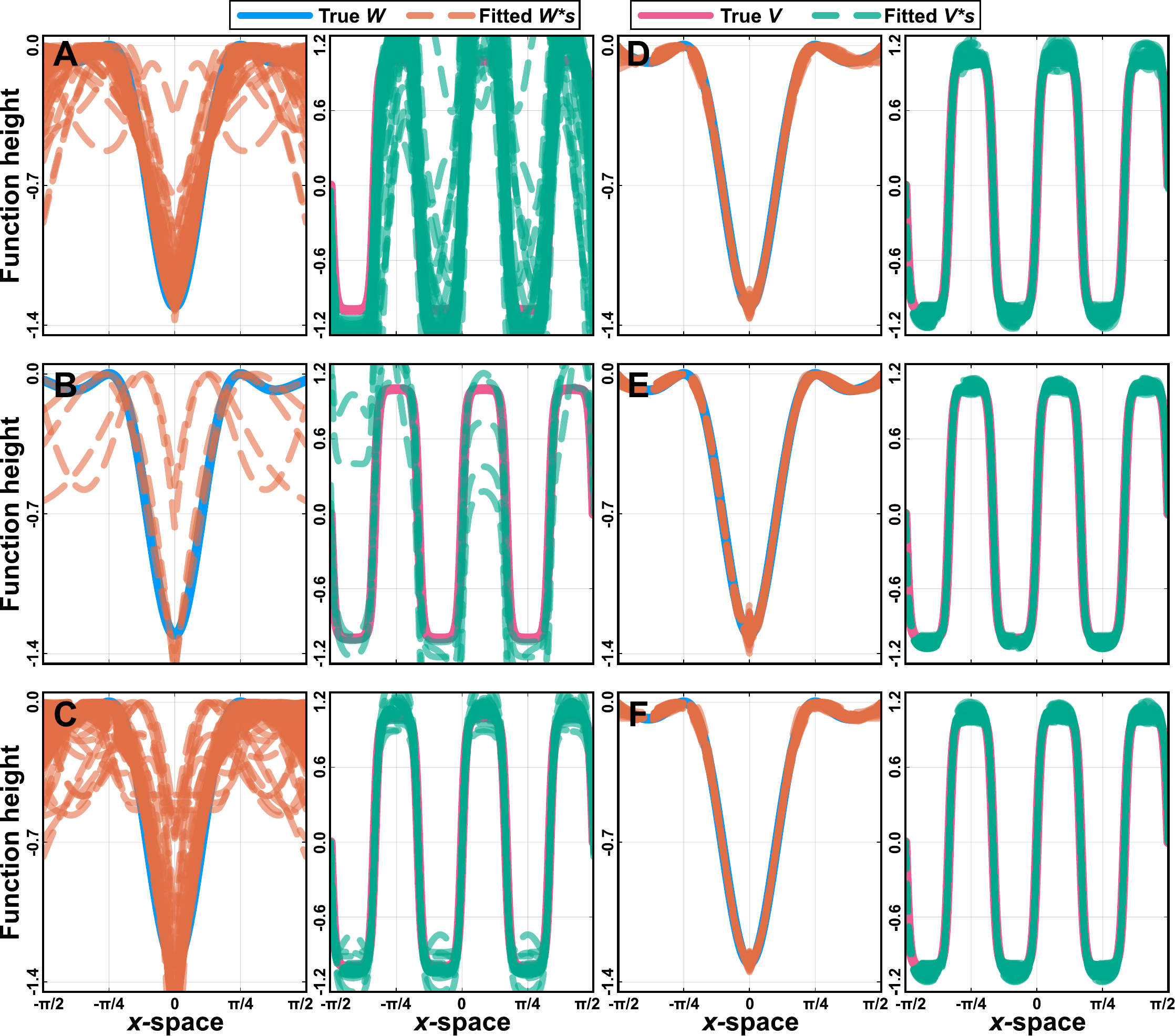}
    \caption{\textbf{Recovery of two functional components requires at least two steady-state solutions.} (A-C) For each of the solutions of the PDE in Figure~\ref{fig:fit_WVk_to_sols}, we attempt to fit both $W$ and $V$ using that solution only ($\kappa$ is set to be known). Using the approach from Figure~\ref{fig:fig_4_sup_practical_identifiability}, we plot all fitted $W^*$ and $V^*$ achieving a loss function value no more than three times that of the optimal one. In each case, a wide range of functional components generates good fits, suggesting that the recovery problem is non-identifiable. This is consistent with the result presented in Theorem~\ref{thm:identifiability}. (D-F) For each combination of two solutions of the solution of the PDE in Figure~\ref{fig:fit_WVk_to_sols}, we attempt to fit both $W$ and $V$ using those two solutions only. Here, in all cases, the ensemble plots show that we can correctly recover the true functions, confirming that practical identifiability holds when two solutions are available.}
    \label{fig:fig_4_sup_partial_sols}
\end{figure}

% ------------------------------------------------------------------------------------------------

\begin{figure}[htbp]
    \centering   \includegraphics[width=0.85\linewidth]{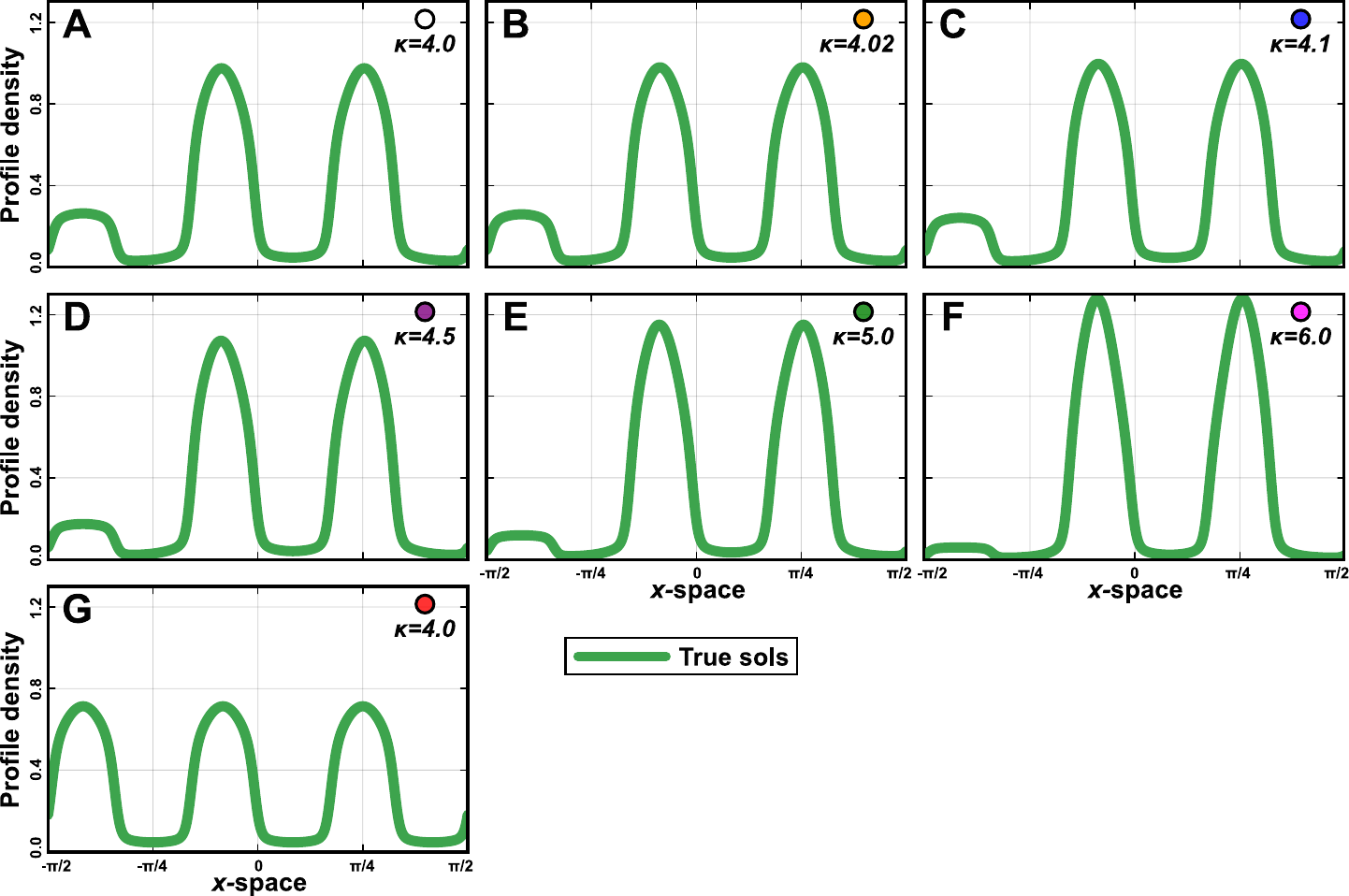}
    \caption{\textbf{Solution profiles vary across the PDE's bifurcation diagram.} (A-G) Solution profiles for the seven sample locations in Figure~\ref{fig:bif_analysis}A. Identities are marked by top right corner coloured dots. A-F correspond to solutions on the same branch, while G corresponds to a solution on a different branch. Note that profiles A,B,F,G are identical to Figure~\ref{fig:bif_analysis}C,D,E,F (but displayed here as well for completeness).}
    \label{fig:fig_5_sols}
\end{figure}

% ------------------------------------------------------------------------------------------------

\begin{figure}[htbp]
    \centering   
    \includegraphics[width=0.85\linewidth]{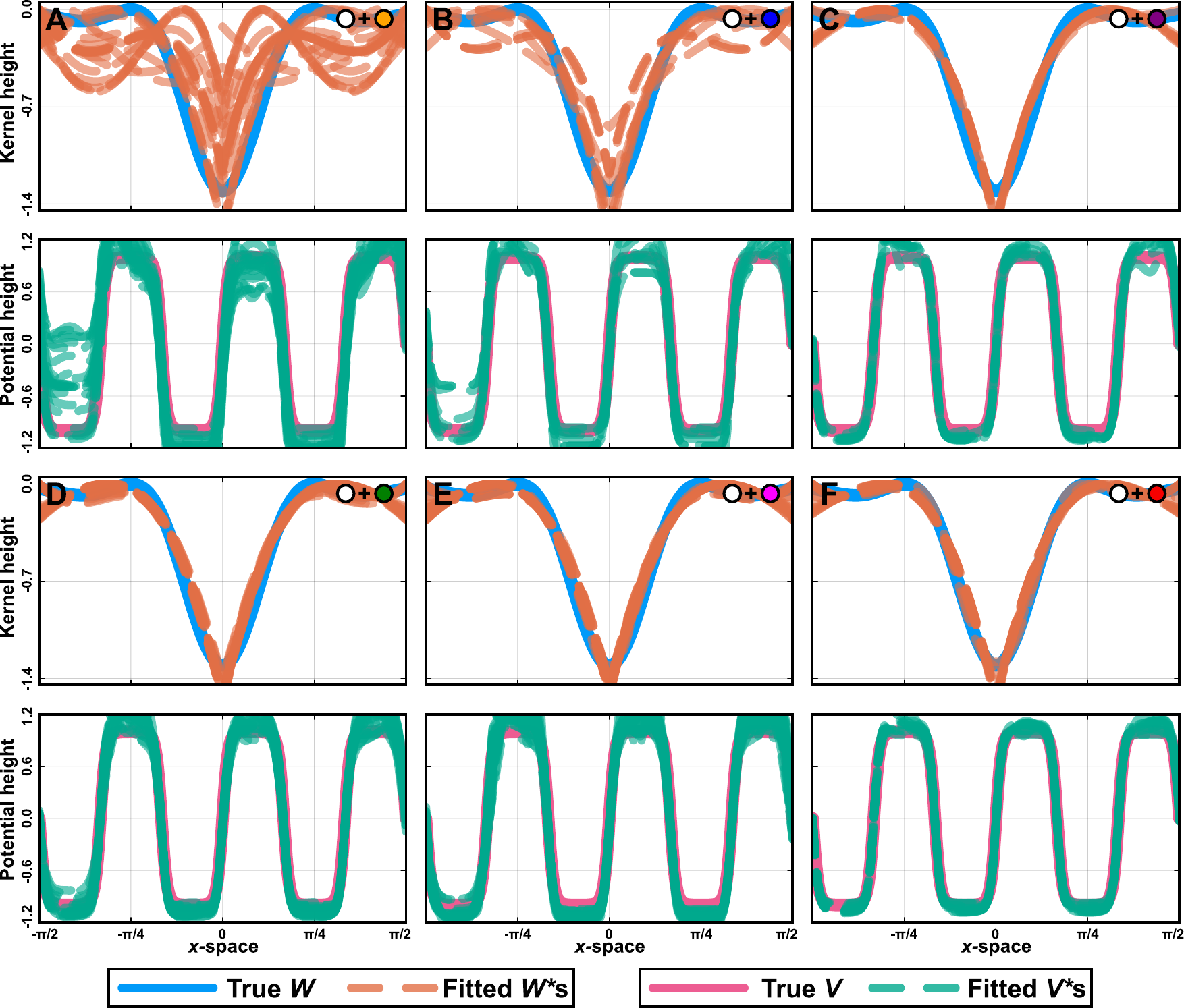}
    \caption{\textbf{$V$ and $W$ can be recovered using two solutions, conditioned on their bifurcation diagram sampling locations.} (A-F) For each of the combinations of two solutions described in Figure~\ref{fig:bif_analysis}B, we show the ensemble of recovered $W^*$ and $V^*$ (and compare them to the true functions). Using the approach from Figure~\ref{fig:fig_4_sup_practical_identifiability}, we plot all fitted $W^*$ and $V^*$ achieving a loss function value no more than three times that of the optimal one. These plots correspond to Figure~\ref{fig:bif_analysis}G-I, but include the fits for the solution combinations not displayed there. Same-branch recovery performance increases as the $\kappa$ distance between the two solutions used increases. However, already at a relatively small distance, $V$ and $W$ can be recovered well. Note that ensemble plots A,E,F are identical to Figure~\ref{fig:bif_analysis}G,H,I (but displayed here as well for completeness).}
    \label{fig:fig_5_dual_fits}
\end{figure}

% ------------------------------------------------------------------------------------------------

\end{document}